\documentclass[letterpaper]{article}
\usepackage[preprint]{aaai2027}
\usepackage[hyphens]{url}
\usepackage{graphicx}
\usepackage{natbib}
\usepackage{caption}
\usepackage{booktabs}
\usepackage{amsmath}
\usepackage{amssymb}
\definecolor{grarelink}{RGB}{91,155,213}
\usepackage[colorlinks=true,linkcolor=black,citecolor=green,urlcolor=grarelink]{hyperref}
\makeatletter
\let\grare@orig@PackageError\PackageError
\def\grare@aaai{aaai}
\def\PackageError#1#2#3{%
  \def\grare@pkg{#1}%
  \ifx\grare@pkg\grare@aaai
    \ifnum\pdfstrcmp{#2}{You must not use hyperref in AAAI papers.}=0
      \PackageWarning{aaai}{Hyperref check bypassed for this preprint}%
    \else
      \grare@orig@PackageError{#1}{#2}{#3}%
    \fi
  \else
    \grare@orig@PackageError{#1}{#2}{#3}%
  \fi}
\AtBeginDocument{\let\PackageError\grare@orig@PackageError}
\makeatother
\nocopyright

\title{GraRe: Grasp Candidate Re-Ranking for Frozen 6-DoF Grasp Detectors}
\author{Jibao Yuan\textsuperscript{\rm 1},
Yuhui Zhao\textsuperscript{\rm 1},
Yinzhen Lv\textsuperscript{\rm 2},
Chao Xu\textsuperscript{\rm 1},\\
Shun Li\textsuperscript{\rm 1},
Chenxi Deng\textsuperscript{\rm 1},
Shaofei Chen\textsuperscript{\rm 1}\corresponding}
\affiliations{\textsuperscript{\rm 1}College of Intelligence Science and Technology, National University of Defense Technology, Changsha, Hunan, China\\
\textsuperscript{\rm 2}Department of Engineering and Information Technology, The University of Melbourne, Parkville, Victoria 3010, Australia}

\begin{document}

\maketitle

\begin{abstract}
Existing 6-DoF grasp detectors typically rank grasp candidates by detector confidence. However, our analysis on GraspNet-1Billion shows that detector confidence is often poorly aligned with grasp quality, leaving successful grasp candidates at low ranks. Motivated by this observation, we study whether learned re-ranking can improve candidate ordering while keeping detector parameters and grasp candidates unchanged. We propose GraRe, which estimates grasp quality from candidate attributes, shell-stratified local geometry, and object context. Candidate attributes condition the local geometric and object-context representations, and a Transformer fuses all three feature types. The predicted quality is combined with detector confidence to produce the final ranking. Experiments on GraspNet-1Billion with five frozen detectors show consistent improvements, with gains of up to 13.56 points in Average AP. Real-robot experiments further demonstrate robust grasping in cluttered scenes. These results show that improving candidate ranking provides a practical way to enhance frozen 6-DoF grasp detectors.
Project page is available at \href{https://minakanmi-yuki.github.io/grare/}{\textcolor{grarelink}{\texttt{\textit{https://minakanmi-yuki.github.io/grare/}}}}.
\end{abstract}

\section{Introduction}

Grasping in cluttered scenes is fundamental to robotic manipulation in real-world environments. 
This capability supports applications such as industrial bin picking, warehouse automation, and household object manipulation~\citep{davella2024cepb,ciocarlie2014household}.
However, in such scenes, occlusion reduces object visibility, while nearby objects increase the risk of gripper collisions~\citep{jiang2021giga,wei2021gpr}.
For previously unseen objects, accurate 3D object models may be unavailable, motivating grasping methods that operate without predefined object models~\citep{boularias2015unknown}.
With advances in depth sensing and deep learning, data-driven methods for 6-DoF grasp detection have shown strong performance in cluttered scenes~\citep{sundermeyer2021contactgraspnet,breyer2020vgn}.

In general, data-driven 6-DoF grasp detection methods generate a set of grasp candidates from an RGB-D frame or point cloud. Each candidate contains a 6-DoF grasp pose, a gripper width, and an associated detector confidence. During execution, the robot attempts grasp candidates in descending order of detector confidence~\citep{fang2023robust,ma2022scalebalanced,wu2024economic}. Therefore, grasping performance depends not only on the quality of candidate generation but also on the accuracy of candidate ranking.

\begin{figure}[!t]
  \centering
  \includegraphics[width=\columnwidth]{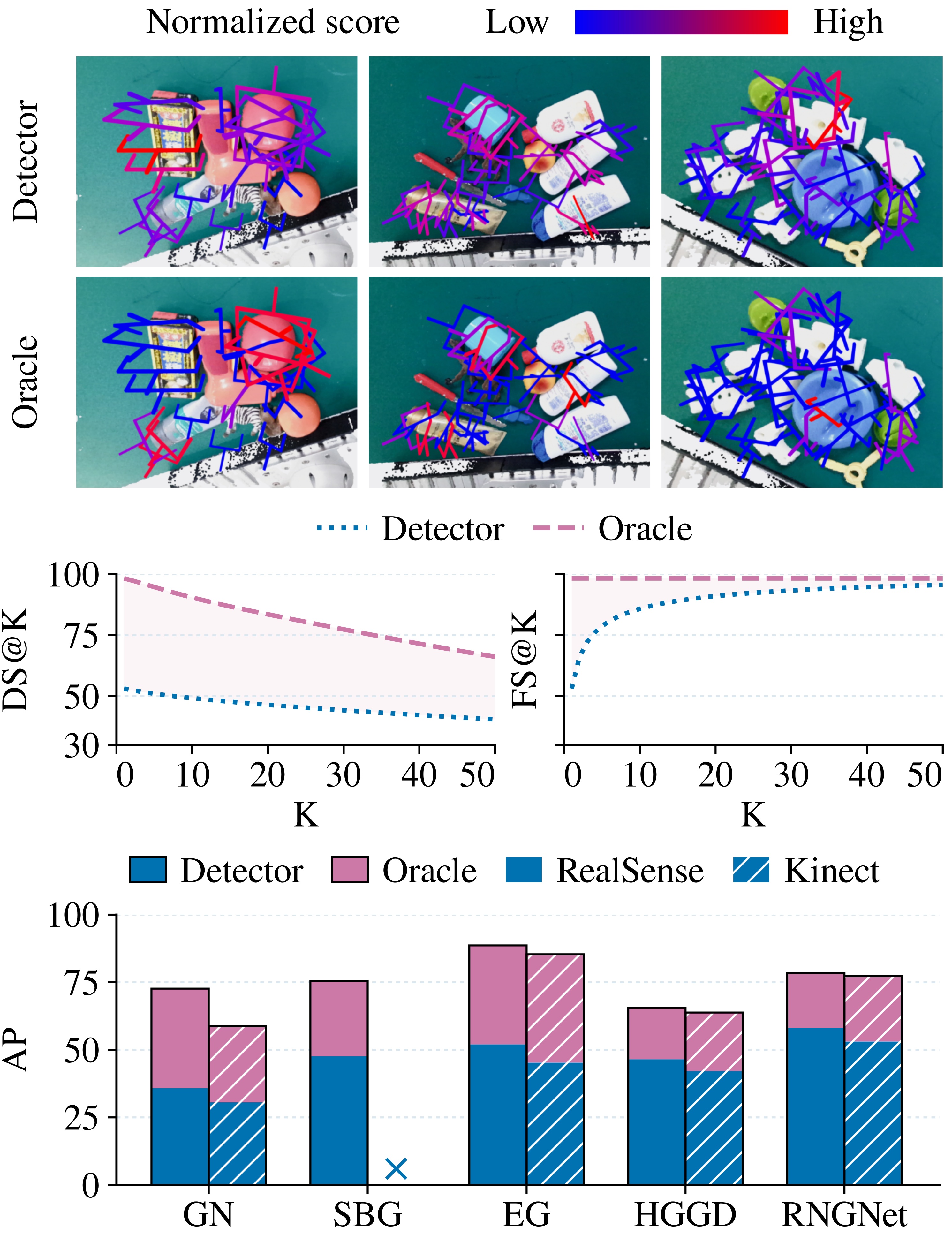}
  \caption{Ranking headroom analysis on GraspNet-1Billion shows that better ordering improves ranking metrics and AP.}
  \label{fig:teaser}
\end{figure}

Grasping failures may arise because a detector fails to generate successful grasp candidates or assigns low confidence to candidates that would succeed. If the latter is common, better candidate ordering could improve grasping performance. To examine this possibility, we compare the detector order with an oracle order that reorders the same grasp candidates using ground-truth labels (Figure~\ref{fig:teaser}). The three examples show that the detector and oracle prioritize grasp candidates in markedly different regions of the same scene. The gap in candidate ranking metrics reveals substantial room to improve the ordering of these candidates. Oracle ordering also yields higher Average AP for the same candidate sets on GraspNet-1Billion~\citep{fang2020graspnet}. This finding raises a key question: Can grasp candidates be ranked more accurately without modifying candidate generation?

Motivated by this question, we study grasp candidate re-ranking for frozen 6-DoF grasp detectors. Given the detector's grasp candidates, the task predicts a new ordering while keeping all candidates unchanged. We propose GraRe to address this task.
Its key idea is to evaluate each candidate using three complementary types of information: candidate attributes, shell-stratified local geometry, and object context.
The shell-stratified representation preserves geometry across different distances from the gripper, while object context describes the candidate relative to the visible object.
Candidate attributes condition both geometric representations before a Transformer fuses all three to predict grasp quality.
GraRe combines the predicted quality with detector confidence for final ranking.

Experiments on GraspNet-1Billion with five frozen detectors show consistent improvements, with gains of up to 13.56 points in Average AP.
Ranking analysis shows that GraRe promotes successful grasp candidates, while real-robot experiments demonstrate robust grasping in cluttered scenes.

The main contributions of this work are summarized as follows:
\begin{itemize}

\item We analyze the mismatch between detector confidence and grasp quality through detector--oracle comparisons on unchanged grasp candidates. This analysis reveals ranking headroom and motivates learning to improve the ordering produced by frozen 6-DoF grasp detectors.

\item We propose GraRe, which uses candidate attributes to condition shell-stratified local geometry and object context. A Transformer fuses the three representations to predict grasp quality. The predicted quality is combined with detector confidence for final ranking.

\item We evaluate GraRe with five frozen detectors on GraspNet-1Billion and conduct real-robot experiments. The results show Average AP gains of up to 13.56 points, improved ranking of successful grasp candidates, and robust grasping in cluttered scenes.

\end{itemize}

\section{Related Work}

Data-driven 6-DoF grasp detectors generate grasp poses and associated scores from visual observations. 
GraspNet-1Billion provides dense annotations and a common evaluation protocol for grasping in clutter~\citep{fang2020graspnet}. Subsequent methods improve detection through graspness, scale-balanced learning, heatmap guidance, and efficient supervision~\citep{wang2021graspness,ma2022scalebalanced,chen2023hggd,wu2024economic}. 

Learned grasp evaluation and ranking have been explored through several approaches. Dex-Net 2.0 trains a grasp-quality CNN with synthetic depth images and analytical robustness labels for parallel-jaw grasp selection~\citep{mahler2017dexnet2}. GPD and PointNetGPD evaluate sampled candidates using local geometry~\citep{tenpas2017grasp,liang2019pointnetgpd}, while 6-DoF GraspNet combines variational generation with a separately trained evaluator and iterative pose refinement~\citep{mousavian20196dof}. Kent and Toris learn pairwise preferences to rank externally supplied grasp candidates~\citep{kent2018pairwiseranking}. Duarte et al. incorporate shape-completion uncertainty into existing scores to reorder grasps produced by GPD and PointNetGPD~\citep{duarte2025uncertainty}. GraspGen trains a separate discriminator on diffusion-generated grasps, combining candidate poses with frozen object-encoder features to learn generator-specific failure modes~\citep{murali2025graspgen}. GraRe likewise learns from generated candidates, but uses outputs from the frozen detector being augmented and evaluates the same scoring design separately across detector architectures.

GRaCE ranks grasps using a utility function over prioritized criteria and additionally refines poses through GRaCE-OPT~\citep{taunyazov2024grace}; GPR also improves grasp proposals through pose refinement~\citep{wei2021gpr}. Ma et al. train an independent ScoreNet on noisy object clouds and labeled grasps, then combine score and contact objectives to iteratively refine poses~\citep{ma2024domainprior}. LERF-TOGO combines semantic relevance with grasp confidence for language-guided selection~\citep{rashid2023lerftogo}. Building on these studies, GraRe examines learned quality prediction for improving the ordering of unchanged candidate sets from multiple frozen detectors.

Geometric representations determine the information available for candidate evaluation. GtG 2.0 represents points inside and around the gripper as graphs and regresses grasp-quality scores from candidates produced by a Grasp Pose Generator (GPG) and relabeled with the GraspNet evaluator~\citep{moghadam2025gtg2}. Its scalar labels already distinguish collisions from low-quality grasps. GraRe uses evaluator-derived supervision for a friction-based quality target and separate collision and empty-grasp classifiers. GraphGrasp models local--global scene features and spatial relationships between objects~\citep{graphgrasp2026}, while ShellNet aggregates point-cloud features within concentric spherical shells~\citep{zhang2019shellnet}. GraRe conditions gripper-frame radial-shell and visible-object representations on candidate attributes, fuses them with a Transformer, and combines predicted quality with detector confidence. Evaluation on unchanged candidates isolates the contribution of ranking to grasp detection performance.

\begin{figure*}[t]
  \centering
  \includegraphics[width=\textwidth]{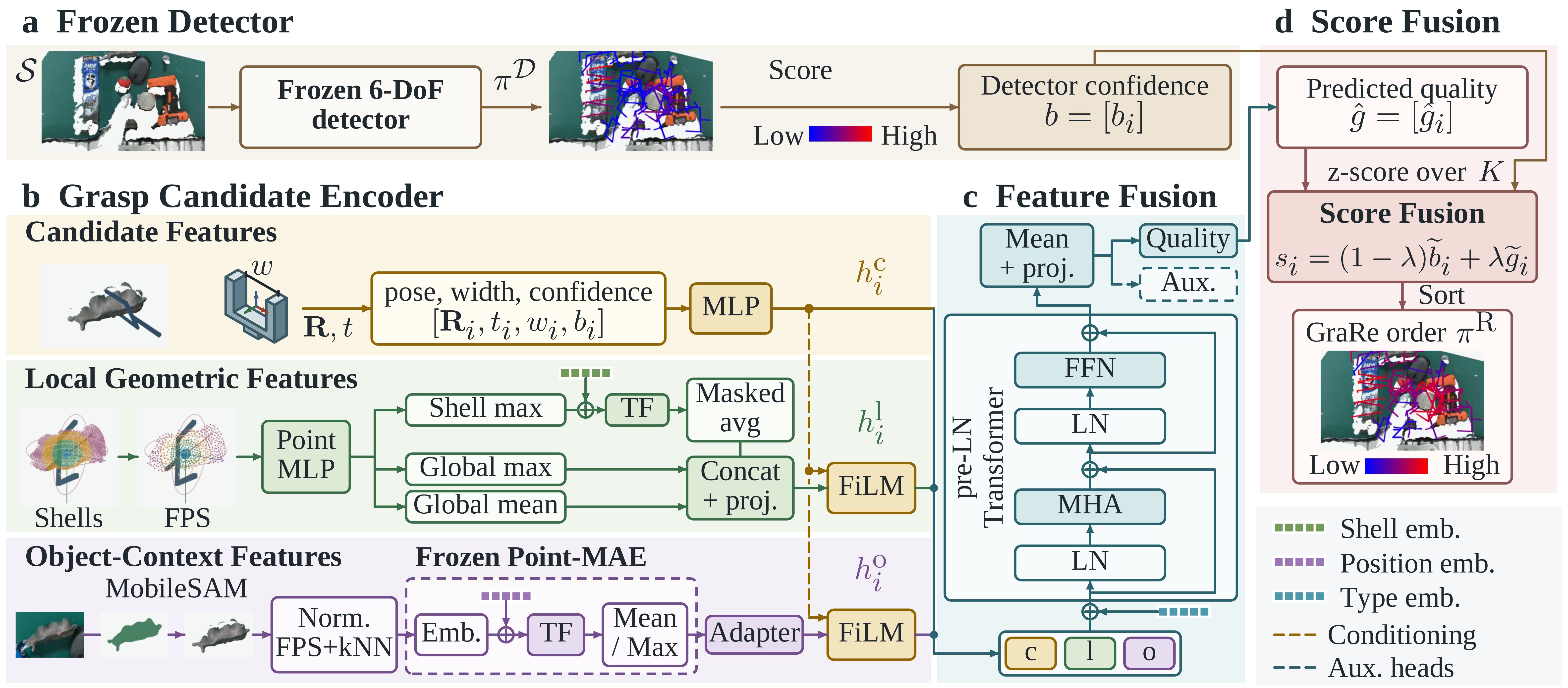}
  \caption{Overview of GraRe. (a) A frozen detector generates grasp candidates. (b) Candidate features condition local geometric and object-context features through FiLM. (c) A Transformer fuses the three representations to predict grasp quality. (d) Score Fusion combines predicted quality with detector confidence to re-rank the unchanged grasp candidates. \mbox{TF denotes Transformer block.}}
  \label{fig:architecture}
\end{figure*}

\section{Problem Formulation}
Let $\mathcal{S}$ be a scene point cloud and $I$ its aligned RGB image. A frozen 6-DoF grasp detector $\mathcal{D}$ for a parallel-jaw gripper takes $\mathcal{S}$ as input and produces $K$ grasp candidates, denoted by $\mathcal{C}=\{c_i\}_{i=1}^{K}$. Each candidate is represented as $c_i=(p_i,w_i,b_i)$. The 6-DoF grasp pose is $p_i=(\mathbf{R}_i,t_i)\in\mathrm{SE}(3)$. The rotation $\mathbf{R}_i\in\mathrm{SO}(3)$ and translation $t_i\in\mathbb{R}^{3}$ specify the gripper pose in the camera coordinate system. The scalar $w_i\in\mathbb{R}^{+}$ is the gripper width, and $b_i\in\mathbb{R}$ is the detector confidence. The detector ranks the candidates by $b_i$. We denote this order by $\pi^{\mathcal{D}}\in\Pi_K$.

Grasp candidate re-ranking keeps $\mathcal{C}$ unchanged and predicts a new ordering of its elements. Let $\Pi_K$ denote the set of all permutations of the $K$ candidates. For any $\pi\in\Pi_K$, $\mathcal{E}(\mathcal{C},\pi)$ measures the quality of the resulting ranking. On GraspNet-1Billion, $\mathcal{E}$ is Average Precision (AP) under the official evaluation protocol, which uses multiple friction coefficients~\citep{fang2020graspnet}. The best possible ranking for $\mathcal{C}$ is $\pi^\star=\arg\max_{\pi\in\Pi_K}\mathcal{E}(\mathcal{C},\pi)$. A learned re-ranker predicts $\pi^{\mathrm{R}}$. Its goal is to improve over the detector order, so that $\mathcal{E}(\mathcal{C},\pi^{\mathrm{R}})\geq\mathcal{E}(\mathcal{C},\pi^{\mathcal{D}})$.

GraRe does not predict $\pi^{\mathrm{R}}$ directly. Let $\mathcal{T}_{\theta}$ denote GraRe with learnable parameters $\theta$. For each candidate, it predicts a grasp quality $\hat{g}_i=\mathcal{T}_{\theta}(c_i,\mathcal{S},I)$ from the candidate, scene point cloud, and aligned RGB image. We combine $\hat{g}_i$ with the detector confidence $b_i$ to obtain the final ranking score $s_i$. Sorting the candidates by $s_i$ in descending order gives the re-ranked order $\pi^{\mathrm{R}}$.

\section{Method}
GraRe re-ranks grasp candidates using candidate attributes, local geometry, and object context. It combines the predicted quality with detector confidence, and Figure~\ref{fig:architecture} shows the overall architecture.

\subsection{Feature Extraction}
Given a grasp candidate $c_i=(p_i,w_i,b_i)$, the scene point cloud $\mathcal{S}$, and its aligned RGB image $I$, GraRe extracts candidate, local geometric, and object-context features. These features are fused to predict $\hat{g}_i=\mathcal{T}_{\theta}(c_i,\mathcal{S},I)$.

\textbf{Candidate Features.} Candidate attributes provide information not contained in the geometric features. We encode the grasp pose $p_i$, gripper width $w_i$, and detector confidence $b_i$ with an MLP:
\begin{equation}
h_i^{\mathrm{candidate}}
=
f_{\mathrm{candidate}}
\left(
[p_i,w_i,b_i]
\right)
\end{equation}
The output $h_i^{\mathrm{candidate}}$ conditions the local geometric and object-context features and is later fused with them.

\textbf{Local Geometric Features.} Local geometry describes possible finger contacts and free space for gripper motion. Applying farthest point sampling (FPS)~\citep{qi2017pointnet2} to the full neighborhood may undersample some distance ranges. We therefore transform scene points into the gripper coordinate system and partition them into radial shells with boundaries $r_0<r_1<\cdots<r_J$:
\begin{equation}
\begin{aligned}
\mathcal{Y}_{i,j}
&=\bigl\{\mathbf{R}_i^{\top}(x-t_i)\mid x\in\mathcal{S},\ \\
&\qquad\lVert x-t_i\rVert_2\in[r_{j-1},r_j)\bigr\}
\end{aligned}
\end{equation}

We apply FPS separately to each shell:
\begin{equation}
\mathcal{X}_{i,j}^{\mathrm{local}}
=\textsc{FPS}_{\kappa_j}
\left(
\mathcal{Y}_{i,j}
\right)
\end{equation}
Here, $\kappa_j$ is the sampling budget. The local point set is:
\begin{equation}
\mathcal{X}_i^{\mathrm{local}}
=
{\textstyle\bigcup\nolimits_{j=1}^{J}}
\mathcal{X}_{i,j}^{\mathrm{local}}
\end{equation}

Each shell is encoded by a shared MLP $\phi_{\mathrm{local}}$ and max pooling~\citep{qi2017pointnet}. Let $m_{i,j}$ indicate whether shell $j$ is nonempty:
\begin{equation}
\eta_{i,j}=\max_{y\in\mathcal{X}_{i,j}^{\mathrm{local}}}\phi_{\mathrm{local}}(y), \qquad m_{i,j}=1
\end{equation}
Empty shells are excluded by the attention mask $\mathcal{M}_i$. We add a learned embedding $\alpha_j$ to retain the radial position of each shell:
\begin{equation}
\widetilde{\eta}_i=\left[\eta_{i,j}+\alpha_j\right]_{j=1}^{J}
\end{equation}
A pre-normalized Transformer models interactions among the shell features~\citep{vaswani2017attention,xiong2020layer}:
\begin{equation}
\left\{
  \begin{array}{ll}
  \overline{\eta}_i=\widetilde{\eta}_i+\mathrm{MHA}\left(\mathrm{LN}(\widetilde{\eta}_i);\mathcal{M}_i\right)
  \\
  G_i^{\mathrm{local}}=\overline{\eta}_i+\mathrm{FFN}\left(\mathrm{LN}(\overline{\eta}_i)\right)
  \end{array}
  \right.
\end{equation}
Here, $\mathrm{MHA}$ is multi-head self-attention and $\mathrm{FFN}$ is a two-layer feed-forward network with GELU activation.

We average the valid Transformer outputs and compute global max and mean features:
\begin{equation}
\left\{
\begin{array}{l}
z_i^{\mathrm{shell}}
=
\frac{
\sum_{j=1}^{J}m_{i,j}G_{i,j}^{\mathrm{local}}
}{
\sum_{j=1}^{J}m_{i,j}
}
\\
z_i^{\mathrm{max}}
=
\max_{y\in\mathcal{X}_i^{\mathrm{local}}}
\phi_{\mathrm{local}}(y)
\\
z_i^{\mathrm{mean}}
=
\frac{1}{|\mathcal{X}_i^{\mathrm{local}}|}
\sum_{y\in\mathcal{X}_i^{\mathrm{local}}}
\phi_{\mathrm{local}}(y)
\end{array}
\right.
\end{equation}
The three features are concatenated and projected to a local descriptor:
\begin{equation}
u_i^{\mathrm{local}}=f_{\mathrm{local}}\left([z_i^{\mathrm{shell}},z_i^{\mathrm{max}},z_i^{\mathrm{mean}}]\right)
\end{equation}
Because local geometry must be interpreted with the candidate attributes, feature-wise linear modulation (FiLM)~\citep{perez2018film} conditions the descriptor on $h_i^{\mathrm{candidate}}$:
\begin{equation}
h_i^{\mathrm{local}}=\mathrm{FiLM}_{\mathrm{local}}\left(u_i^{\mathrm{local}},h_i^{\mathrm{candidate}}\right)
\end{equation}

\textbf{Object-Context Features.} Local features cover only the gripper neighborhood, so we use the visible shape of the associated object as context. We project the grasp center $t_i$ into the aligned RGB image and use the pixel as a prompt for MobileSAM~\citep{zhang2023mobilesam}. The predicted mask selects 3D points from $\mathcal{S}$, which are sampled by FPS to form $\mathcal{X}_i^{\mathrm{object}}$. The RGB image is used only for mask generation.

Clutter and occlusion make $\mathcal{X}_i^{\mathrm{object}}$ incomplete. We encode this visible object point cloud with Point-MAE, which is pretrained by reconstructing masked point-cloud patches~\citep{pang2022pointmae}. Before encoding, we normalize the points to the unit sphere and group them into patches using FPS and $k$-nearest neighbors (kNN):
\begin{equation}
G_i^{\mathrm{object}}=\mathrm{PointMAE}
\left(
\nu(\mathcal{X}_i^{\mathrm{object}})
\right)
\end{equation}
Here, $\nu$ denotes normalization and patch grouping. We freeze the Point-MAE backbone and map its mean- and max-pooled patch features through a trainable adapter:
\begin{equation}
u_i^{\mathrm{object}}
=
f_{\mathrm{object}}
\left(
\mathrm{Pool}_{\mathrm{mean,max}}
\left(
G_i^{\mathrm{object}}
\right)
\right)
\end{equation}
Because one object can support grasps of different quality, FiLM conditions the object descriptor on $h_i^{\mathrm{candidate}}$:
\begin{equation}
h_i^{\mathrm{object}}=\mathrm{FiLM}_{\mathrm{object}}
\left(
u_i^{\mathrm{object}},
h_i^{\mathrm{candidate}}
\right)
\end{equation}
\subsection{Feature Fusion}
Local geometry is interpreted with grasp pose and gripper width. We fuse candidate, local geometric, and object-context features. Let $\mathcal{R}=\{\mathrm{candidate},\mathrm{local},\mathrm{object}\}$ denote the three feature types. We add a learned type embedding $e^r$ and use a pre-normalized Transformer:
\begin{equation}
\left\{
\begin{array}{l}
Z_i
=
\left[h_i^{r}+e^{r}\right]_{r\in\mathcal{R}}
,\quad
H_i^{\mathrm{fusion}}
=
\mathrm{TF}_{\mathrm{fusion}}
\left(
Z_i
\right)
\\
h_i^{\mathrm{fusion}}
=
f_{\mathrm{fusion}}
\left(
\frac{1}{|\mathcal{R}|}
\sum_{r\in\mathcal{R}}
H_{i,r}^{\mathrm{fusion}}
\right)
\\
\hat{g}_i
=
f_{\mathrm{quality}}
\left(
h_i^{\mathrm{fusion}}
\right)
\end{array}
\right.
\end{equation}
\subsection{Training Objective}
During training, the GraspNet evaluator uses annotated scene geometry to obtain the minimum friction coefficient $\mu_i$ required for force closure~\citep{fang2020graspnet}. It also provides the collision and empty-grasp labels $y_i^{\mathrm{coll}}$ and $y_i^{\mathrm{empty}}$. We define the analytical grasp-quality target as:
\begin{equation}
g_i=\tau-\overline{\mu}_i
\end{equation}
Here, $\tau$ is the success threshold. We set $\overline{\mu}_i=\mu_{\mathrm{worst}}$ for non-finite $\mu_i$ and $\overline{\mu}_i=\mu_i$ otherwise. Lower friction produces a higher target, preserving differences discarded by binary labels.

Because $g_i$ does not identify the failure type, we add collision and empty-grasp classifiers. An object-classification head regularizes the object-context feature. The total loss is:
\begin{equation}
\mathcal{L}
=
\mathcal{L}_{\mathrm{quality}}
+
\omega_{\mathrm{coll}}\mathcal{L}_{\mathrm{coll}}
+
\omega_{\mathrm{empty}}\mathcal{L}_{\mathrm{empty}}
+
\omega_{\mathrm{obj}}\mathcal{L}_{\mathrm{obj}}
\end{equation}
The quality loss is Smooth-$\ell_1$, the binary heads use binary cross-entropy, and the object head uses cross-entropy. The coefficients $\omega_{\mathrm{coll}}$, $\omega_{\mathrm{empty}}$, and $\omega_{\mathrm{obj}}$ weight the auxiliary losses. These losses are used only during training.

\subsection{Score Fusion}
The detector confidence $b_i$ and predicted quality $\hat{g}_i$ have different scales. We apply z-score normalization within each candidate set to obtain $\widetilde{b}_i$ and $\widetilde{g}_i$, then compute:
\begin{equation}
s_i=(1-\lambda)\widetilde{b}_i+\lambda\widetilde{g}_i
\end{equation}
where $\lambda\in[0,1]$ controls the contribution of predicted quality. Sorting $s_i$ in descending order gives the final ranking.

\section{Experiments}
\subsection{Benchmark, Baselines, and Metrics}

\textbf{Benchmark.} We evaluate on GraspNet-1Billion~\citep{fang2020graspnet} using RealSense and Kinect observations and the official split of 100 training and 90 test scenes. The Seen, Similar, and Novel scene splits each contain 30 test scenes.

\textbf{Baselines.} We compare GraspNet-Baseline (GN)~\citep{fang2020graspnet,fang2023robust}, Scale-Balanced Grasp (SBG)~\citep{ma2022scalebalanced}, EconomicGrasp (EG)~\citep{wu2024economic}, HGGD~\citep{chen2023hggd}, and RNGNet~\citep{chen2024rngnet} with and without GraRe. These five frozen 6-DoF grasp detectors cover different architectures and performance levels. Detector parameters and grasp candidates remain unchanged.

\textbf{Metrics.} We evaluate GraRe in terms of candidate ranking quality and grasping performance. For ranking evaluation, a successful candidate is non-empty, collision-free, and satisfies force closure with $\mu_i\leq0.4$. Inspired by the logarithmic rank discount used in discounted cumulative gain~\citep{jarvelin2002cumulated}, we define Discounted Success@K (DS@K) as the percentage of successful candidates weighted by $1/\log_2(r+1)$ at rank $r$. For each frame, we divide the weighted success sum by the sum of these weights over its first $\min(K,N)$ candidates, where $N$ is the candidate count, and then average across frames. Frames with no candidates contribute zero. Frame Success@K (FS@K) measures the percentage of frames with at least one successful candidate in the top $K$. Average AP measures grasping performance under the official GraspNet-1Billion protocol.

\subsection{Ranking and Performance}
\begin{figure}[t]
  \centering
  \includegraphics[width=\columnwidth]{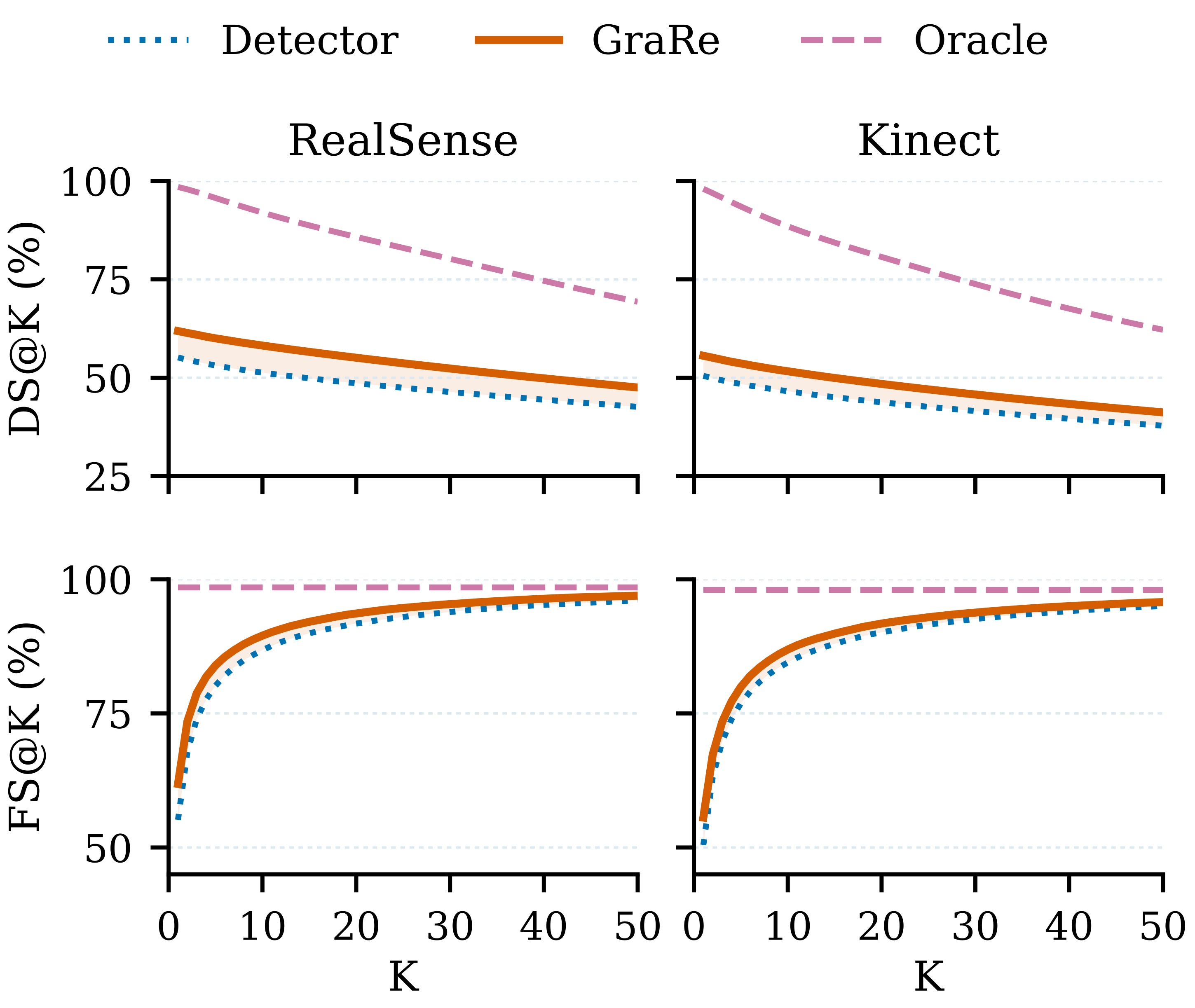}
  \caption{Ranking comparison of $\pi^{\mathcal D}$, $\pi^{\mathrm R}$, and $\pi^\star$ on unchanged grasp candidates, averaged over the evaluated detectors for each camera.}
  \label{fig:ranking_analysis}
\end{figure}

For DS@K and FS@K, the oracle $\pi^\star$ places successful candidates first. Its high FS@1 on both cameras shows that most frames already contain a successful candidate, while the gap to $\pi^{\mathcal D}$ identifies a loss from candidate ordering (Figure~\ref{fig:ranking_analysis}). GraRe narrows this gap, especially at small $K$. As $K$ grows, FS@K approaches the oracle, yet the DS@K advantage of $\pi^{\mathrm R}$ persists. GraRe therefore makes successful candidates available earlier and improves their ordering beyond finding a single successful grasp.

  \begin{table}[t]
    \centering
    {\small
    \setlength{\tabcolsep}{2.0pt}
    \begin{tabular*}{\columnwidth}{@{\extracolsep{\fill}}l|cccc@{}}
    \toprule
    \textbf{Method}
    & \textbf{Seen}
    & \textbf{Similar}
    & \textbf{Novel}
    & \textbf{Average} \\
    \midrule
    GN
    & 47.83/41.98 & 42.79/37.57 & 16.94/12.24 & 35.86/30.60 \\
    $\mathrm{GraRe}_{\mathrm{GN}}$
    & \textbf{64.36}/\textbf{54.06} & \textbf{58.44}/\textbf{46.38} & \textbf{25.45}/\textbf{15.97} & \textbf{49.42}/\textbf{38.80} \\
    SBG
    & 62.27/- & 56.92/- & 23.79/- & 47.66/- \\
    $\mathrm{GraRe}_{\mathrm{SBG}}$
    & \textbf{68.47}/- & \textbf{62.55}/- & \textbf{27.24}/- & \textbf{52.75}/- \\
    EG
    & 69.31/63.76 & 61.49/52.42 & 25.28/19.61 & 52.03/45.26 \\
    $\mathrm{GraRe}_{\mathrm{EG}}$
    & \textbf{74.92}/\textbf{69.72} & \textbf{65.16}/\textbf{57.47} & \textbf{28.35}/\textbf{22.06} & \textbf{56.14}/\textbf{49.75} \\
    HGGD
    & 63.37/60.43 & 52.28/46.74 & 23.83/19.23 & 46.49/42.13 \\
    $\mathrm{GraRe}_{\mathrm{HGGD}}$
    & \textbf{67.58}/\textbf{64.95} & \textbf{55.37}/\textbf{50.03} & \textbf{25.66}/\textbf{19.85} & \textbf{49.53}/\textbf{44.94} \\
    RNGNet
    & 75.32/73.50 & 66.95/59.20 & 32.03/26.44 & 58.10/53.05 \\
    $\mathrm{GraRe}_{\mathrm{RNG}}$
    & \textbf{77.67}/\textbf{75.34} & \textbf{69.64}/\textbf{60.86} & \textbf{32.63}/\textbf{26.58} & \textbf{59.98}/\textbf{54.26} \\
    \bottomrule
    \end{tabular*}
    }
    \caption{AP on GraspNet-1Billion, reported as RealSense/Kinect. GraRe uses $\lambda=1.0$ for GN, $\lambda=0.8$ for SBG, EG, and HGGD, and $\lambda=0.5$ for RNGNet.}
    \label{tab:comparison}
\end{table}

Under the official GraspNet-1Billion protocol, GraRe improves Average AP across all nine detector-camera configurations (Table~\ref{tab:comparison}). The gains are largest for GN, reaching 13.56 and 8.20 points on RealSense and Kinect, respectively. RNGNet, the strongest detector baseline, also gains 1.88 and 1.21 points. The gains on RNGNet show that candidate ordering remains a source of improvement even for high-performing detectors. Together, these results show that better ordering of unchanged grasp candidates translates into stronger grasping performance without modifying candidate generation.

\subsection{Ablation Studies}

\begin{table}[t]
\centering
{\small
\setlength{\tabcolsep}{2.0pt}
\begin{tabular*}{\columnwidth}{@{\extracolsep{\fill}}l|ccccc@{}}
  \toprule
  \textbf{Variant} & \textbf{GN} & \textbf{SBG} & \textbf{EG} & \textbf{HGGD} & \textbf{RNGNet} \\
  \midrule
  Detector & 35.86 & 47.66 & 52.03 & 46.49 & 58.10 \\
  Full GraRe & 49.42 & 52.75 & 56.14 & 49.53 & 59.98 \\
  w/o Candidate & 47.63 & 51.74 & 54.19 & 47.99 & 59.98 \\
  w/o Local & 44.99 & 50.60 & 54.45 & 47.77 & 58.76 \\
  w/o Object & 48.71 & 52.59 & 55.98 & 49.92 & 59.83 \\
  w/o Shell-wise FPS & 48.27 & 52.64 & 55.96 & 47.90 & 58.69 \\
  w/ Concat Fusion & 48.72 & 52.73 & 56.04 & 49.88 & 59.85 \\
  w/o Aux. Losses & 48.92 & 52.27 & 55.70 & 49.47 & 59.77 \\
  \bottomrule
\end{tabular*}
}
\caption{Ablation study on RealSense, reporting Average AP with the Score Fusion weights in Table~\ref{tab:comparison}.}
\label{tab:ablation_realsense}
\end{table}

\begin{table}[t]
\centering
{\small
\setlength{\tabcolsep}{2.0pt}
\begin{tabular*}{\columnwidth}{@{\extracolsep{\fill}}l|ccccc@{}}
  \toprule
  \textbf{Variant} & \textbf{GN} & \textbf{SBG} & \textbf{EG} & \textbf{HGGD} & \textbf{RNGNet} \\
  \midrule
  Detector & 30.60 & -- & 45.26 & 42.13 & 53.05 \\
  Full GraRe & 38.80 & -- & 49.75 & 44.94 & 54.26 \\
  w/o Candidate & 37.21 & -- & 48.15 & 43.73 & 53.79 \\
  w/o Local & 36.48 & -- & 48.43 & 43.15 & 53.60 \\
  w/o Object & 38.04 & -- & 49.73 & 45.05 & 54.44 \\
  w/o Shell-wise FPS & 38.12 & -- & 49.71 & 43.33 & 53.66 \\
  w/ Concat Fusion & 38.12 & -- & 49.92 & 45.24 & 54.42 \\
  w/o Aux. Losses & 38.50 & -- & 49.40 & 44.55 & 54.10 \\
  \bottomrule
\end{tabular*}
}
\caption{Ablation study on Kinect, reporting Average AP with the Score Fusion weights in Table~\ref{tab:comparison}.}
\label{tab:ablation_kinect}
\end{table}

Tables~\ref{tab:ablation_realsense} and \ref{tab:ablation_kinect} assess the contributions of GraRe's features, sampling, fusion, and supervision.
Candidate Features improve most evaluated configurations. Their contribution supports conditioning geometric features on candidate attributes, so that geometry is interpreted with the grasp pose and gripper width. Local Geometric Features improve every evaluated detector-camera configuration, supporting the use of possible finger contacts and free space for gripper motion in quality estimation. Shell-wise FPS improves most available comparisons, although the EG Kinect sampling variant is 0.04 AP points lower than Full GraRe.

Object-Context Features improve GN on both cameras, SBG on RealSense, EG on both cameras, and RNGNet on RealSense. Their contribution therefore depends on the detector and camera.

The Concat Fusion variant retains all three feature types and replaces Transformer-based Feature Fusion with concatenation and an MLP. Full GraRe improves GN and SBG over this variant, whereas EG Kinect is 0.17 AP points higher with Concat Fusion. These results indicate that the benefit of Transformer-based fusion depends on the detector-camera configuration.

Auxiliary losses improve most configurations. Collision and empty-grasp supervision distinguish failure types not identified by the scalar quality target, while object-classification supervision regularizes Object-Context Features. Their gains support learning these complementary distinctions alongside grasp quality.

\subsection{Fusion Weight Sensitivity}

\begin{figure}[t]
  \centering
  \includegraphics[width=\columnwidth]{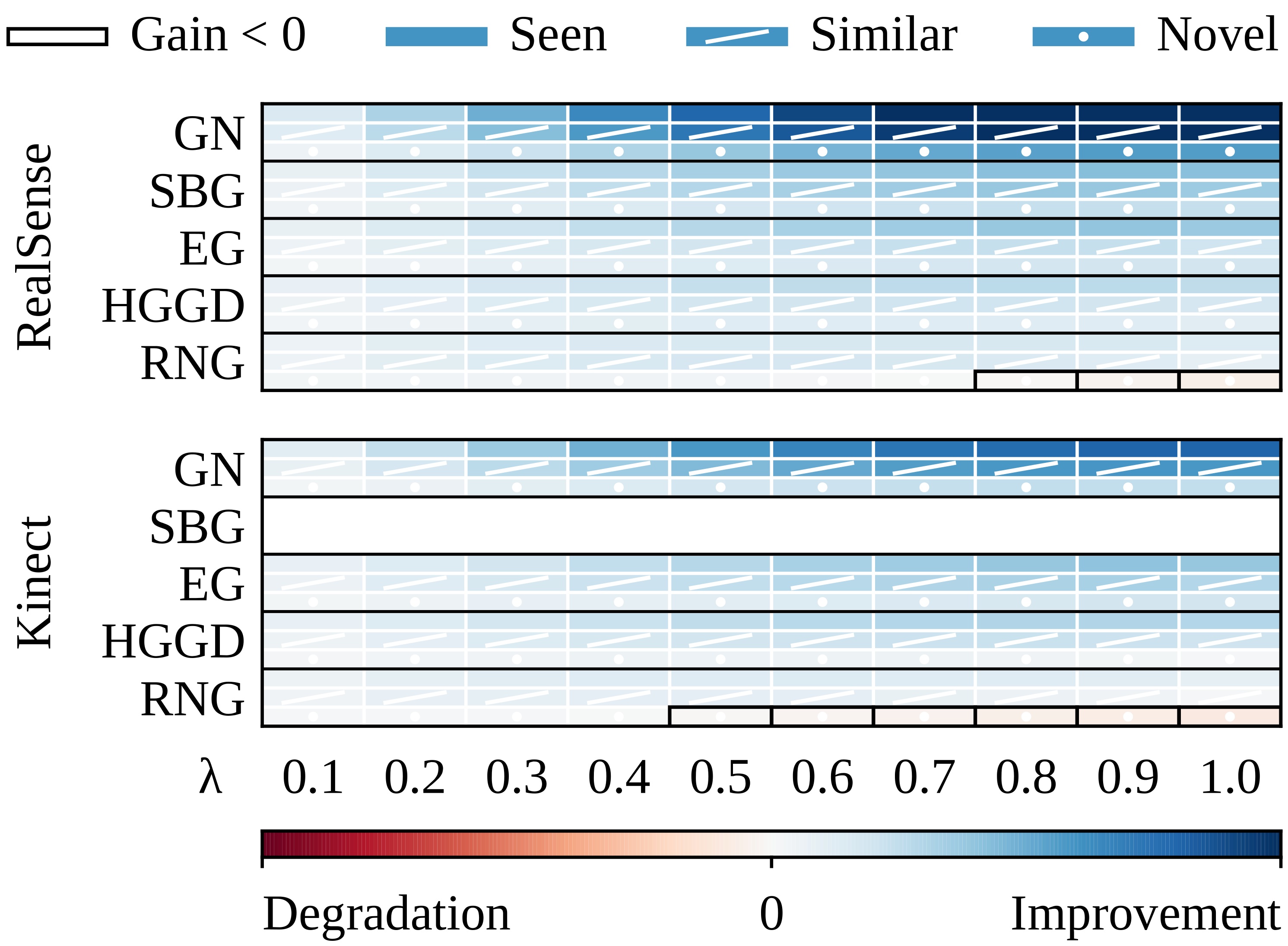}
  \caption{Average scene-level AP gain over the detector order $\pi^{\mathcal D}$ across Score Fusion weights $\lambda$.}
  \label{fig:lambda_sensitivity}
\end{figure}

We sweep $\lambda$ from 0.1 to 1.0 in steps of 0.1. Seen and Similar scene splits generally favor larger $\lambda$, but gains often peak before $\lambda=1$ (Figure~\ref{fig:lambda_sensitivity}). GN favors values close to 1 on both cameras, while SBG, EG, and HGGD peak between 0.7 and 0.9 on Seen and Similar. RNGNet favors more balanced weights.

The Novel scene split distinguishes these responses further. SBG and EG continue to improve up to $\lambda=1$, whereas HGGD and RNGNet peak earlier, especially on Kinect. Both cameras show similar detector-level trends, with greater sensitivity for HGGD and RNGNet on the Kinect Novel scene split.

For configurations with an interior optimum, combining predicted quality and detector confidence yields higher AP than either score alone. The two scores therefore retain complementary information about candidate quality. The preferred balance depends on the detector: GN favors quality prediction, whereas RNGNet benefits from a larger contribution from detector confidence. Score Fusion is therefore needed to combine these complementary signals, with $\lambda$ controlling the strength of quality-based correction for each detector.

\subsection{Success and Failure Analysis}

\begin{figure}[t]
  \centering
  \includegraphics[width=\columnwidth]{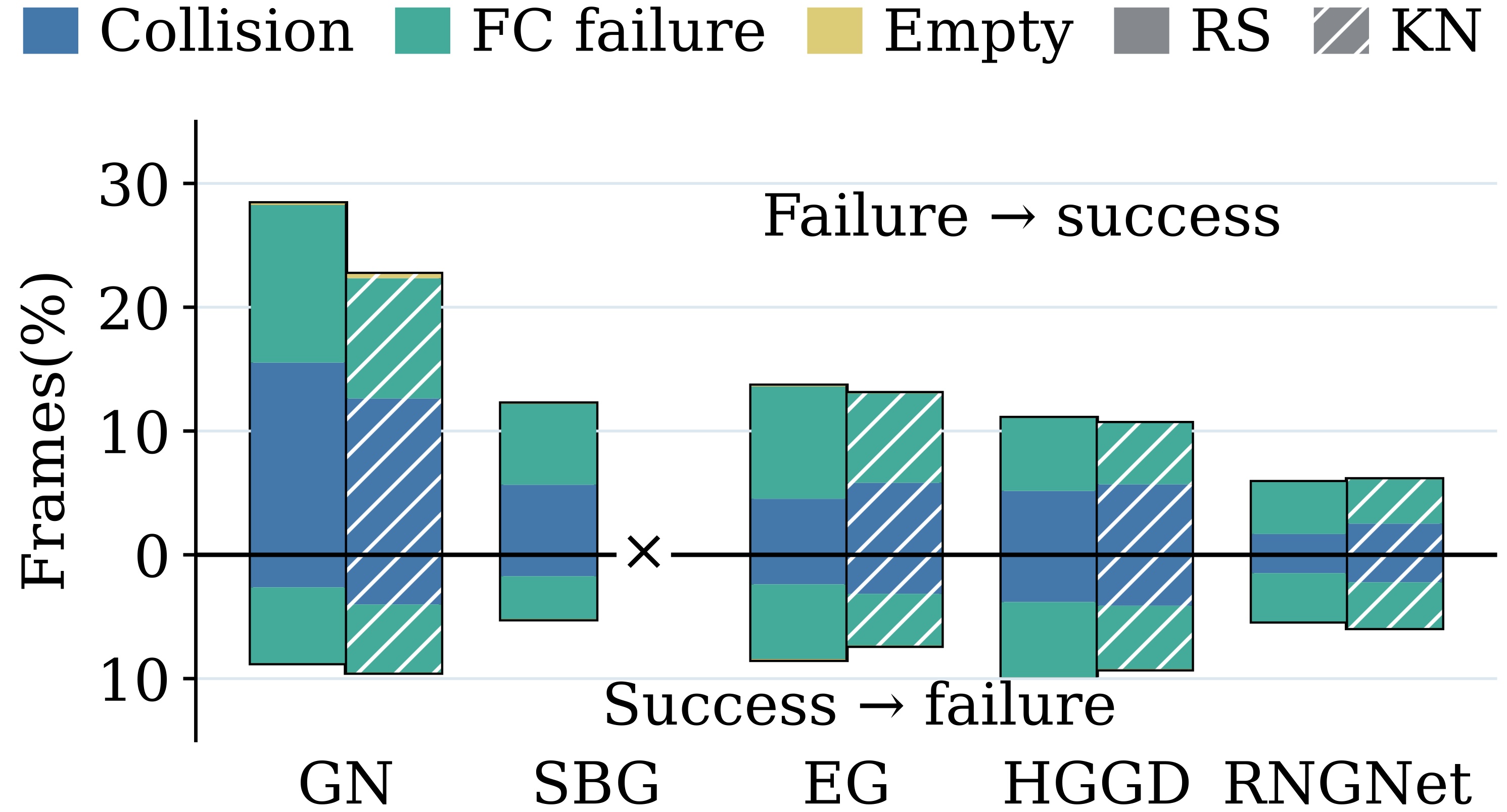}
  \caption{Top-1 outcome changes from $\pi^{\mathcal D}$ to $\pi^{\mathrm R}$. FC failures are non-empty, collision-free candidates that fail the success criterion.}
  \label{fig:error_transitions}
\end{figure}

Collisions and force-closure failures account for 47.86\% and 49.7\% of failure-to-success changes, respectively (Figure~\ref{fig:error_transitions}). GraRe thus improves both collision avoidance and selection among collision-free candidates.

\begin{figure}[t]
  \centering
  \includegraphics[width=\columnwidth]{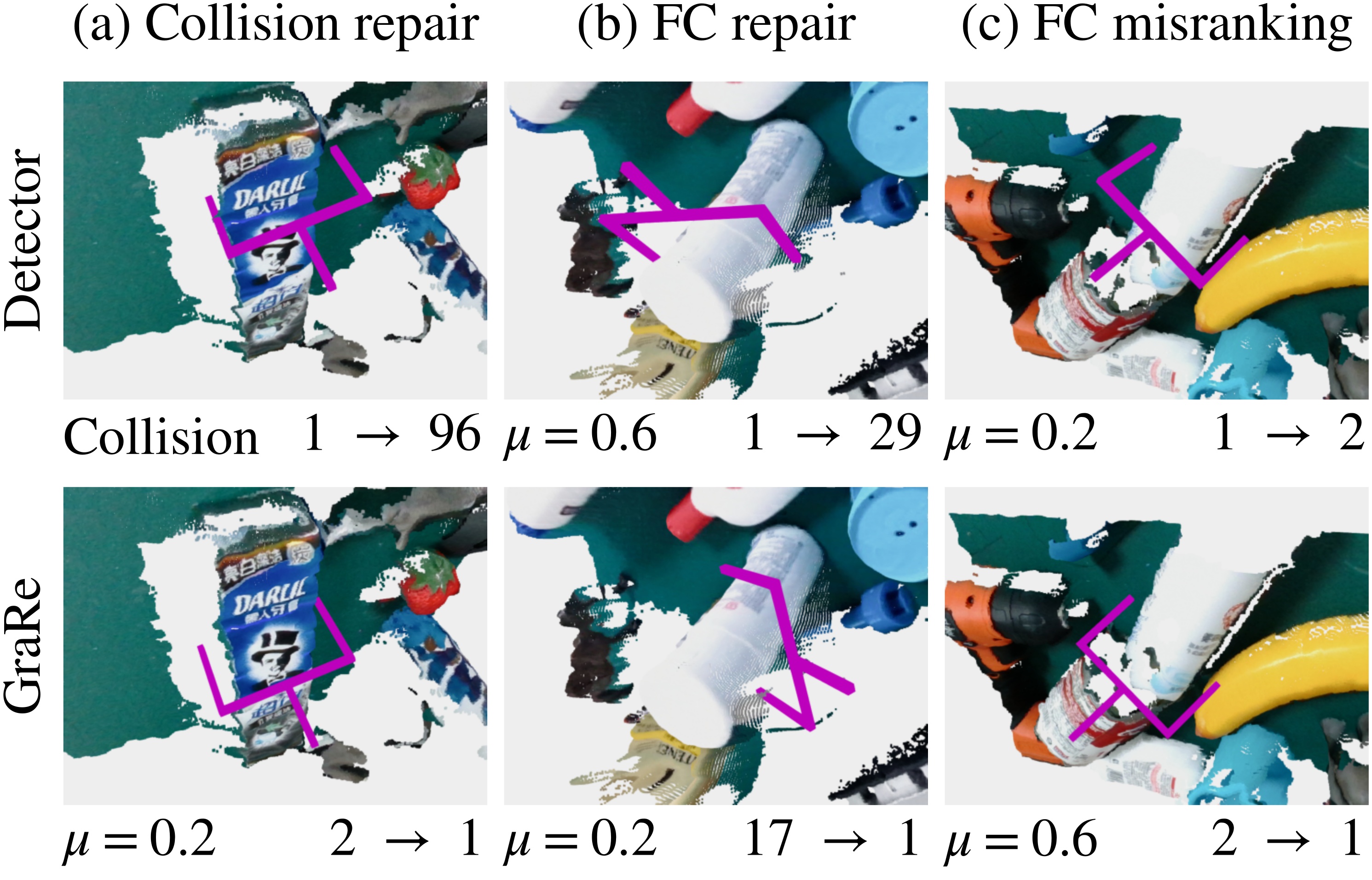}
  \caption{Top-ranked grasp candidates under $\pi^{\mathcal D}$ and $\pi^{\mathrm R}$. Columns (a)--(c) correspond to SBG--RS, GN--KN, and RNGNet--RS. Arrows show each candidate's rank change from $\pi^{\mathcal D}$ to $\pi^{\mathrm R}$.}
  \label{fig:qualitative_ranking}
\end{figure}

Figure~\ref{fig:qualitative_ranking}(a) illustrates collision avoidance with SBG. In (b), GraRe promotes a GN candidate from rank 17 to rank 1, replacing a candidate with $\mu_i=0.6$ with one requiring $\mu_i=0.2$. Both candidates are non-empty and collision-free on the same bottle. This contrast shows why object geometry must be interpreted with the candidate's grasp pose.

Force-closure failures account for 63.19\% of success-to-failure changes. In Figure~\ref{fig:qualitative_ranking}(c), the RNGNet candidates have similar gripper widths and closing directions but differ in position and approach direction. Both are non-empty and collision-free, yet their minimum friction coefficients are $\mu_i=0.2$ and $\mu_i=0.6$. Detector confidence correctly favors the successful candidate, but the margin is small. Predicted quality reverses this pose-dependent quality relation, and its opposing margin outweighs the detector's margin during Score Fusion, swapping the top two candidates.

GraRe estimates quality for each candidate independently, whereas selecting the leading candidate depends on their relative quality. This case shows how an error in that relative quality can overturn an otherwise correct first choice. Effective re-ranking therefore requires resolving the quality differences between specific grasp poses, including among non-empty, collision-free candidates.

\subsection{Real-Robot Experiments}

\begin{figure}[t]
  \centering
  \includegraphics[width=0.95\columnwidth]{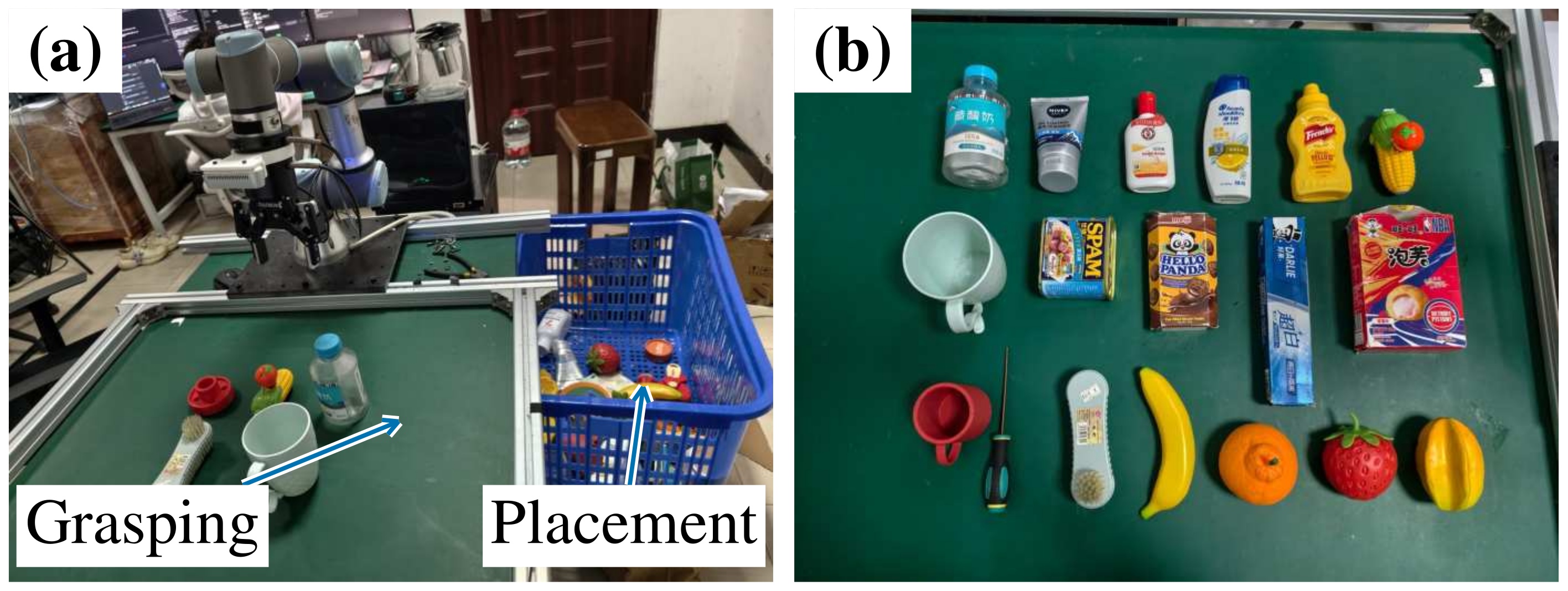}
  \caption{Real-robot setup and test objects.}
  \label{fig:robot_setup}
\end{figure}

\textbf{Setup.} We evaluate GN, SBG, and EG under $\pi^{\mathcal D}$ and $\pi^{\mathrm R}$ using a UR3, a RealSense D435, and a Robotiq 2F-85 gripper (Figure~\ref{fig:robot_setup}). Ten cluttered scenes contain objects from YCB, GraspNet-1Billion~\citep{calli2015ycb,fang2020graspnet}, and outside these datasets. The robot selects an executable candidate in the chosen order and observes the scene after each attempt, with up to 30 attempts per trial.

Grasp success rate (GSR) and completion rate (CR) measure successful attempts and fully cleared trials as percentages. Maximum consecutive failures (MCF) is the longest within-trial failure streak across the ten scenes. Outcomes are verified from recordings.

\textbf{Results.} GraRe raises GSR from 73.4\% to 88.9\% for GN, from 70.4\% to 87.2\% for SBG, and from 68.2\% to 91.0\% for EG (Table~\ref{tab:real_robot}). The number of cleared scenes increases from 1 to 10, 3 to 9, and 7 to 10, respectively, while the maximum failure streak falls from 4–6 attempts to 2. Better candidate selection thus improves both individual grasp success and sustained object removal over successive observations.

\begin{table}[t]
  \centering
  {\small
  \setlength{\tabcolsep}{3.5pt}
  \begin{tabular*}{\columnwidth}
    {@{\extracolsep{\fill}}lccc@{}}
    \toprule
    \textbf{Detector}
    & \textbf{GSR}
    & \textbf{CR}
    & \textbf{MCF} \\
    \midrule
    GN & 73.4/88.9 & 10/100 & 4/2 \\
    SBG & 70.4/87.2 & 30/90 & 6/2 \\
    EG & 68.2/91.0 & 70/100 & 4/2 \\
    \bottomrule
  \end{tabular*}
  }
  \caption{Real-robot results under $\pi^{\mathcal D}$/$\pi^{\mathrm R}$.}
  \label{tab:real_robot}
\end{table}

\begin{figure}[t]
  \centering
  \includegraphics[width=\columnwidth]{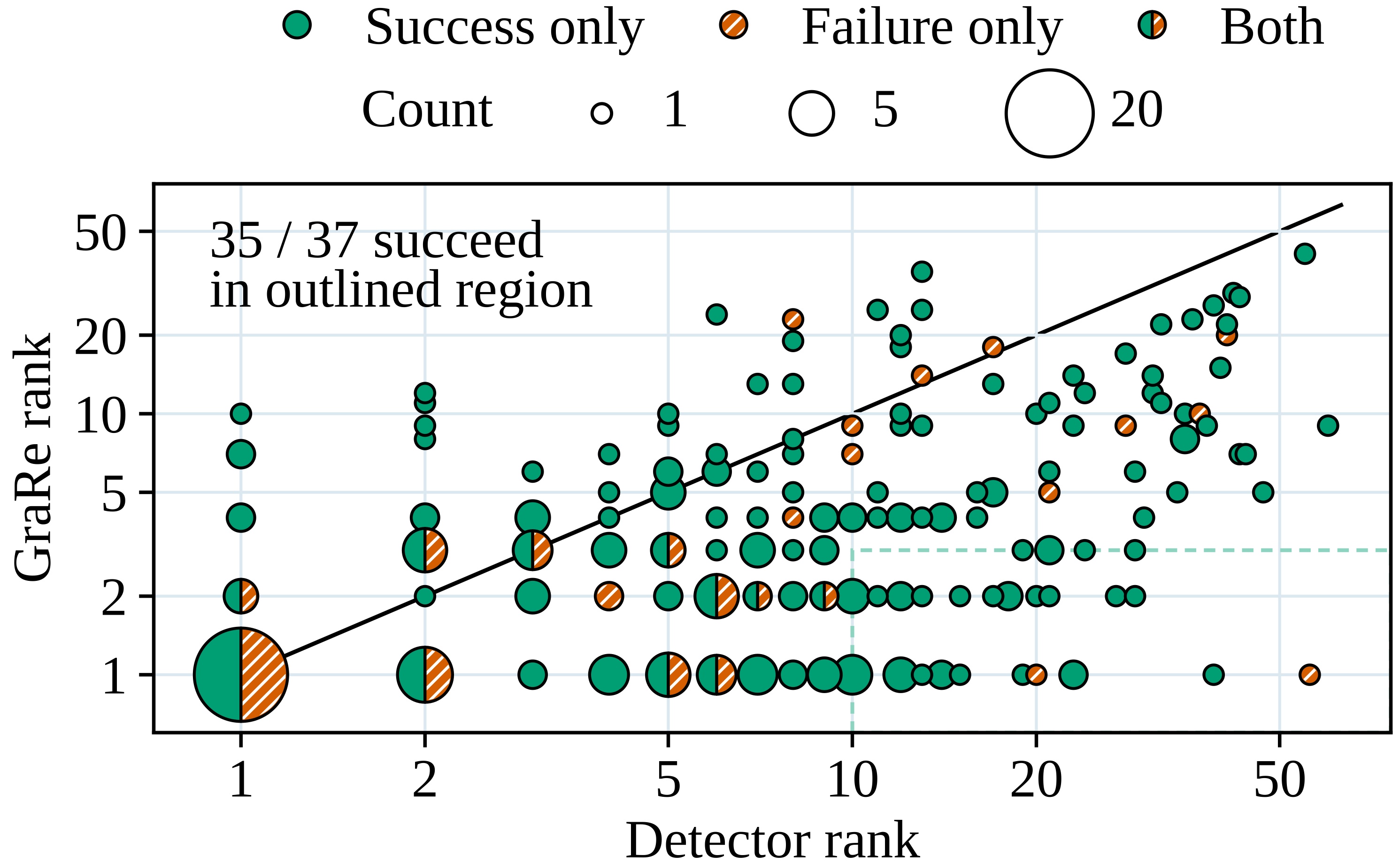}
  \caption{Ranks of 237 grasps executed under $\pi^{\mathrm R}$. Circle area encodes execution count; equal split colors indicate mixed outcomes. Dashed boundaries mark ranks $\geq10$ under $\pi^{\mathcal D}$ and $\leq3$ under $\pi^{\mathrm R}$.}
  \label{fig:real_robot_reranking}
\end{figure}

\begin{figure}[t]
  \centering
  \includegraphics[width=\columnwidth]{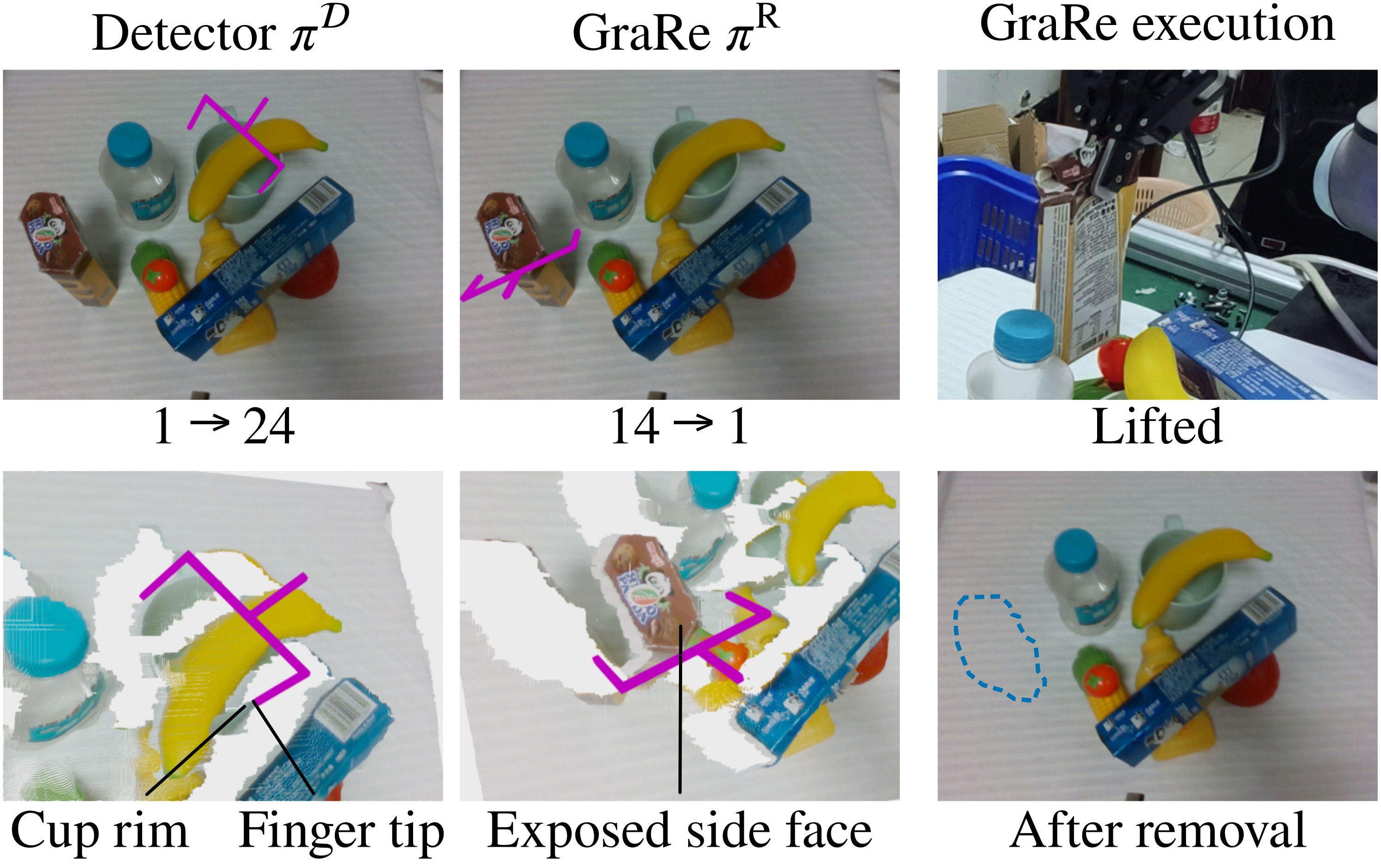}
  \caption{Grasp candidate selection and execution in an SBG scene. Arrows show rank changes from $\pi^{\mathcal D}$ to $\pi^{\mathrm R}$.}
  \label{fig:real_robot_selection}
\end{figure}

\textbf{Ranking and Execution.} Of the 37 executed candidates promoted from ranks $\geq10$ in $\pi^{\mathcal D}$ to the top three in $\pi^{\mathrm R}$, 35 succeed (Figure~\ref{fig:real_robot_reranking}). Their successful execution shows that detector confidence can undervalue physically effective grasp candidates. Figure~\ref{fig:real_robot_selection} shows the same issue across objects. The detector favors a banana candidate with a finger close to the cup rim. GraRe instead selects a carton candidate whose grasp spans exposed side faces with more gripper space, and the robot lifts and removes the carton. This contrast links candidate selection to possible finger contacts and free space for gripper motion.

\textbf{Inference Latency.} On an RTX 2060, Detector inference including instance segmentation takes 2.25, 2.33, and 2.03 s for GN, SBG, and EG, respectively. Given the Detector candidates, GraRe's object FPS, local-shell construction, and Point-MAE-based prediction and re-ranking add 0.84, 0.52, and 0.50 s for GN, SBG, and EG, respectively.

\section{Limitations}
Despite its improved grasping performance, GraRe has several limitations. GraRe assigns a fusion weight $\lambda$ to each detector, although the relative value of detector confidence and predicted quality varies across scene splits (Figure~\ref{fig:lambda_sensitivity}). An open question is which properties of the scene and candidate set determine this balance. Both geometric representations also depend on surfaces visible in a single observation. Additional views could reveal occluded contact regions and help distinguish candidates that appear similar from the current view.

\section{Conclusion}
Our analysis shows that existing 6-DoF grasp detectors can generate successful grasps but rank them too low. We address this problem with GraRe, which uses candidate features, local geometric features, and object-context features to re-rank existing grasp candidates while keeping detector parameters and grasp poses unchanged. GraRe improves the Average AP of five detectors on GraspNet-1Billion and increases grasp success and completion rates in real-robot experiments. These results show that existing detectors can be improved by refining candidate ranking without changing grasp generation.

\bibliography{references}

\clearpage
\appendix
\twocolumn[
  \begin{center}
    {\LARGE\bfseries Supplementary Material\par}
  \end{center}
  \vspace{0.5em}
]
\section{Experimental Details}
\label{sec:experimental_details}

\subsection{Dataset and Candidate Sets}

We use the official GraspNet-1Billion split, with 100 training scenes and 90 test scenes. The test set contains 30 scenes each for the Seen, Similar, and Novel splits, with 256 frames per scene. We reserve 10\% of the training scenes for validation, keeping all frames from each scene in the same partition.

The benchmark covers nine detector--camera settings: GN, EG, HGGD, and RNGNet with both RealSense and Kinect, and SBG with RealSense. Within each setting, Detector and GraRe use the same candidate set. GraRe changes only the ranking, preserving all candidate poses and gripper widths.

\subsection{Feature and Network Settings}

Candidate Features use a 14-dimensional vector containing the flattened rotation matrix, translation, gripper width, and detector confidence. Local Geometric Features use observed points expressed in the gripper coordinate frame. We apply farthest-point sampling separately within each radial shell. When a shell contains fewer points than its budget, the remaining entries are zero-padded and marked invalid for the local encoder.

For Object-Context Features, MobileSAM masks provide the visible-object depth points. Each object point cloud is centered and normalized to a unit sphere before Point-MAE grouping. The sampling and mask settings are listed in Table~\ref{tab:feature_settings}.

Candidate, local geometric, and object-context descriptors each have 128 dimensions. The frozen Point-MAE backbone produces a 768-dimensional descriptor by concatenating mean- and max-pooled patch features; a trainable adapter maps this descriptor to 128 dimensions. Feature Fusion operates on the three feature tokens of each candidate. Its output is projected to 64 dimensions and passed to a linear quality head. Network dimensions are summarized in Table~\ref{tab:network_settings}.

\subsection{Training Details}

\textbf{Supervision.} We use the continuous quality target defined in the main paper with $\tau=0.4$ and $\mu_{\mathrm{worst}}=2.0$. The quality head is trained with Smooth L1 loss. Collision and empty-grasp prediction use binary cross-entropy, and object classification uses cross-entropy over 88 classes. The corresponding auxiliary weights are 0.10, 0.05, and 0.05. Unavailable object-class labels are excluded from the classification loss.

\begin{table}[!ht]
\centering
{\small
\setlength{\tabcolsep}{3pt}
\begin{tabular*}{\columnwidth}{@{\extracolsep{\fill}}lc@{}}
\toprule
\textbf{Setting} & \textbf{Value} \\
\midrule
Shell boundaries (mm) & 0, 5, 15, 25, 40 \\
Points per shell & 64, 128, 128, 192 \\
Local point budget & 512 \\
Object point budget & 512 \\
Prompt cluster radius & 30 mm \\
Minimum mask area & 200 pixels \\
Maximum mask area & 40\% of the image \\
Point-MAE groups & 32 \\
Points per group & 32 \\
\bottomrule
\end{tabular*}
}
\caption{Feature-construction settings.}
\label{tab:feature_settings}
\end{table}

\textbf{Optimization.} We optimize the trainable parameters with AdamW and select the model by quality loss on held-out validation scenes. Table~\ref{tab:optimization_settings} summarizes the optimization settings.

\begin{table}[!ht]
\centering
{\small
\setlength{\tabcolsep}{3pt}
\begin{tabular*}{\columnwidth}{@{\extracolsep{\fill}}lc@{}}
\toprule
\textbf{Setting} & \textbf{Value} \\
\midrule
Candidate MLP dimensions & $14\!\to\!64\!\to\!128\!\to\!128$ \\
Local point MLP dimensions & $3\!\to\!64\!\to\!128\!\to\!128$ \\
Feature dimension & 128 \\
Shell Transformer layers / heads & 1 / 4 \\
Shell Transformer FFN dimension & 256 \\
Feature Fusion layers / heads & 1 / 4 \\
Feature Fusion FFN dimension & 256 \\
Fused feature dimension & 64 \\
Quality head & Linear, $64\!\to\!1$ \\
Dropout & 0.1 \\
\bottomrule
\end{tabular*}
}
\caption{Network settings of the full GraRe model.}
\label{tab:network_settings}
\end{table}

\begin{table}[!ht]
\centering
{\small
\setlength{\tabcolsep}{3pt}
\begin{tabular*}{\columnwidth}{@{\extracolsep{\fill}}lcc@{}}
\toprule
\textbf{Setting} & \textbf{RealSense} & \textbf{Kinect} \\
\midrule
Optimizer & AdamW & --- \\
Initial learning rate & $2\times10^{-4}$ & --- \\
Weight decay & $1\times10^{-4}$ & --- \\
Batch size (candidates) & 2,048 & --- \\
Gradient clipping norm & 1.0 & --- \\
Minimum learning rate & $10^{-5}$ & $10^{-6}$ \\
\bottomrule
\end{tabular*}
}
\caption{Optimization settings. Kinect-specific values apply to GN, HGGD, and RNGNet; dashes indicate the corresponding RealSense value.}
\label{tab:optimization_settings}
\end{table}

\subsection{Evaluation Protocol}

We use the official GraspNet evaluation protocol with friction thresholds of 0.2, 0.4, 0.6, 0.8, 1.0, and 1.2 and the top-50 grasp evaluation. Results are reported separately for Seen, Similar, and Novel scenes, together with Average AP. DS@$K$ and FS@$K$ follow the definitions in the main paper.

For Score Fusion, detector confidence and predicted quality are standardized separately within each frame. Candidates are then sorted by the fused score defined in the main paper. We refer to this final fused ranking as the GraRe order. The fusion weight is fixed to 1.0 for GN, 0.8 for SBG, EG, and HGGD, and 0.5 for RNGNet, with the same weight used across available cameras for each detector.

\section{Additional Ranking Results}
\label{sec:additional_ranking_results}

\textbf{Metric definitions.} For each detector--camera setting, let $F$ be the total number of evaluated frames and $N_f$ the number of candidates in frame $f$. Let $a_{fi}\in\{0,1\}$ indicate whether candidate $i$ in frame $f$ is non-empty, collision-free, and has a finite minimum friction coefficient $\mu_{fi}\leq0.4$ required for force closure. For any of the orders $\pi_f^{\mathcal D}$, $\pi_f^{\mathrm R}$, and $\pi_f^\star$, write $\pi_f(r)$ for the candidate index at rank $r$. For $K\geq1$, define $L_f(K)=\min(K,N_f)$ and $d_r=1/\log_2(r+1)$.

Frame Success@K measures the fraction of frames containing at least one successful candidate in the first $K$ positions:
\begin{equation}
\mathrm{FS}@K = \frac{100}{F}\sum_{f=1}^{F}
\mathbf{1}\!\left[\sum_{r=1}^{L_f(K)}a_{f,\pi_f(r)}>0\right].
\label{eq:supp_fs_at_k}
\end{equation}

Discounted Success@K measures success within each prefix with greater weight assigned to earlier ranks:
\begin{equation}
\mathrm{DS}@K = \frac{100}{F}
\sum_{\substack{f=1\\N_f>0}}^{F}
\frac{\sum_{r=1}^{L_f(K)}d_r a_{f,\pi_f(r)}}
{\sum_{r=1}^{L_f(K)}d_r}.
\label{eq:supp_ds_at_k}
\end{equation}

Here, $\mathbf{1}[\cdot]$ is the indicator function. Both metrics are reported as percentages and averaged equally over all frames. Frames without candidates contribute zero.

\begin{figure}[!ht]
\centering
\includegraphics[width=\columnwidth]{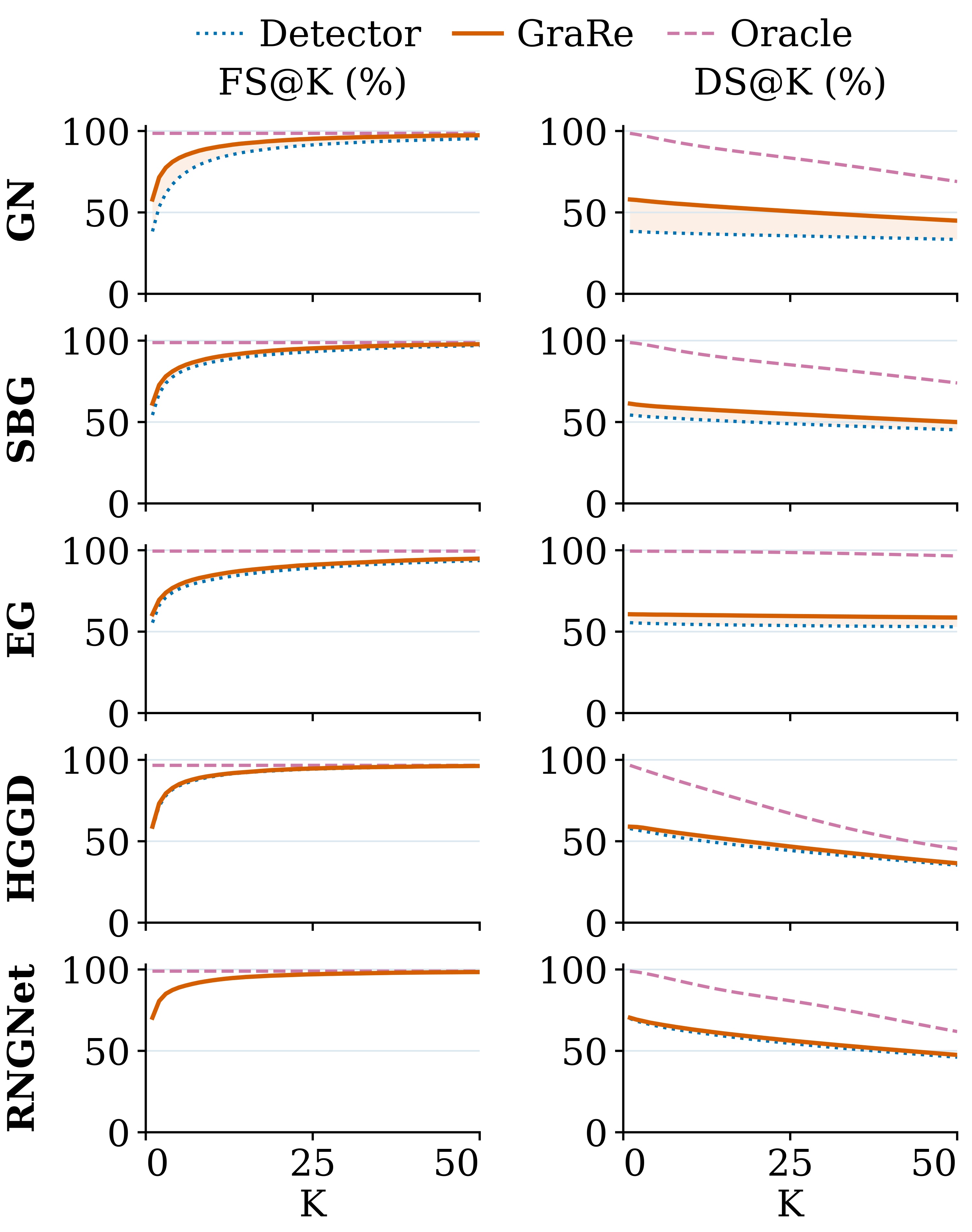}
\caption{FS@$K$ (left) and DS@$K$ (right) for RealSense. Shading marks the Detector--GraRe gap.}
\label{fig:additional_ranking_realsense}
\end{figure}

\begin{figure}[!ht]
\centering
\includegraphics[width=\columnwidth]{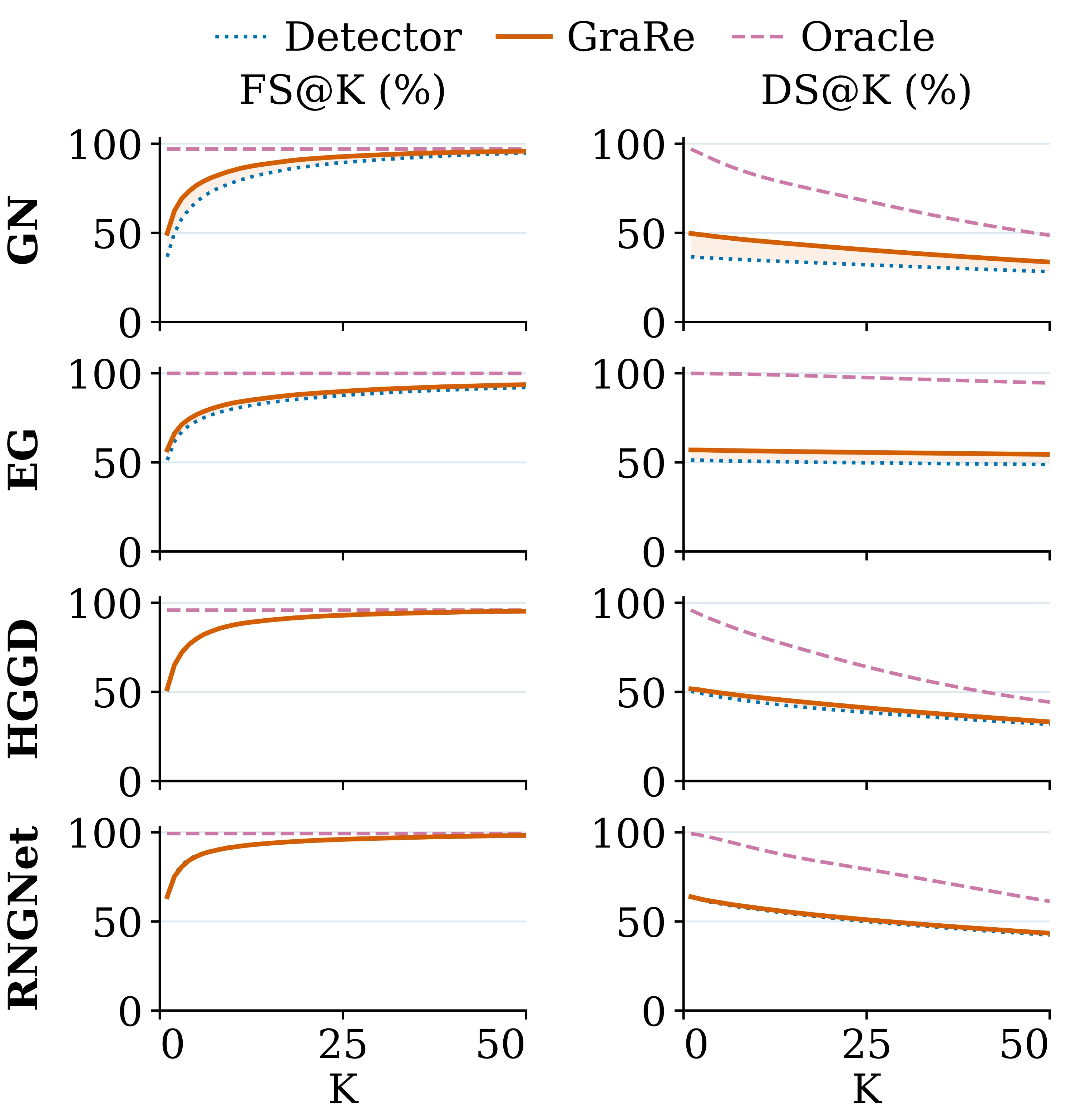}
\caption{Per-detector ranking curves on Kinect, using the same styles as Figure~\ref{fig:additional_ranking_realsense}.}
\label{fig:additional_ranking_kinect}
\end{figure}

Figures~\ref{fig:additional_ranking_realsense} and~\ref{fig:additional_ranking_kinect} compare ranking quality across all nine available detector--camera settings. Each curve averages 23,040 test frames for $K=1,\ldots,50$. The oracle orders successful candidates first within the same candidate set.

\textbf{Earlier access to successful candidates.} GraRe improves FS@1 in all nine settings. GN shows the largest gains, from 38.32\% to 57.96\% on RealSense and from 36.56\% to 49.73\% on Kinect. EG gains 5.17 and 5.71 percentage points, respectively.

\textbf{Ranking quality beyond the first success.} GraRe improves DS@$K$ at every $K=1,\ldots,50$ in all nine settings, with DS@50 gains of 1.02--11.55 percentage points. At $K=50$, FS ranges from 93.62\% to 98.41\%.

\textbf{Differences between FS and DS.} Higher DS does not guarantee higher FS at every ranking prefix. For RNGNet on Kinect at $K=5$, DS rises from 59.95\% to 60.59\%, while FS falls from 87.11\% to 86.48\%.

\textbf{Candidate availability.} Oracle FS@1 ranges from 95.88\% to 99.93\%, indicating that most frames contain a successful candidate. Differences in DS across detectors also reflect candidate availability; comparisons within each detector--camera setting reveal the effect of ordering.

\subsection{Successful-Candidate Availability}

\begin{figure}[!ht]
\centering
\includegraphics[width=\columnwidth]{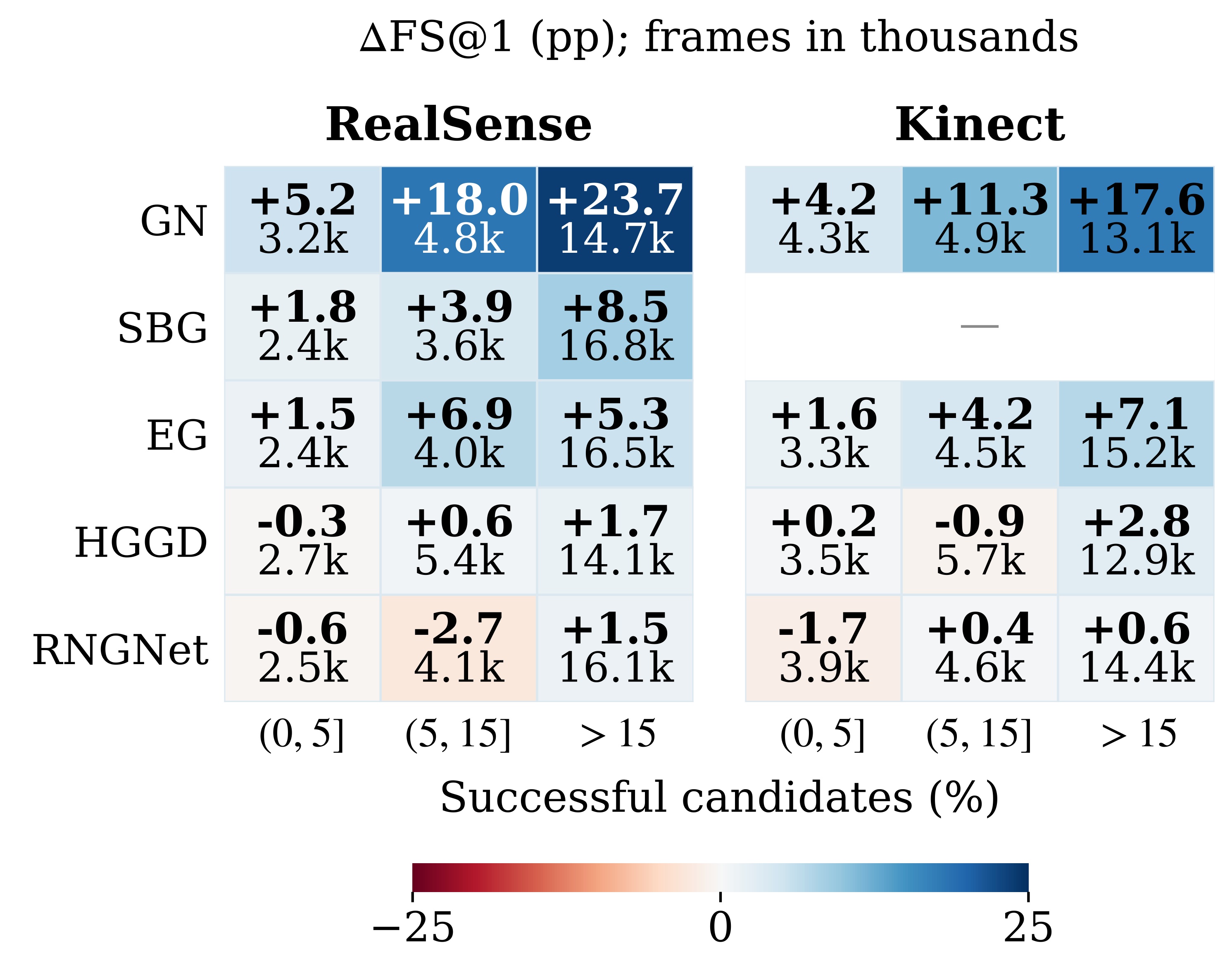}
\caption{Change in FS@1 across successful-candidate proportions. Each cell gives the gain in percentage points and the number of frames in thousands.}
\label{fig:ranking_candidate_fraction}
\end{figure}

We group frames by the proportion of successful candidates,
\begin{equation}
\rho_f=\frac{1}{N_f}\sum_{i=1}^{N_f}a_{fi},\qquad N_f>0,
\label{eq:supp_successful_candidate_fraction}
\end{equation}
using the success indicators defined above. We use four groups: $\rho_f=0$, $0<\rho_f\leq0.05$, $0.05<\rho_f\leq0.15$, and $\rho_f>0.15$. FS@1 is computed within each group. Figure~\ref{fig:ranking_candidate_fraction} shows the three positive groups; both orders necessarily fail when $\rho_f=0$. All evaluated frames have non-empty candidate sets.

Scarce successful candidates are harder to select. For GN on RealSense, FS@1 rises from 3.74\% to 8.91\% when $0<\rho_f\leq0.05$, compared with 53.49\% to 77.19\% when $\rho_f>0.15$.

Across the nine settings, 91.46--99.83\% of the remaining GraRe top-1 failures occur in frames that contain a successful candidate.

\subsection{Original Ranks of Selected Candidates}

\begin{figure}[!ht]
\centering
\includegraphics[width=\columnwidth]{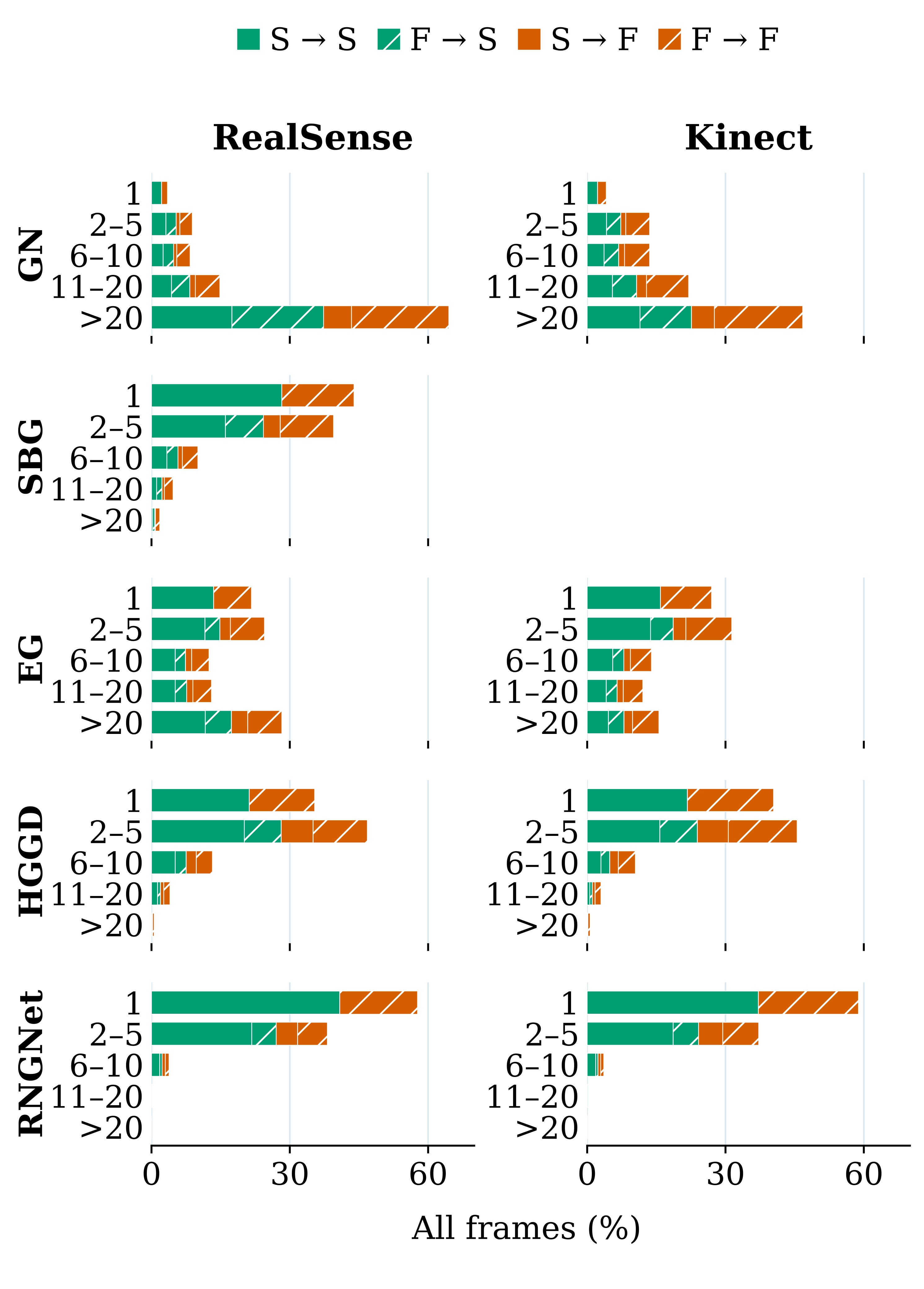}
\caption{Original Detector ranks of GraRe's top-1 candidates. Stacked segments show outcome changes as percentages of all 23,040 frames per setting; S and F denote success and failure.}
\label{fig:ranking_origin_rank}
\end{figure}

We trace each GraRe top-1 candidate to its rank in the Detector order and group these ranks into 1, 2--5, 6--10, 11--20, and greater than 20. Figure~\ref{fig:ranking_origin_rank} separates the four possible Detector-to-GraRe outcome changes, retaining both successful and unsuccessful selections.

GN selects candidates originally ranked below the top 20 in 64.51\% of RealSense frames and 46.75\% of Kinect frames. By comparison, more than 95\% of RNGNet's choices on either camera come from the original top five. GraRe therefore adjusts both deeper candidate rankings and choices near the front.

\section{Additional Benchmark Results}
\label{sec:additional_benchmark_results}

\textbf{AP at a friction threshold.} We follow the official GraspNet evaluation, including candidate selection and collision checks. Let $M_f$ be the number of candidates retained by the evaluator in frame $f$, ordered by the score of the evaluated method. Let $v_{fr}(\mu)$ indicate whether the candidate at rank $r$ is non-empty, collision-free, and achieves force closure at friction threshold $\mu$. Its top-$K$ precision is
\begin{equation}
P_f(K,\mu)=\frac{1}{K}\sum_{r=1}^{\min(K,M_f)}v_{fr}(\mu).
\label{eq:supp_precision_mu}
\end{equation}
For the frame set $\mathcal F_S$ of split $S$, we compute
\begin{equation}
\mathrm{AP}_S@\mu=\frac{100}{50|\mathcal F_S|}
\sum_{f\in\mathcal F_S}\sum_{K=1}^{50}P_f(K,\mu).
\label{eq:supp_ap_mu}
\end{equation}
Missing candidate positions contribute zero. The reported benchmark AP averages over the six friction thresholds $\mathcal M=\{0.2,0.4,0.6,0.8,1.0,1.2\}$:
\begin{equation}
\mathrm{AP}_S=\frac{1}{6}\sum_{\mu\in\mathcal M}\mathrm{AP}_S@\mu.
\label{eq:supp_ap_average_mu}
\end{equation}
The Average curve pools all test frames. Since Seen, Similar, and Novel contain equal numbers of frames, it is also the arithmetic mean of their three AP@$\mu$ curves.

\begin{figure}[!ht]
\centering
\includegraphics[width=\columnwidth]{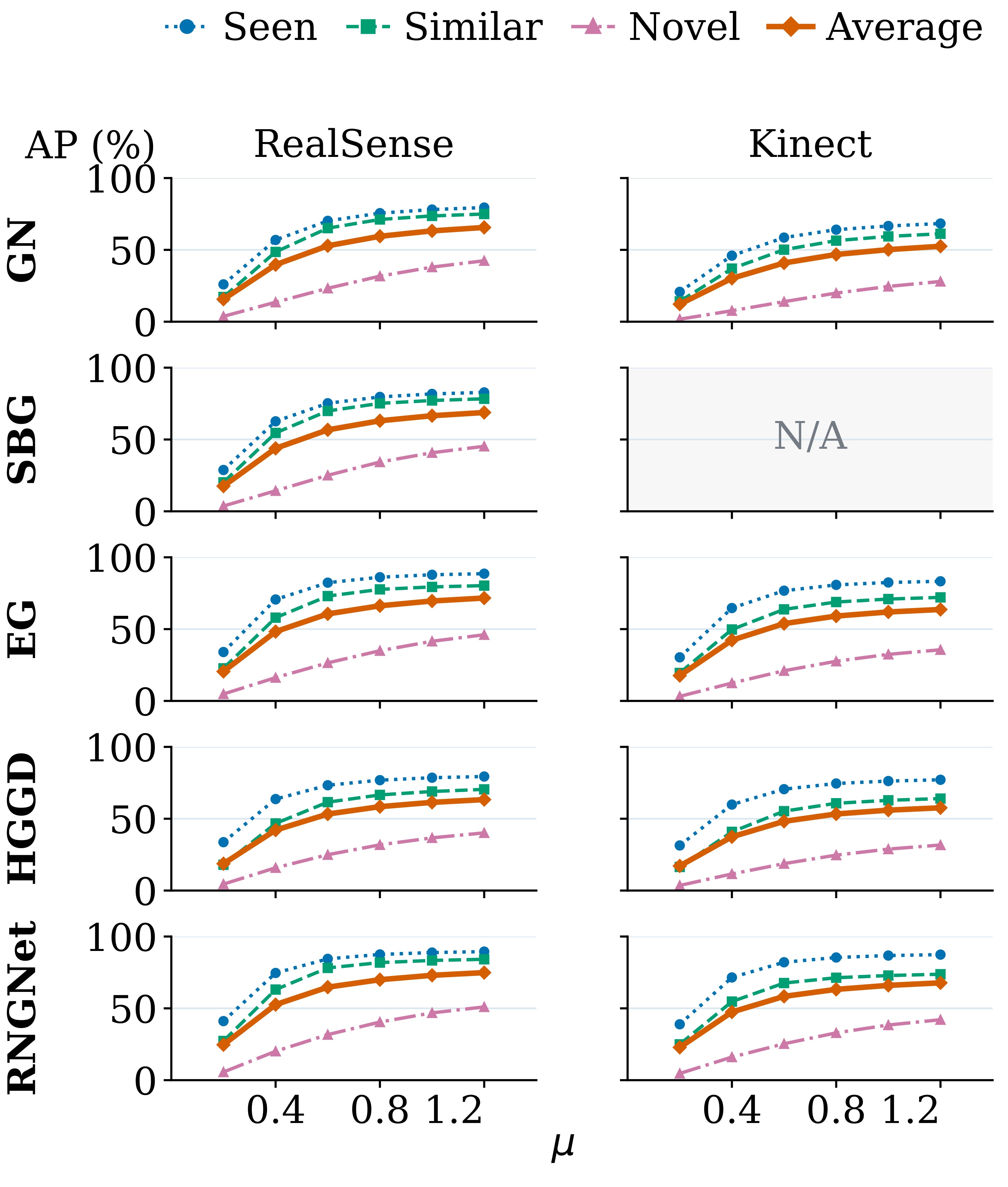}
\caption{GraRe AP across friction thresholds for Seen, Similar, Novel, and their average. SBG--Kinect is unavailable.}
\label{fig:additional_benchmark_ap_mu}
\end{figure}

\textbf{Performance across friction thresholds.} Figure~\ref{fig:additional_benchmark_ap_mu} compares GraRe across force-closure requirements and scene splits.

\textbf{Variation across training runs.} Table~\ref{tab:benchmark_variation} summarizes Average AP over three training runs. The standard deviations range from 0.06 to 0.22 percentage points.

\begin{table}[!ht]
\centering
{\small
\setlength{\tabcolsep}{3pt}
\begin{tabular*}{\columnwidth}{@{\extracolsep{\fill}}lcc@{}}
\toprule
\textbf{Detector} & \textbf{RealSense} & \textbf{Kinect} \\
\midrule
GN & $49.35 \pm 0.06$ & $38.66 \pm 0.16$ \\
SBG & $52.69 \pm 0.10$ & --- \\
EG & $56.05 \pm 0.08$ & $49.64 \pm 0.10$ \\
HGGD & $49.42 \pm 0.13$ & $44.75 \pm 0.20$ \\
RNGNet & $59.82 \pm 0.22$ & $54.15 \pm 0.11$ \\
\bottomrule
\end{tabular*}
}
\caption{Average AP (\%) over three training runs, reported as mean $\pm$ sample standard deviation. SBG--Kinect is unavailable.}
\label{tab:benchmark_variation}
\end{table}

\textbf{Uncertainty across test scenes.} We estimate 95\% confidence intervals for Average AP gains using a paired scene bootstrap, stratified by scene split, with 10,000 resamples (Table~\ref{tab:benchmark_scene_ci}). These intervals quantify uncertainty from test-scene composition.

\begin{table}[!ht]
\centering
{\small
\setlength{\tabcolsep}{3pt}
\begin{tabular*}{\columnwidth}{@{\extracolsep{\fill}}lccc@{}}
\toprule
\textbf{Detector} & \textbf{$\Delta$AP} & \textbf{95\% CI} & \textbf{Improved scenes} \\
\midrule
\multicolumn{4}{c}{\textbf{Kinect}} \\
GN & 8.21 & $[7.48,8.96]$ & 89/90 \\
EG & 4.49 & $[3.97,5.02]$ & 89/90 \\
HGGD & 2.81 & $[2.45,3.17]$ & 77/90 \\
RNGNet & 1.21 & $[0.98,1.45]$ & 75/90 \\
\midrule
\multicolumn{4}{c}{\textbf{RealSense}} \\
GN & 13.57 & $[12.65,14.52]$ & 90/90 \\
SBG & 5.09 & $[4.63,5.57]$ & 90/90 \\
EG & 4.12 & $[3.67,4.56]$ & 88/90 \\
HGGD & 3.04 & $[2.72,3.37]$ & 87/90 \\
RNGNet & 1.88 & $[1.64,2.13]$ & 83/90 \\
\bottomrule
\end{tabular*}
}
\caption{GraRe gains in Average AP and 95\% scene-bootstrap confidence intervals, in percentage points. The final column counts scenes with a positive AP gain.}
\label{tab:benchmark_scene_ci}
\end{table}
\textbf{Coverage of scene-level gains.} GraRe improves Average AP in all nine detector--camera settings, with every 95\% confidence interval above zero. AP improves in 75--90 of the 90 test scenes per setting, indicating that the gains extend across scenes.

\textbf{Comparison with state-of-the-art methods.} Table~\ref{tab:benchmark_sota} compares single-view grasp detection on GraspNet-1Billion. We report the five frozen detectors with and without GraRe alongside published results for other methods~\citep{wang2021graspness,wang2024erfgbs,graphgrasp2026,iwase2025zerograsp,cai2025r2sgrasp}.

\begin{table*}[!t]
\centering
\captionsetup{skip=5pt}
\caption{AP (\%) on GraspNet-1Billion, reported separately for RealSense and Kinect. CD denotes collision detection; $\dagger$ marks additional grasp-training data. Dashes indicate unreported or unavailable results.}
\label{tab:benchmark_sota}
{\small
\setlength{\tabcolsep}{3pt}
\renewcommand{\arraystretch}{1.06}
\begin{tabular*}{\textwidth}{@{\extracolsep{\fill}}l*{8}{r}@{}}
\toprule
 & \multicolumn{4}{c}{\textbf{RealSense}} & \multicolumn{4}{c}{\textbf{Kinect}} \\
\cmidrule(lr){2-5}\cmidrule(lr){6-9}
\textbf{Method} & \textbf{Seen} & \textbf{Similar} & \textbf{Novel} & \textbf{Average} & \textbf{Seen} & \textbf{Similar} & \textbf{Novel} & \textbf{Average} \\
\midrule
GN & 47.83 & 42.79 & 16.94 & 35.86 & 41.98 & 37.57 & 12.24 & 30.60 \\
$\mathrm{GraRe}_{\mathrm{GN}}$ & 64.36 & 58.44 & 25.45 & 49.42 & 54.06 & 46.38 & 15.97 & 38.80 \\
GSNet+CD & 67.12 & 54.81 & 24.31 & 48.75 & 63.50 & 49.18 & 19.78 & 44.15 \\
SBG & 62.27 & 56.92 & 23.79 & 47.66 & --- & --- & --- & --- \\
$\mathrm{GraRe}_{\mathrm{SBG}}$ & 68.47 & 62.55 & 27.24 & 52.75 & --- & --- & --- & --- \\
EG & 69.31 & 61.49 & 25.28 & 52.03 & 63.76 & 52.42 & 19.61 & 45.26 \\
$\mathrm{GraRe}_{\mathrm{EG}}$ & 74.92 & 65.16 & 28.35 & 56.14 & 69.72 & 57.47 & 22.06 & 49.75 \\
HGGD & 63.37 & 52.28 & 23.83 & 46.49 & 60.43 & 46.74 & 19.23 & 42.13 \\
$\mathrm{GraRe}_{\mathrm{HGGD}}$ & 67.58 & 55.37 & 25.66 & 49.53 & 64.95 & 50.03 & 19.85 & 44.94 \\
Wang et al.+CD & 74.33 & 64.36 & 27.56 & 55.42 & --- & --- & --- & --- \\
R2SGrasp$^\dagger$ & 64.71 & 56.93 & 26.31 & 49.32 & 66.12 & 54.47 & 24.74 & 48.44 \\
ZeroGrasp & 70.53 & 62.51 & 26.46 & 53.17 & --- & --- & --- & --- \\
ZeroGrasp+FT$^\dagger$ & 72.43 & 65.45 & 28.49 & 55.46 & --- & --- & --- & --- \\
GraphGrasp & 64.88 & 56.91 & 24.83 & 48.87 & --- & --- & --- & --- \\
RNGNet & 75.32 & 66.95 & 32.03 & 58.10 & 73.50 & 59.20 & 26.44 & 53.05 \\
$\mathrm{GraRe}_{\mathrm{RNG}}$ & \textbf{77.67} & \textbf{69.64} & \textbf{32.63} & \textbf{59.98} & \textbf{75.34} & \textbf{60.86} & \textbf{26.58} & \textbf{54.26} \\
\bottomrule
\end{tabular*}
}
\end{table*}

GraRe improves Average AP in all nine detector--camera settings. With RNGNet, it achieves the highest Average AP among the compared methods on both cameras and improves all three scene splits. These results show that refining candidate order can improve grasp detection while leaving the candidate set unchanged.

\section{Additional Ablation Results}
\label{sec:additional_ablation_results}

Table~\ref{tab:additional_ablation}(a--d) examines feature contributions, local geometry, auxiliary supervision, and quality targets. Average AP (\%) is reported separately for RealSense and Kinect. Dashes indicate unreported results; SBG--Kinect is unavailable.

\setlength{\floatsep}{4pt}
\setlength{\textfloatsep}{6pt}
\captionsetup[table]{skip=2pt}

\begin{table*}[!t]
\centering
\captionsetup{skip=5pt}
\caption{Additional ablations of GraRe, reporting Average AP (\%) separately for RealSense and Kinect. Detector and Full GraRe are shared references for all four groups. Dashes indicate unreported results; SBG--Kinect is unavailable.}
\label{tab:additional_ablation}
\label{tab:additional_ablation_feature}
\label{tab:additional_ablation_geometry}
\label{tab:additional_ablation_auxiliary}
\label{tab:additional_ablation_objective}
{\small
\setlength{\tabcolsep}{2.5pt}
\renewcommand{\arraystretch}{1.06}
\begin{tabular*}{\textwidth}{@{\extracolsep{\fill}}l*{10}{r}@{}}
\toprule
 & \multicolumn{5}{c}{\textbf{RealSense}} & \multicolumn{5}{c}{\textbf{Kinect}} \\
\cmidrule(lr){2-6}\cmidrule(lr){7-11}
\textbf{Variant} & \textbf{GN} & \textbf{SBG} & \textbf{EG} & \textbf{HGGD} & \textbf{RNGNet} & \textbf{GN} & \textbf{SBG} & \textbf{EG} & \textbf{HGGD} & \textbf{RNGNet} \\
\midrule
Detector & 35.86 & 47.66 & 52.03 & 46.49 & 58.10 & 30.60 & - & 45.26 & 42.13 & 53.05 \\
Full GraRe & 49.42 & 52.75 & 56.14 & 49.53 & 59.98 & 38.80 & - & 49.75 & 44.94 & 54.26 \\
\midrule
\multicolumn{11}{l}{\textbf{(a) Feature inputs}} \\
\addlinespace[1pt]
w/o Candidate & 47.63 & 51.74 & 54.19 & 47.99 & 59.98 & 37.21 & - & 48.15 & 43.73 & 53.79 \\
w/o Local & 44.99 & 50.60 & 54.45 & 47.77 & 58.76 & 36.48 & - & 48.43 & 43.15 & 53.60 \\
w/o Object-Context & 48.71 & 52.59 & 55.98 & 49.92 & 59.83 & 38.04 & - & 49.73 & 45.05 & 54.44 \\
Candidate-only & 42.59 & 49.88 & 53.99 & 47.09 & 58.68 & 34.08 & - & 47.67 & 42.39 & 53.53 \\
Local-only & 46.66 & 51.61 & 53.69 & 47.86 & 59.82 & 37.22 & - & 48.02 & 43.18 & 54.01 \\
Object-only & 34.04 & 44.18 & 43.70 & 39.24 & 54.55 & 28.29 & - & 37.82 & 36.91 & 48.84 \\
\addlinespace[4pt]
\multicolumn{11}{l}{\textbf{(b) Local geometry}} \\
\addlinespace[1pt]
w/o Shell-wise FPS & 48.27 & 52.64 & 55.96 & 47.90 & 58.69 & 38.12 & - & 49.71 & 43.33 & 53.67 \\
shuffled-local & 45.08 & 50.62 & 54.40 & 47.86 & 58.71 & 36.26 & - & 48.41 & 43.25 & 53.39 \\
\addlinespace[4pt]
\multicolumn{11}{l}{\textbf{(c) Auxiliary supervision}} \\
\addlinespace[1pt]
w/o collision/empty & 49.06 & 52.73 & 56.21 & 49.89 & 59.71 & 38.99 & - & 49.76 & 44.85 & 54.16 \\
w/o object-id & 48.91 & 52.60 & 55.77 & 49.46 & 59.80 & 38.49 & - & 49.57 & 44.60 & 54.01 \\
w/o all aux & 48.92 & 52.27 & 55.70 & 49.47 & 59.77 & 38.50 & - & 49.40 & 44.55 & 54.10 \\
\addlinespace[4pt]
\multicolumn{11}{l}{\textbf{(d) Training targets and ranking objectives}} \\
\addlinespace[1pt]
Binary $\tau=.2$ & 47.32 & 51.52 & 55.26 & 48.48 & 58.65 & 37.59 & - & 49.22 & 43.60 & 53.53 \\
Binary $\tau=.6$ & 48.92 & 52.45 & 55.70 & 49.44 & 59.60 & 38.64 & - & 49.68 & 44.76 & 53.96 \\
shuffled targets & 35.78 & 45.13 & 44.07 & 43.61 & 55.31 & 27.77 & - & 40.85 & 37.26 & 48.17 \\
RankNet & 49.43 & 52.99 & 56.59 & 49.60 & 59.69 & 38.69 & - & 49.90 & 44.92 & 54.19 \\
ListMLE & 48.74 & 52.48 & 56.36 & 49.11 & 59.86 & 38.16 & - & 50.10 & 45.01 & 54.37 \\
\bottomrule
\end{tabular*}
}
\end{table*}

\subsection{Feature Contributions}

Table~\ref{tab:additional_ablation}(a) compares feature removal with using each feature type alone. Full GraRe achieves higher Average AP than every reported single-feature variant, supporting the complementary roles of Candidate, Local Geometric, and Object-Context Features in quality prediction.

\subsection{Local Geometric Representation}

Table~\ref{tab:additional_ablation}(b) examines local geometric representations and candidate--geometry correspondence. Shuffling local point clouds among candidates reduces Average AP in every reported comparison. This result supports associating local geometry with the particular grasp pose being evaluated.

\subsection{Auxiliary Supervision}

Table~\ref{tab:additional_ablation}(c) separates collision/empty-grasp supervision from object classification. The \emph{w/o object-id} variant removes the object-classification loss. Removing all auxiliary losses reduces Average AP in every reported setting, supporting supervision that captures failure modes and object information alongside scalar grasp quality.

\subsection{Training Targets and Ranking Objectives}

Table~\ref{tab:additional_ablation}(d) compares the continuous quality target with two binary targets and the RankNet and ListMLE objectives. The continuous target outperforms both binary targets in every reported comparison, supporting the use of graded friction requirements rather than a binary success label. RankNet and ListMLE provide alternative ways to learn candidate ordering.

\section{Additional Sensitivity Results}
\label{sec:additional_sensitivity_results}

We examine how the Score Fusion weight affects individual test scenes and how sensitive the results are to small changes around the fixed weights. The models and their candidate sets remain unchanged throughout the analysis.

\textbf{Rationale for the fixed fusion weights.} We use empirical fusion weights shared across cameras for each detector. Their relative magnitudes are consistent with the observed differences in baseline performance and ranking headroom. GN has the lowest Detector Average AP (30.60--35.86) and the largest mean Oracle--Detector FS@1 gap (60.34 percentage points), whereas RNGNet has the highest Detector Average AP (53.05--58.10) and the smallest gap (32.34 points). SBG, EG, and HGGD occupy the intermediate range, with Detector Average AP of 42.13--52.03 and mean FS@1 gaps of 42.18--46.30 points. AP ranges span the evaluated cameras; each mean gap averages those cameras for one detector.

This pattern provides a qualitative rationale for the weighting scheme: quality-only ranking for GN ($\lambda=1.0$), a greater contribution from predicted quality for SBG, EG, and HGGD ($\lambda=0.8$), and equal weighting of the two standardized scores for RNGNet ($\lambda=0.5$). The exact weights remain empirical choices.

\subsection{Scene-level Fusion Sensitivity}

For each scene $s$, we report the change in scene-level Average AP relative to the detector order:
\begin{equation}
\Delta \mathrm{AP}_{s}(\lambda)
=\mathrm{AP}_{s}(\lambda)-\mathrm{AP}_{s}(0).
\label{eq:supp_scene_lambda_gain}
\end{equation}
The endpoints $\lambda=0$ and $\lambda=1$ rank candidates by Detector confidence and predicted quality, respectively. Each scene-level value averages its 256 test frames under the official protocol. We evaluate $\lambda\in\{0,0.1,\ldots,1.0\}$ while preserving the original scene order within the Seen, Similar, and Novel splits. The quality predictor uses the same input features throughout the sweep.

\begin{figure}[!ht]
\centering
\includegraphics[width=\columnwidth]{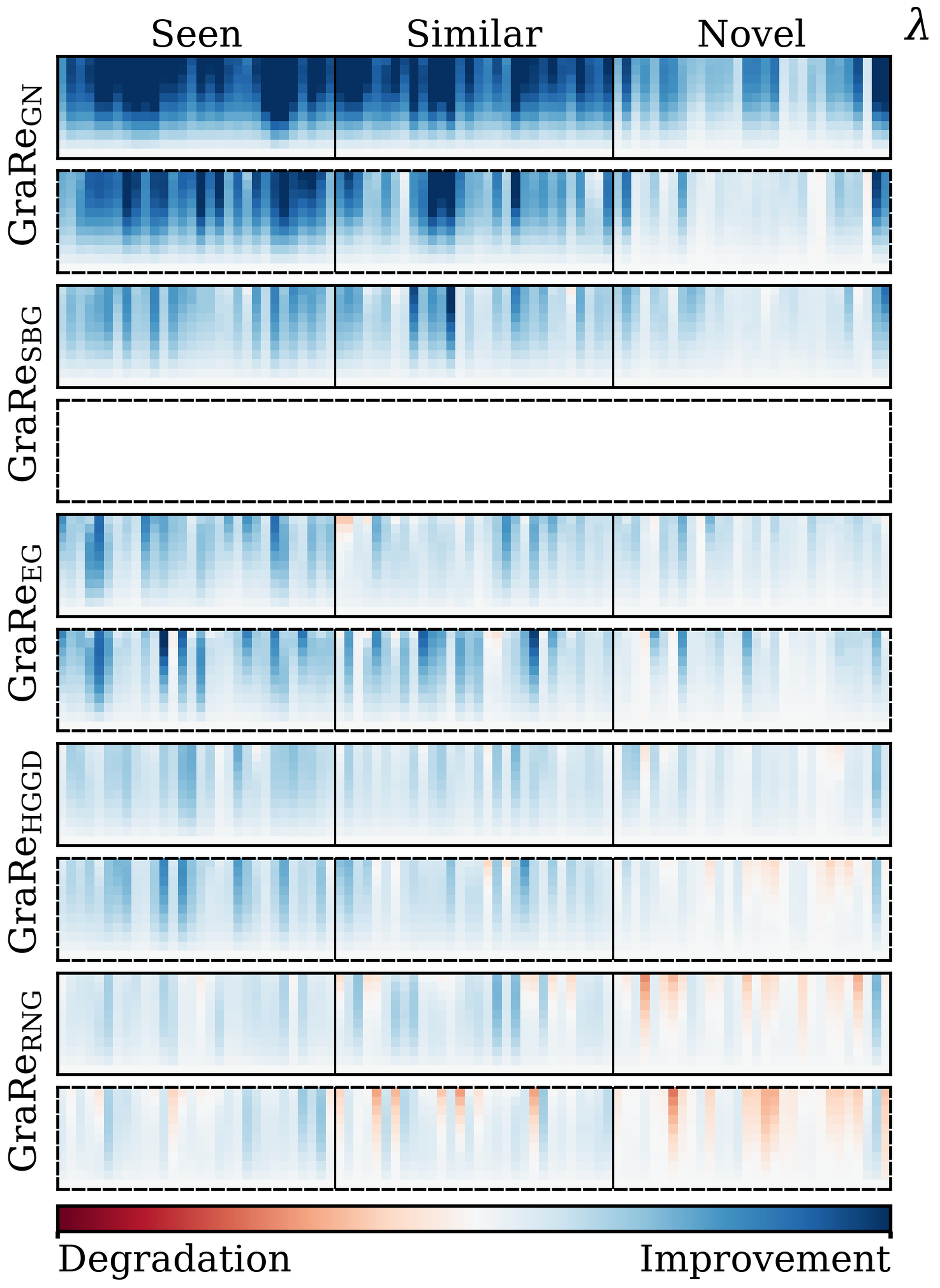}
\caption{Scene-level Average AP gains relative to the Detector order across Score Fusion weights. Each detector block contains a RealSense row (solid border) and a Kinect row (dashed border); columns retain the 90 test scenes in Seen, Similar, and Novel order. Rows run from $\lambda=0$ at the bottom to $\lambda=1$ at the top, and the color scale is shared across all rows.}
\label{fig:additional_sensitivity_lambda_sweep_scenewise}
\end{figure}

The combined heatmap shows that increasing the quality weight does not benefit every scene equally. For RNGNet on Kinect, increasing $\lambda$ from 0.5 to 1.0 reduces the Average AP gain from 1.09 to 0.01 points, while the number of improved scenes falls from 66 to 49 out of 90. The mean gain on Novel scenes changes from $-0.17$ to $-1.44$ points. Thus, ranking by predicted quality alone can improve the aggregate result only marginally while worsening the Novel split.

Similar aggregate values can also hide a change in which scenes benefit. For EG on Kinect, the Average AP gains at $\lambda=0.8$ and $\lambda=1.0$ are 4.35 and 4.33 points, respectively, but the number of improved scenes decreases from 89 to 82. The scene-level view therefore complements the split averages: detector confidence can remain useful even when the quality-only endpoint has a similar mean score.

\textbf{Top-1 outcome transitions.} Scene-level AP gains do not show which leading candidate changes outcome. We therefore trace the top-ranked candidate on each frame while varying the Score Fusion weight. For frame $f$, let $\mathcal C_f=\{c_{f,i}\}_{i=1}^{N_f}$ with $c_{f,i}=(p_{f,i},w_{f,i},b_{f,i})$, and let $\pi_f^{\mathcal D}$ and $\pi_f^{\mathrm R}(\lambda)$ denote the Detector and final GraRe orders. With $\tau=0.4$, the success indicator is
\begin{equation}
\begin{aligned}
a_{f,i}=\mathbf 1\!\big[&\operatorname{finite}(\mu_{f,i})\land y_{f,i}^{\mathrm{coll}}=0\\[-2pt]
&{}\land y_{f,i}^{\mathrm{empty}}=0\land \mu_{f,i}\le\tau\big].
\end{aligned}
\label{eq:supp_transition_success}
\end{equation}
Here, zero collision and empty-grasp labels denote a collision-free and non-empty candidate. The top-1 statuses are
\begin{equation}
\chi_f^{\mathcal D}=a_{f,\pi_f^{\mathcal D}(1)},\qquad
\chi_{f,\lambda}^{\mathrm R}=a_{f,\pi_f^{\mathrm R}(\lambda)(1)}.
\label{eq:supp_transition_status}
\end{equation}
For a split $S$ with frame set $\mathcal F_S$, we count
\begin{equation}
\begin{aligned}
n_{\mathrm{F\to S}}(S,\lambda)
&=\sum_{f\in\mathcal F_S}
\mathbf 1[\chi_f^{\mathcal D}=0\land\chi_{f,\lambda}^{\mathrm R}=1],\\
n_{\mathrm{S\to F}}(S,\lambda)
&=\sum_{f\in\mathcal F_S}
\mathbf 1[\chi_f^{\mathcal D}=1\land\chi_{f,\lambda}^{\mathrm R}=0].
\end{aligned}
\label{eq:supp_transition_counts}
\end{equation}
We report each count as $100n/|\mathcal F_S|$, so every frame remains in the denominator and contributes at most one transition; frames without candidates contribute zero to both counts. F$\to$S means that the Detector's top-1 candidate fails while GraRe's top-1 candidate succeeds, whereas S$\to$F is the reverse. The resulting net change is $\Delta\mathrm{FS}@1=R_{\mathrm{F\to S}}-R_{\mathrm{S\to F}}$.

\begin{figure}[!ht]
\centering
\includegraphics[width=\columnwidth]{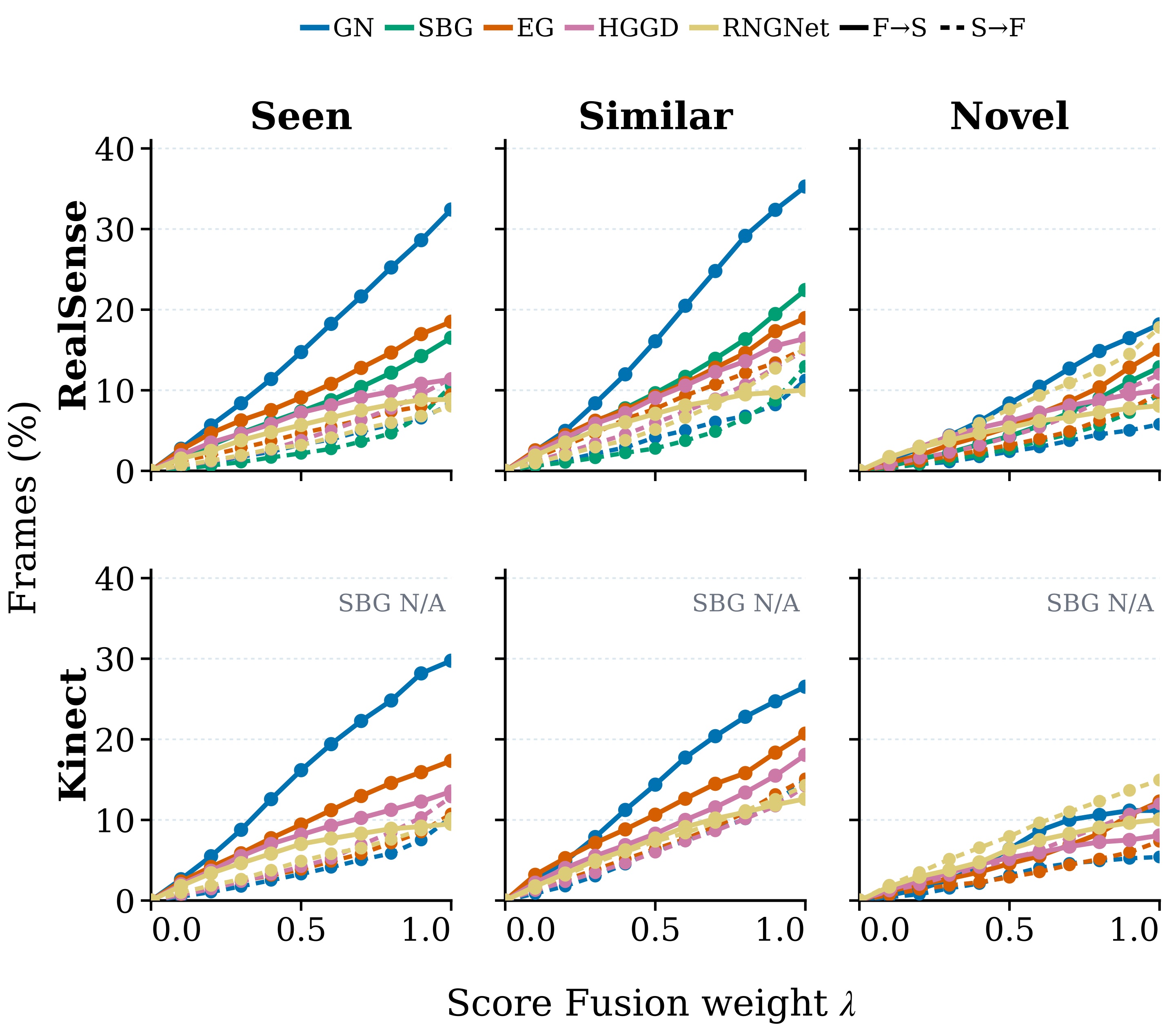}
\caption{Top-1 outcome transitions over Score Fusion weights. Solid and dashed curves show F$\to$S and S$\to$F, respectively; rates use all frames within each scene split.}
\label{fig:additional_sensitivity_lambda_transitions}
\end{figure}

We apply the Score Fusion rule at each weight $\lambda\in\{0,0.1,\ldots,1.0\}$ to the same candidates and record the resulting top-1 transitions. Figure~\ref{fig:additional_sensitivity_lambda_transitions} reports these rates for all nine detector--camera pairs.

At the fixed fusion weights, recoveries outweigh regressions in all nine detector--camera settings. GN shows the largest net FS@1 gains: 20.18 percentage points on RealSense and 12.27 on Kinect at $\lambda=1$. EG gains 4.63 and 5.16 points, respectively, at $\lambda=0.8$. The preferred balance varies across detectors and scene splits: SBG--RealSense peaks near $\lambda=0.9$, whereas larger quality weights introduce more regressions on Novel scenes for HGGD and RNGNet. These differences support balancing predicted quality with detector confidence.

\subsection{Score Preferences and Complementarity}

\begin{figure}[!ht]
\centering
\includegraphics[width=\columnwidth]{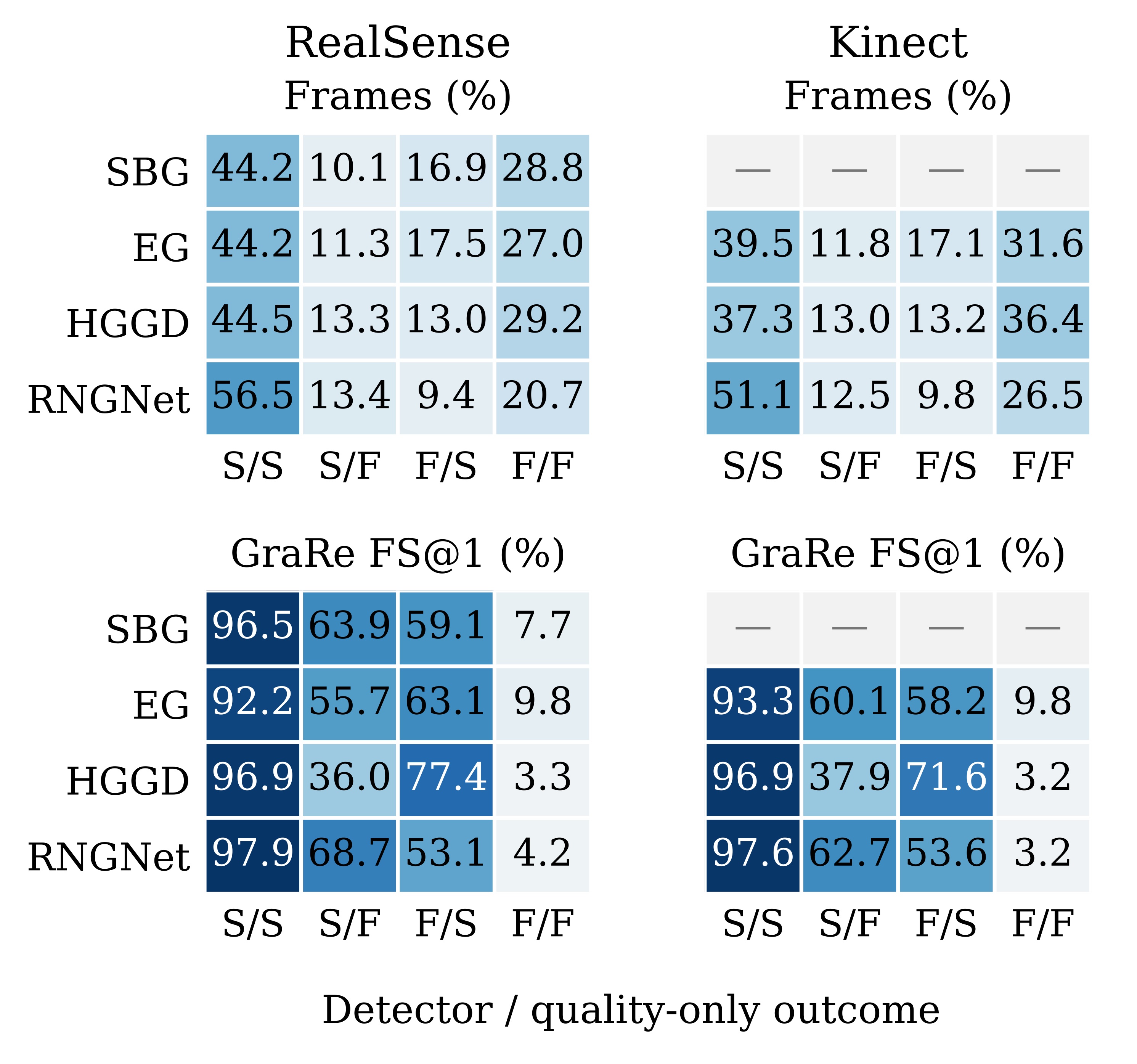}
\caption{Outcome groups for Detector / quality-only top choices. The upper grids show each group's share of all frames; the lower grids show GraRe FS@1 within that group. S and F denote success and failure.}
\label{fig:score_complementarity}
\end{figure}

We group frames by the outcomes of the top candidates selected by Detector confidence and by predicted quality alone ($\lambda=1$), then examine GraRe's final choice at the fixed fusion weight. The four groups contain frames where both choices succeed, only the Detector choice succeeds, only the quality-only choice succeeds, or both choices fail. For GN, the fixed weight is already one, so GraRe and quality-only ordering coincide. Figure~\ref{fig:score_complementarity} shows the seven detector--camera pairs with intermediate fusion weights.

When only the Detector choice succeeds, GraRe selects a successful candidate in 35.99--68.67\% of frames across these seven pairs. When only the quality-only choice succeeds, this fraction is 53.08--77.37\%. Neither score subsumes the other's successful choices, and fusion retains successful selections from both types of disagreement.

Fusion can also select candidates that are not ranked first by either endpoint. When both endpoint choices fail, GraRe succeeds in 3.18--9.81\% of frames by selecting another candidate. Conversely, it fails in 2.11--7.82\% of frames where both endpoint choices succeed. These additional recoveries and regressions jointly determine the benefit of fusion.

\subsection{Sensitivity around the Fixed Weights}

Tables~\ref{tab:sensitivity_neighborhood_realsense} and~\ref{tab:sensitivity_neighborhood_kinect} compare each fixed weight with its neighboring grid values separately for the two cameras. All entries are Average AP (\%). The fixed weights are $\lambda_0=1.0$ for GN, $0.8$ for SBG, EG, and HGGD, and $0.5$ for RNGNet.

\begin{table*}[!t]
\centering
\begin{minipage}[t]{0.48\textwidth}
\centering
{\small
\renewcommand{\arraystretch}{0.82}
\setlength{\tabcolsep}{3pt}
\begin{tabular*}{\columnwidth}{@{\extracolsep{\fill}}lccccc@{}}
\toprule
\textbf{Control} & \textbf{GN} & \textbf{SBG} & \textbf{EG} & \textbf{HGGD} & \textbf{RNGNet} \\
\midrule
Detector & 35.86 & 47.66 & 52.03 & 46.49 & 58.10 \\
Full GraRe & 49.32 & 52.74 & 56.00 & 49.45 & 59.92 \\
$\lambda_0-0.1$ & 49.31 & 52.48 & 55.78 & 49.45 & 59.84 \\
$\lambda_0+0.1$ & -- & 52.86 & 56.12 & 49.37 & 59.90 \\
Quality only & 49.32 & 52.73 & 55.85 & 49.17 & 58.95 \\
\bottomrule
\end{tabular*}
}
\captionof{table}{Average AP (\%) on RealSense at the fixed Score Fusion weights and neighboring grid values. $\lambda_0$ denotes the benchmark weight; $\lambda=1$ uses predicted quality alone.}
\label{tab:sensitivity_neighborhood_realsense}
\end{minipage}\hfill
\begin{minipage}[t]{0.48\textwidth}
\centering
{\small
\renewcommand{\arraystretch}{0.82}
\setlength{\tabcolsep}{3pt}
\begin{tabular*}{\columnwidth}{@{\extracolsep{\fill}}lccccc@{}}
\toprule
\textbf{Control} & \textbf{GN} & \textbf{SBG} & \textbf{EG} & \textbf{HGGD} & \textbf{RNGNet} \\
\midrule
Detector & 30.60 & - & 45.26 & 42.13 & 53.05 \\
Full GraRe & 38.69 & - & 49.61 & 44.94 & 54.14 \\
$\lambda_0-0.1$ & 38.73 & - & 49.29 & 44.91 & 54.16 \\
$\lambda_0+0.1$ & -- & - & 49.81 & 44.88 & 54.06 \\
Quality only & 38.69 & - & 49.59 & 44.71 & 53.06 \\
\bottomrule
\end{tabular*}
}
\captionof{table}{Average AP (\%) on Kinect at the fixed Score Fusion weights and neighboring grid values. $\lambda_0$ denotes the benchmark weight; $\lambda=1$ uses predicted quality alone. SBG--Kinect is unavailable.}
\label{tab:sensitivity_neighborhood_kinect}
\end{minipage}
\end{table*}

The fixed weights lie within 0.20 Average AP points of the highest values observed on the evaluated grid across all nine detector--camera pairs. Changing a fixed weight by one grid step of 0.1, where the perturbed value remains in $[0,1]$, reduces Average AP by at most 0.33 points. Small local changes therefore have limited effects on the aggregate score, whereas larger changes can alter which scenes benefit.

\section{Additional Visualization Analysis}
\label{sec:additional_visualization}

We examine twelve additional GraspNet-1Billion frames across seven detector--camera settings, covering both cameras and all three scene splits. Outcomes follow Equation~\ref{eq:supp_transition_success}. Each pair compares Detector and GraRe choices from the same candidate set, showing the observed RGB-D points from a shared viewpoint and retaining the original candidate poses and widths. Arrows give each candidate's Detector-to-GraRe rank change. RS and KN denote RealSense and Kinect.

\subsection{Successful Re-ranking}

\begin{figure}[!h]
\centering
\includegraphics[width=\columnwidth]{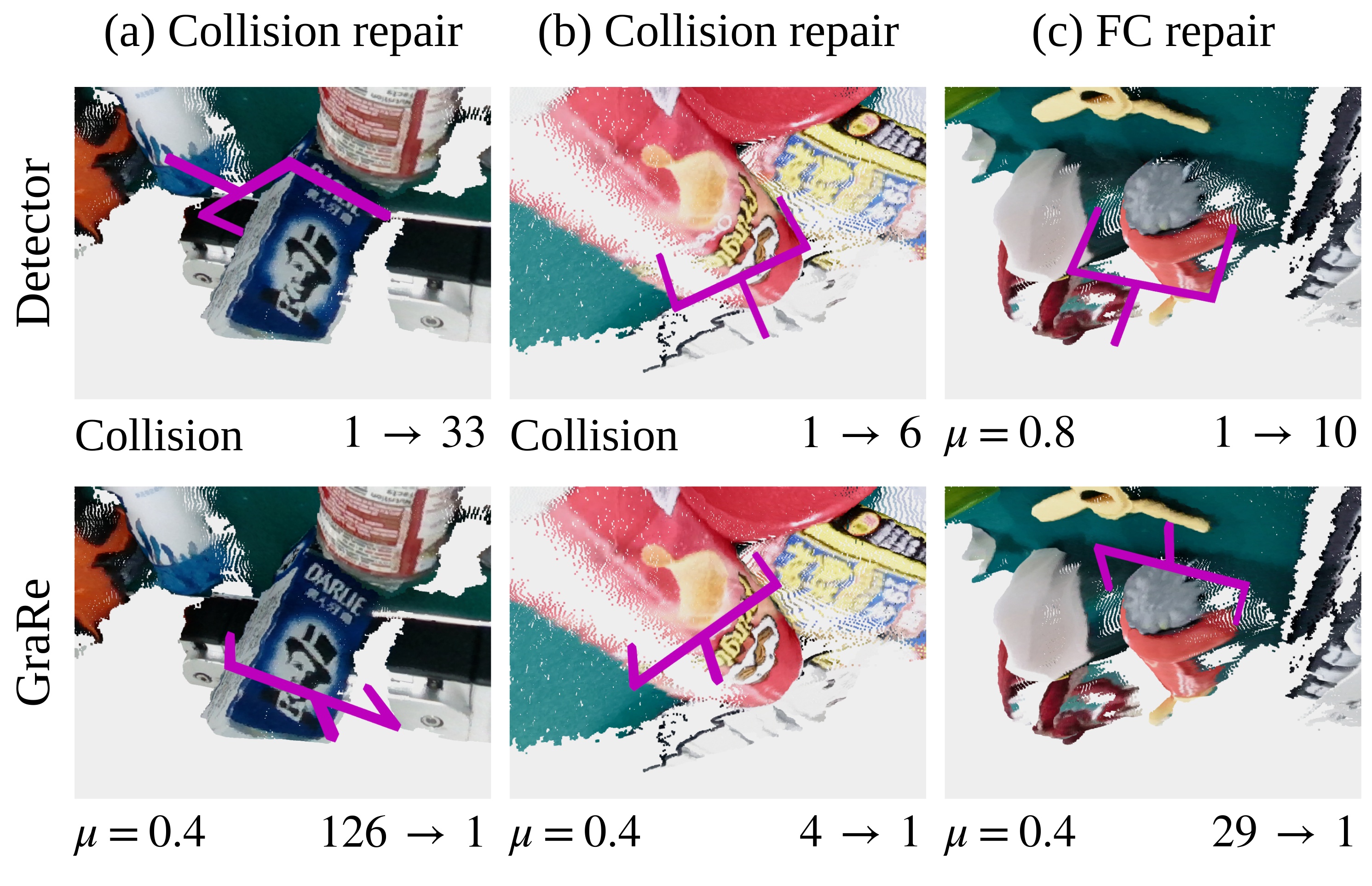}
\caption{Successful re-ranking: (a) GN--RS, Seen; (b) HGGD--KN, Similar; (c) GN--KN, Novel. Detector choices are above GraRe choices. FC denotes force closure.}
\label{fig:additional_visualization_success_1}
\end{figure}

\begin{figure}[!h]
\centering
\includegraphics[width=\columnwidth]{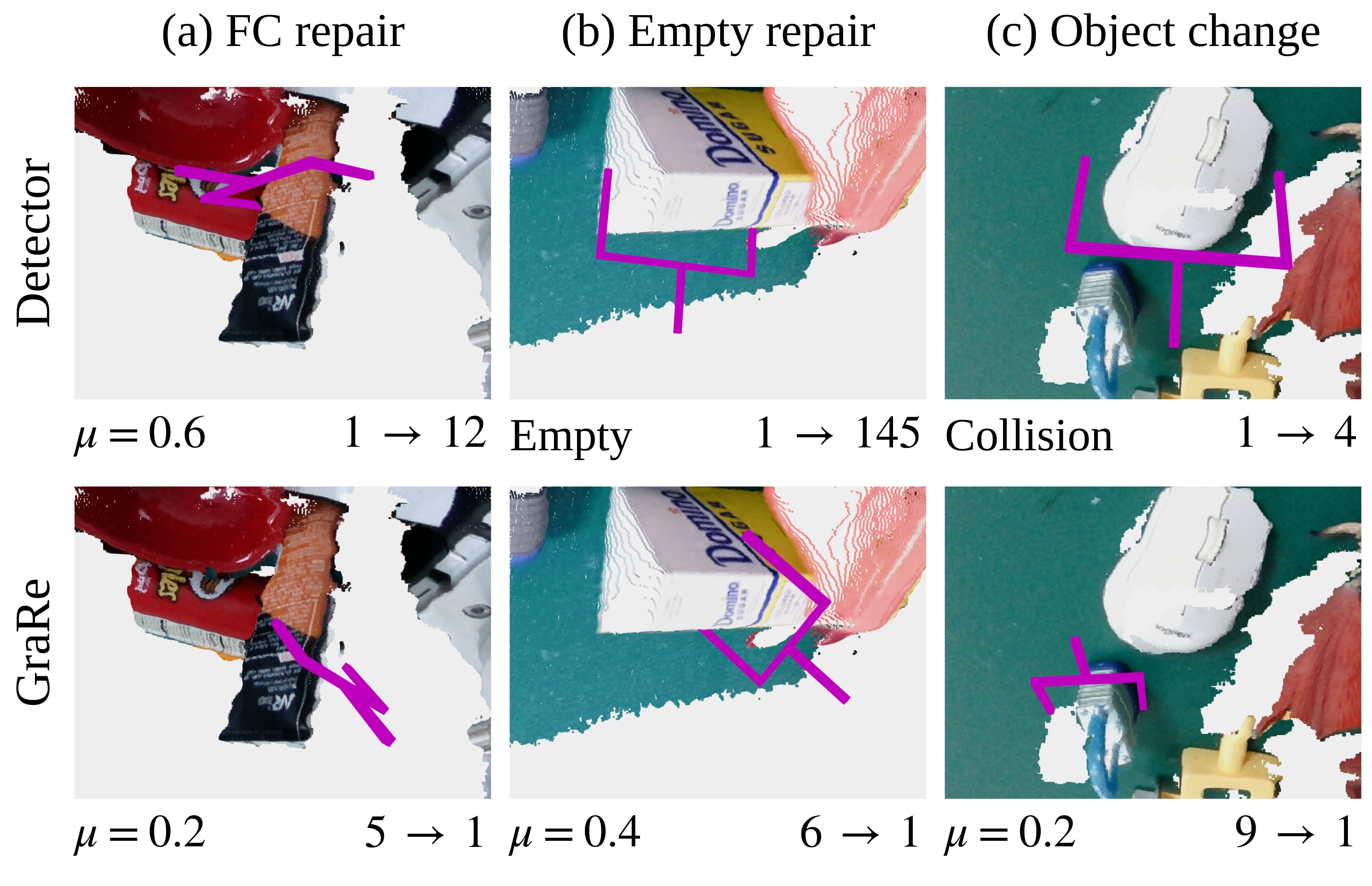}
\caption{Further successful re-ranking: (a) RNGNet--RS, Similar; (b) GN--KN, Similar; (c) SBG--RS, Novel. Columns cover quality differences, an empty-grasp correction, and a change of selected object.}
\label{fig:additional_visualization_success_2}
\end{figure}

Figures~\ref{fig:additional_visualization_success_1} and~\ref{fig:additional_visualization_success_2} show corrections of collision, force-closure, and empty-grasp failures. The collision corrections include different changes in gripper placement. In the HGGD example, the two candidates have closing axes only $11.5^\circ$ apart, but their centers differ by 40.4\,mm. Their different positions relative to the object and its neighbors accompany different collision labels. The GN example instead changes both position and closing direction substantially. These cases illustrate why the grasp neighborhood must be interpreted together with the candidate pose.

The force-closure corrections compare non-empty, collision-free candidates on the same object. In the RNGNet example, the selected candidate has a gripper width 34.1\,mm smaller and a closing axis $82.5^\circ$ away from the original choice, with a lower analytical friction requirement. The empty-grasp correction changes the closing region's position and orientation relative to the carton. The SBG example extends the comparison across objects: GraRe replaces a colliding grasp near the white object with a successful candidate on the neighboring blue object. Re-ranking can therefore change both where an object is grasped and which object is selected, using candidates already present in the detector output.

\subsection{Ranking Regressions}

\begin{figure}[!h]
\centering
\includegraphics[width=\columnwidth]{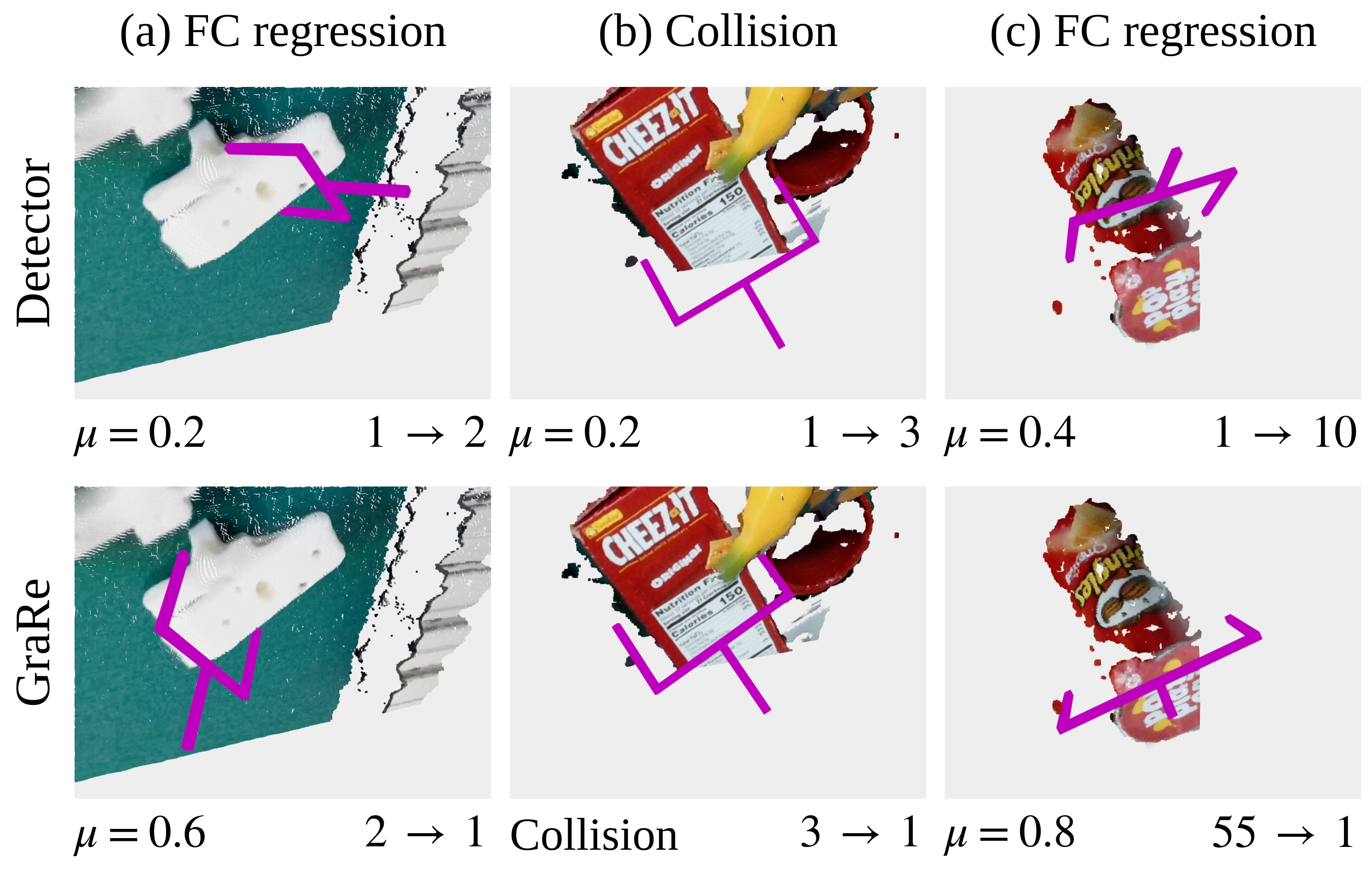}
\caption{Ranking regressions: (a) RNGNet--KN, Novel; (b) HGGD--RS, Seen; (c) GN--RS, Similar. Each successful Detector choice is replaced by a failed GraRe choice.}
\label{fig:additional_visualization_regression}
\end{figure}

Figure~\ref{fig:additional_visualization_regression} shows successful choices replaced by failed candidates. In the RNGNet pair, the closing axes are nearly parallel, yet the centers differ by 41.7\,mm and the gripper widths by 14.8\,mm; the approach orientations also differ. The analytical friction requirement increases from 0.2 to 0.6. Similar closing directions thus do not imply similar grasp quality. The HGGD pair shows a different error: a nearby candidate with a similar orientation receives a higher fused score despite its collision label. GN also replaces a successful candidate with a non-empty, collision-free candidate that fails the friction threshold, showing that quality-only ordering can also make this type of error.

\textbf{Score preferences.} To examine the decision behind a changed order, we compare the two normalized contributions within a frame. For candidates $c_i$ and $c_j$, the main paper's Score Fusion rule gives
\begin{equation}
\begin{aligned}
s_i-s_j={}&(1-\lambda)(\widetilde b_i-\widetilde b_j)\\[-2pt]
&{}+\lambda(\widetilde g_i-\widetilde g_j).
\end{aligned}
\label{eq:supp_visualization_margin}
\end{equation}
Figure~\ref{fig:additional_visualization_scores}(a) decomposes the fused-score difference between the RNGNet GraRe choice and the Detector choice. Detector confidence favors the successful candidate, contributing $-0.033$ to this difference. The opposing quality contribution is $+0.142$, producing a final margin of $+0.109$ for the failed candidate. The predicted quality term outweighs the detector preference and reverses their order.

\subsection{Remaining Failures}

\begin{figure}[!h]
\centering
\includegraphics[width=\columnwidth]{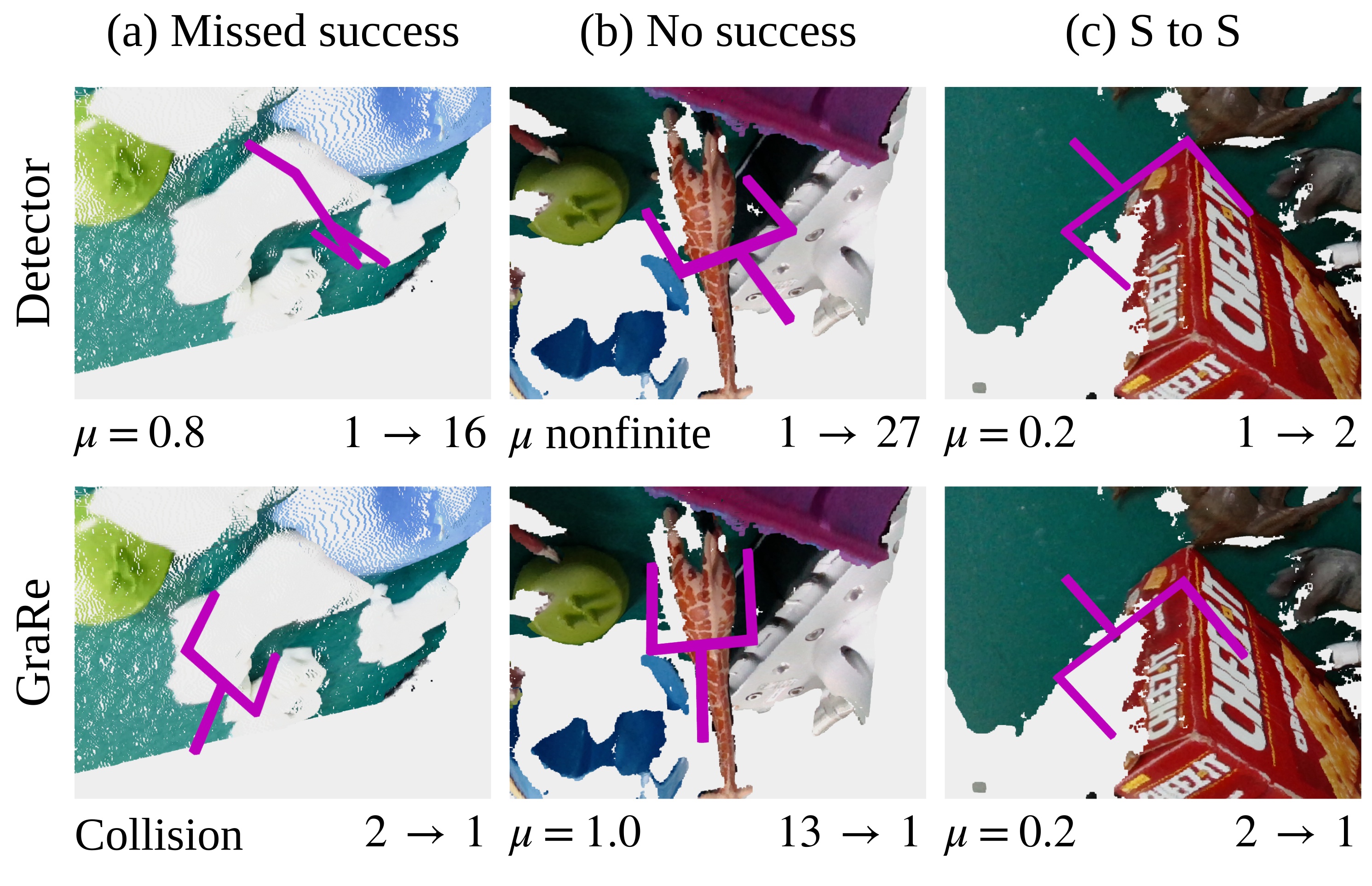}
\caption{Remaining failures and a success-preserving comparison: (a) HGGD--KN, Novel, with successful candidates available; (b) RNGNet--RS, Novel, with none; (c) SBG--RS, Seen.}
\label{fig:additional_visualization_remaining}
\end{figure}

\begin{figure}[!h]
\centering
\includegraphics[width=\columnwidth]{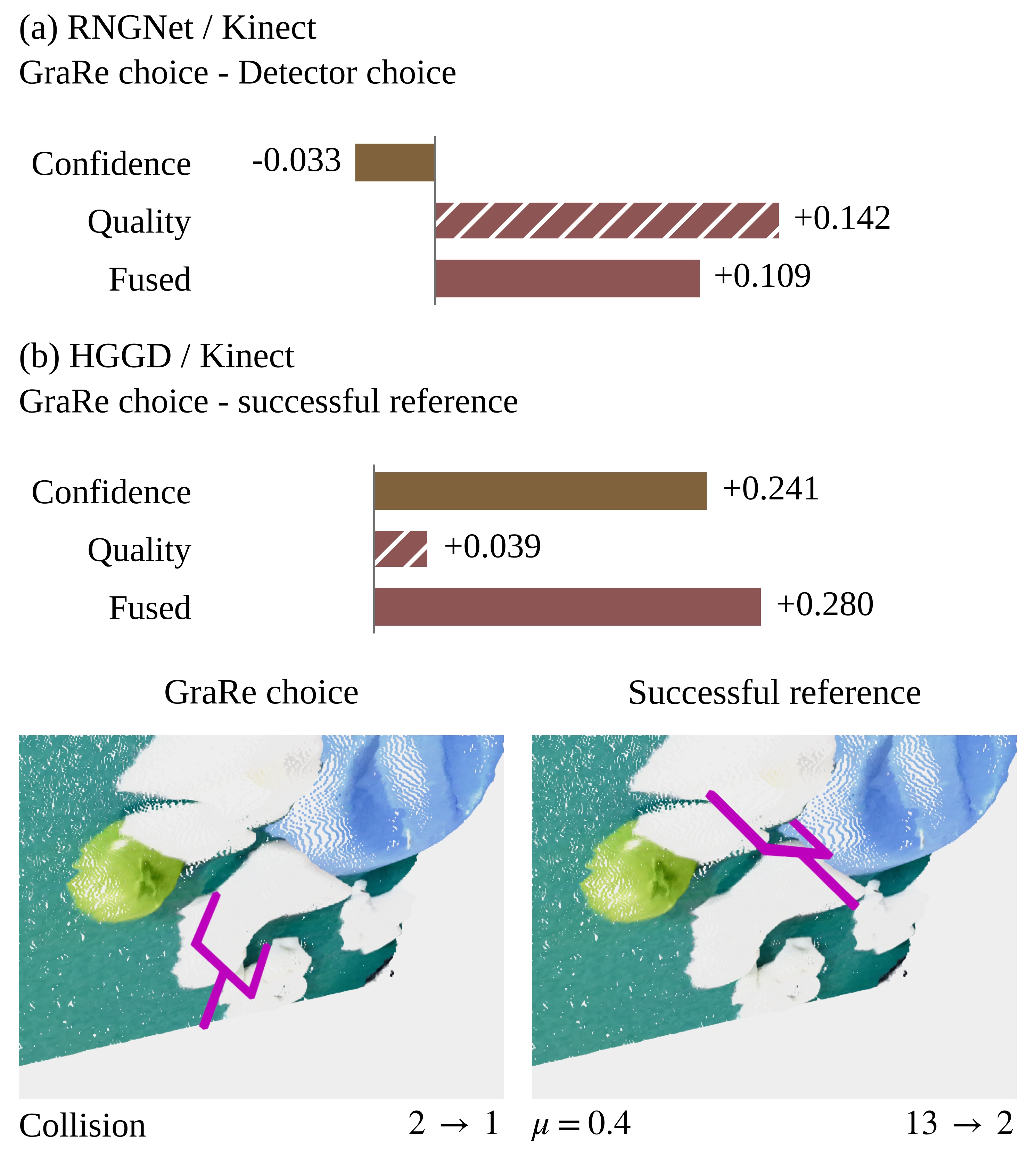}
\caption{Score differences for (a) the RNGNet regression and (b) the HGGD missed-success case. The lower views compare the HGGD choice with the first candidate in GraRe order that satisfies the success criterion.}
\label{fig:additional_visualization_scores}
\end{figure}

Figure~\ref{fig:additional_visualization_remaining} separates two reasons why the leading candidate can remain unsuccessful. In the HGGD frame, 2 of the 66 candidates satisfy the success criterion. GraRe promotes one of them from rank 13 to rank 2, but a colliding candidate remains first. The successful reference in Figure~\ref{fig:additional_visualization_scores}(b) lies on another object. Both detector confidence and predicted quality favor the colliding candidate over this reference, and quality-only ordering also selects the colliding candidate. The frame therefore retains room for better ordering even though GraRe brings a successful candidate close to the front.

The RNGNet frame has no successful candidate among all 113 candidates. Its new leading candidate has a finite friction requirement of 1.0, but still fails the success criterion. Under this fixed candidate set and criterion, changing the order alone cannot produce a successful first choice. The SBG comparison provides a complementary reference: the top two candidates exchange positions while both retain $\mu=0.2$, preserving the binary outcome. Together, the examples distinguish incorrect preferences among available candidates from limits of the candidate set, and show that a changed order need not change the success status.

\subsection{Shell-wise Local Geometry}

\begin{figure}[!ht]
\centering
\includegraphics[width=\columnwidth]{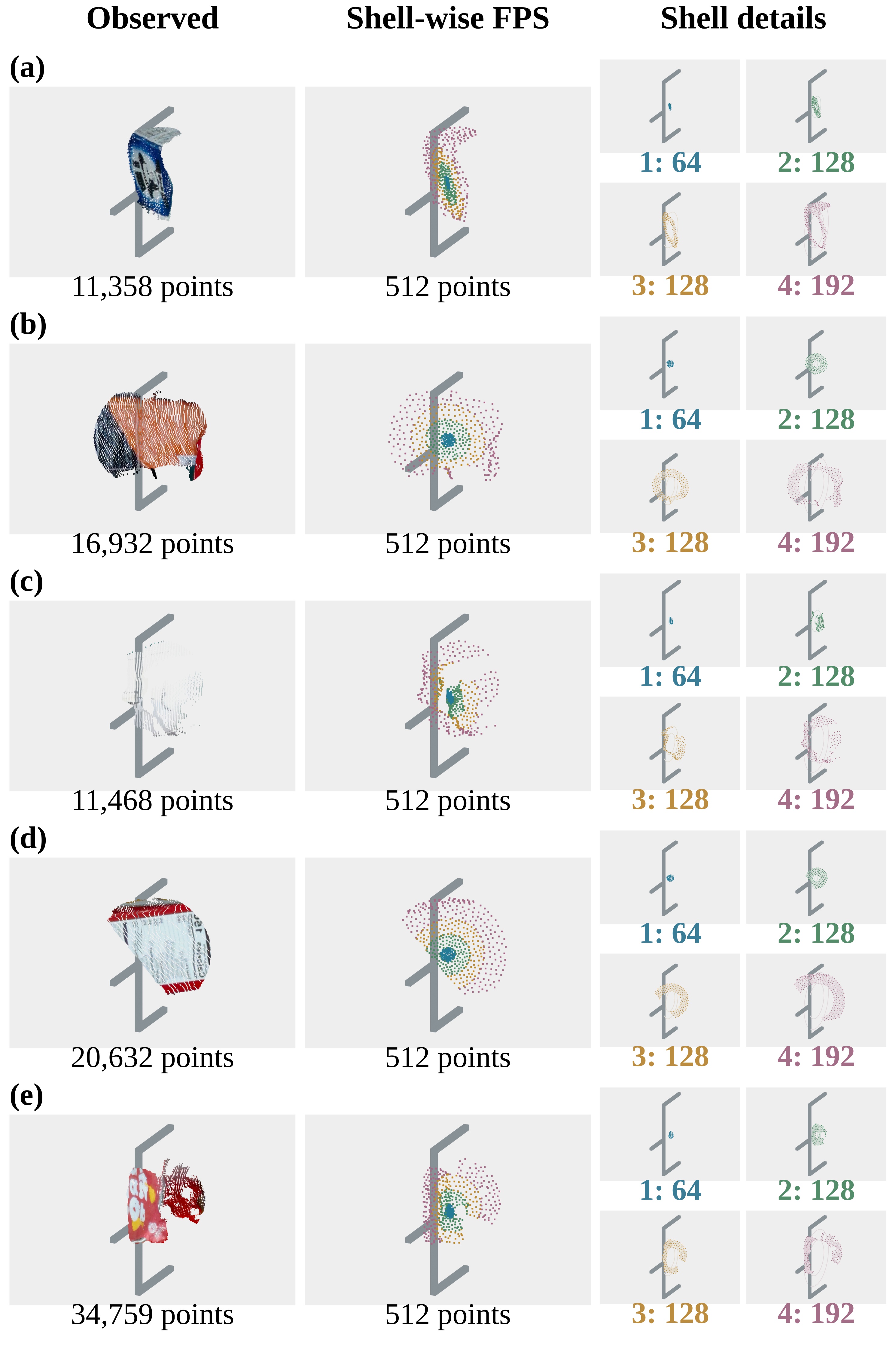}
\caption{Local point-cloud sampling on RealSense. Each row compares the observed neighborhood with the sampled union; the $2\!\times\!2$ details show shell index and retained point count. Rows (a--e) use scenes 120, 138, 160, 125, and 136.}
\label{fig:local_shell_realsense}
\end{figure}

\begin{figure}[!ht]
\centering
\includegraphics[width=\columnwidth]{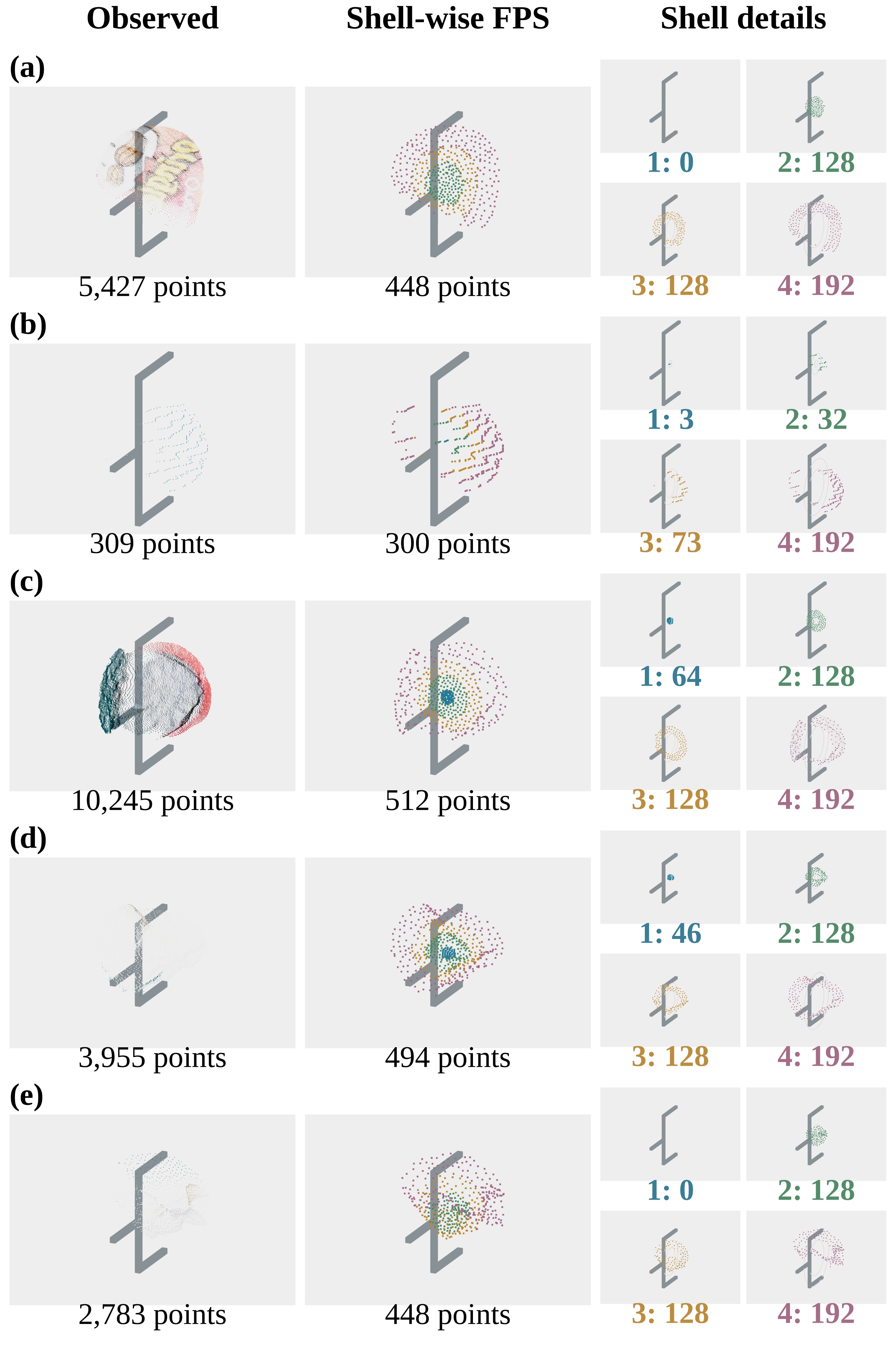}
\caption{Local point-cloud sampling on Kinect, using the same layout and shell colors as Figure~\ref{fig:local_shell_realsense}. Rows (a--e) use scenes 152, 159, 184, 172, and 181.}
\label{fig:local_shell_kinect}
\end{figure}

We show ten local neighborhoods, each alongside the point set retained by shell-wise FPS for the same grasp pose. Figures~\ref{fig:local_shell_realsense} and~\ref{fig:local_shell_kinect} preserve the RGB colors of the input cloud and color the sampled points by radial shell. Each input--output pair shares the same gripper-frame viewpoint and scale. The four smaller panels separate the shells, with faint observed points and highlighted samples. The neighborhood and sampling settings follow Table~\ref{tab:feature_settings}.

Dense neighborhoods are reduced within each radial interval, whereas sparse shells retain their available observations. The RealSense examples fill all four sampling budgets. In Kinect~(a), the innermost shell is empty, and the remaining shells contribute 448 points. Kinect~(b) retains all 3, 32, and 73 observations in the first three shells, while sampling 192 of the 201 outer-shell points. The local representation combines these distance ranges and retains the distinction between densely observed, sparse, and empty shells.

\section{Additional Real-robot Analysis}
\label{sec:additional_real_robot}

The ten real-robot scenes vary in object shape, spacing, and support relations. Their initial configurations show the geometric conditions encountered during grasp selection, while the recorded executions reveal how object removal changes the subsequent scene. We examine selected grasps, unsuccessful attempts, and the progress toward scene clearance in more detail, using the platform and outcome definitions in the main paper.

\textbf{Trial counts.} Table~\ref{tab:real_robot_counts} reports the counts for the six real-robot settings. Each detector and ranking policy contributes ten scene-clearing trials, one for each scene configuration. GSR is computed from the total number of successful attempts divided by the total number of executed attempts across these trials. CR is the fraction of trials ending with a fully cleared scene. A successful attempt removes at least one object from the grasping region. Figure~\ref{fig:additional_real_robot_outcomes} shows the attempt sequences and final scene states for the 30 GraRe trials.

\begin{table*}[!t]
\centering
\begingroup
\normalsize
\setlength{\tabcolsep}{3pt}
\begin{tabular*}{\textwidth}{@{\extracolsep{\fill}}llrrrrr@{}}
\toprule
Detector & Policy & Successes / attempts & GSR (\%) & Cleared / trials & CR (\%) & MCF \\
\midrule
GN & Detector & 58 / 79 & 73.4 & 1 / 10 & 10 & 4 \\
GN & GraRe & 72 / 81 & 88.9 & 10 / 10 & 100 & 2 \\
\midrule
SBG & Detector & 57 / 81 & 70.4 & 3 / 10 & 30 & 6 \\
SBG & GraRe & 68 / 78 & 87.2 & 9 / 10 & 90 & 2 \\
\midrule
EG & Detector & 60 / 88 & 68.2 & 7 / 10 & 70 & 4 \\
EG & GraRe & 71 / 78 & 91.0 & 10 / 10 & 100 & 2 \\
\bottomrule
\end{tabular*}
\endgroup
\caption{Counts and rates for the six real-robot settings. GSR uses successful and executed attempts as its numerator and denominator. CR uses fully cleared and evaluated trials. MCF is the longest consecutive failure streak within a trial.}
\label{tab:real_robot_counts}
\end{table*}

\subsection{Scene Composition}

\begin{figure}[!ht]
\centering
\includegraphics[width=\columnwidth]{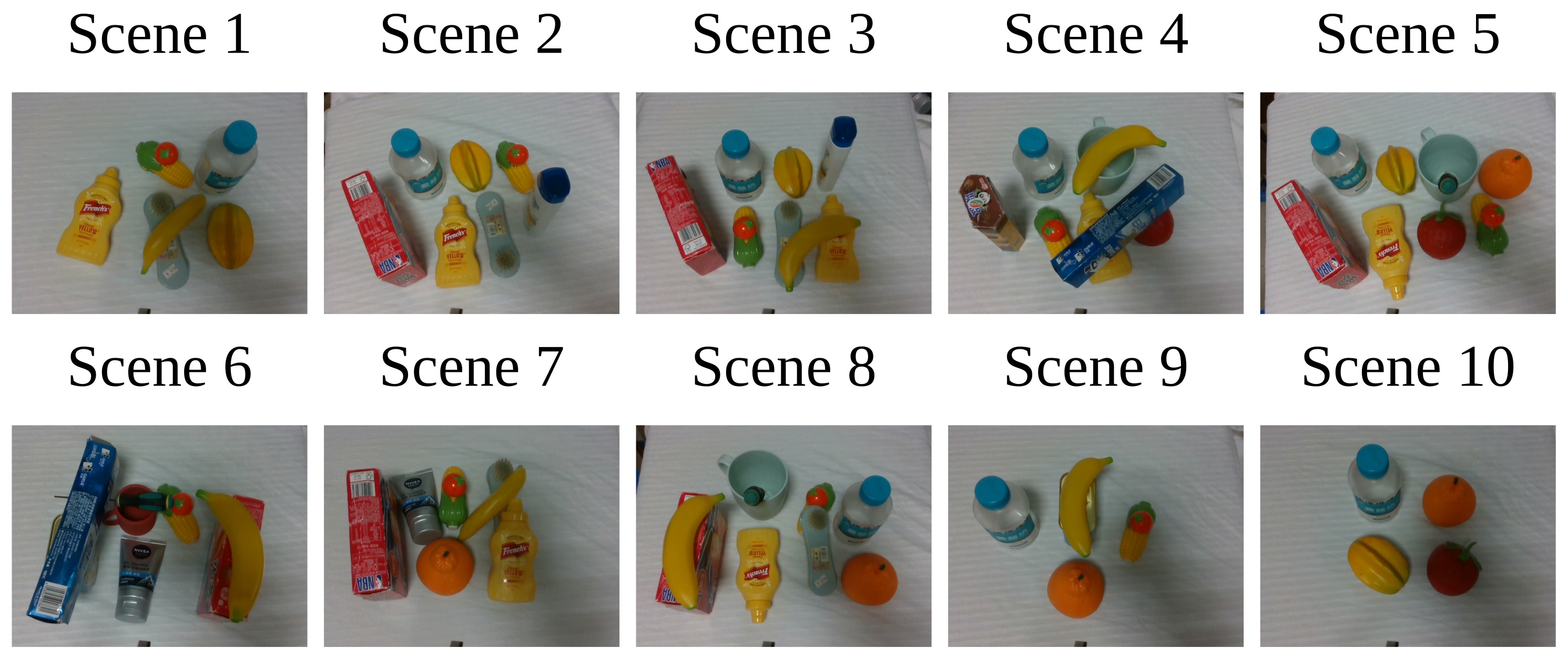}
\caption{Representative initial configurations of Scenes 1--10, shown using the SBG--GraRe observations.}
\label{fig:additional_real_robot_scenes}
\end{figure}

Figure~\ref{fig:additional_real_robot_scenes} shows the ten scene configurations, and Table~\ref{tab:additional_real_robot_scenes} describes their objects and grasping conditions. Scenes 2 and 10 provide more exposed object surfaces, whereas Scenes 1, 3, and 7 contain partial overlaps. Scenes 4, 6, and 8 include objects resting on one another, and Scenes 5 and 8 place a screwdriver inside a mug. These arrangements vary the possible finger contacts and free space for gripper motion. Scene 9 further shows that a sparse arrangement can still contain a local support constraint.

\begin{table}[!ht]
\centering
{\small
\renewcommand{\arraystretch}{1.04}
\setlength{\tabcolsep}{2pt}
\begin{tabular}{@{}p{.11\columnwidth}p{.395\columnwidth}p{.445\columnwidth}@{}}
\toprule
\textbf{Scene} & \textbf{Objects} & \textbf{Grasping conditions} \\
\midrule
1 & {\raggedright Drink bottle, mustard bottle, brush, banana, starfruit, vegetable toys\par} & {\raggedright The banana crosses the brush, with containers of different heights nearby.\par} \\
2 & {\raggedright Drink bottle, mustard bottle, shampoo bottle, brush, red carton, starfruit, vegetable toys\par} & {\raggedright Mostly separated objects expose different shapes and heights.\par} \\
3 & {\raggedright Scene 2 objects and a banana\par} & {\raggedright The banana rests across the brush and mustard bottle, adding overlap and support relations.\par} \\
4 & {\raggedright Drink bottle, mustard bottle, mug, two cartons, banana, strawberry, vegetable toys\par} & {\raggedright The banana spans the cup opening, and a long carton rests across lower objects.\par} \\
5 & {\raggedright Drink bottle, mustard bottle, mug, screwdriver, red carton, orange, strawberry, starfruit, vegetable toys\par} & {\raggedright A screwdriver sits inside the mug, with only part of its shape exposed.\par} \\
6 & {\raggedright Two cartons, metal tin, mug, screwdriver, cream tube, banana, vegetable toys\par} & {\raggedright The carton, banana, and screwdriver rest on other objects, creating several support relations.\par} \\
7 & {\raggedright Red carton, cream tube, mustard bottle, brush, banana, orange, vegetable toys\par} & {\raggedright The banana overlaps the brush among nearby curved and flatter objects.\par} \\
8 & {\raggedright Drink bottle, mustard bottle, mug, screwdriver, red carton, brush, banana, orange, vegetable toys\par} & {\raggedright The scene combines an object inside a mug, stacked objects, and partial occlusion.\par} \\
9 & {\raggedright Drink bottle, metal tin, banana, orange, vegetable toys\par} & {\raggedright The banana rests on the tin within an otherwise sparse arrangement.\par} \\
10 & {\raggedright Drink bottle, orange, strawberry, starfruit\par} & {\raggedright Four separated objects expose cylindrical, rounded, and ridged shapes.\par} \\
\bottomrule
\end{tabular}
}
\caption{Objects and grasping conditions in Scenes 1--10. Fruit and vegetable names describe shape.}
\label{tab:additional_real_robot_scenes}
\end{table}

\subsection{Ranking and Execution}

\begin{figure}[!ht]
\centering
\includegraphics[width=\columnwidth]{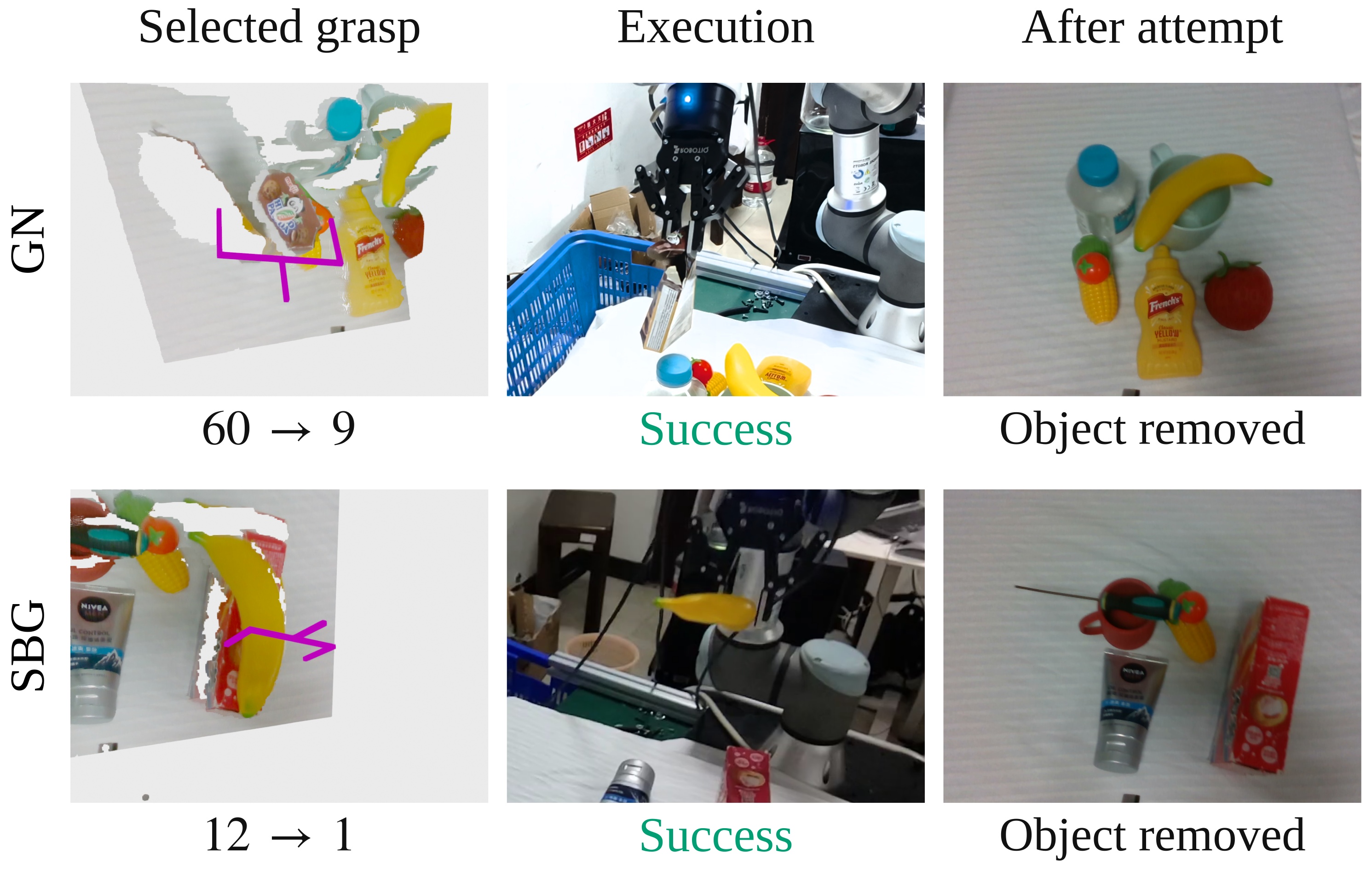}
\caption{Successful execution of selected candidates: GN in Scene 4, attempt 2 (top), and SBG in Scene 6, attempt 3 (bottom).}
\label{fig:additional_real_robot_cases_1}
\end{figure}

\begin{figure*}[!t]
\begin{minipage}[t]{0.482\textwidth}
\vspace{0pt}
\centering
\includegraphics[width=\columnwidth]{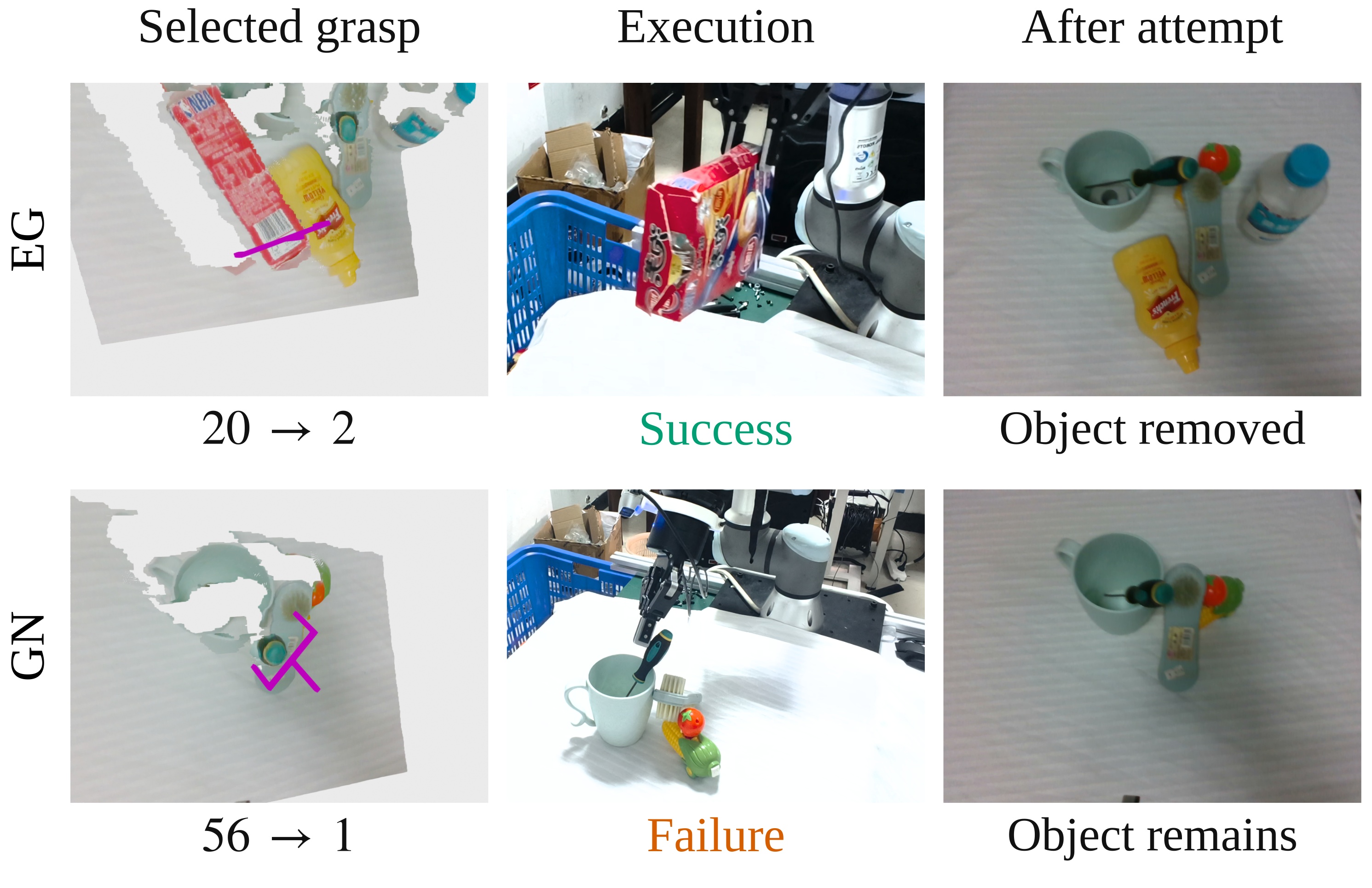}
\caption{A successful EG grasp in Scene 8, attempt 3 (top), and an unsuccessful GN grasp in Scene 8, attempt 7 (bottom).}
\label{fig:additional_real_robot_cases_2}
\end{minipage}
\hfill
\begin{minipage}[t]{0.482\textwidth}
\vspace{0pt}
\centering
\includegraphics[width=.85\columnwidth]{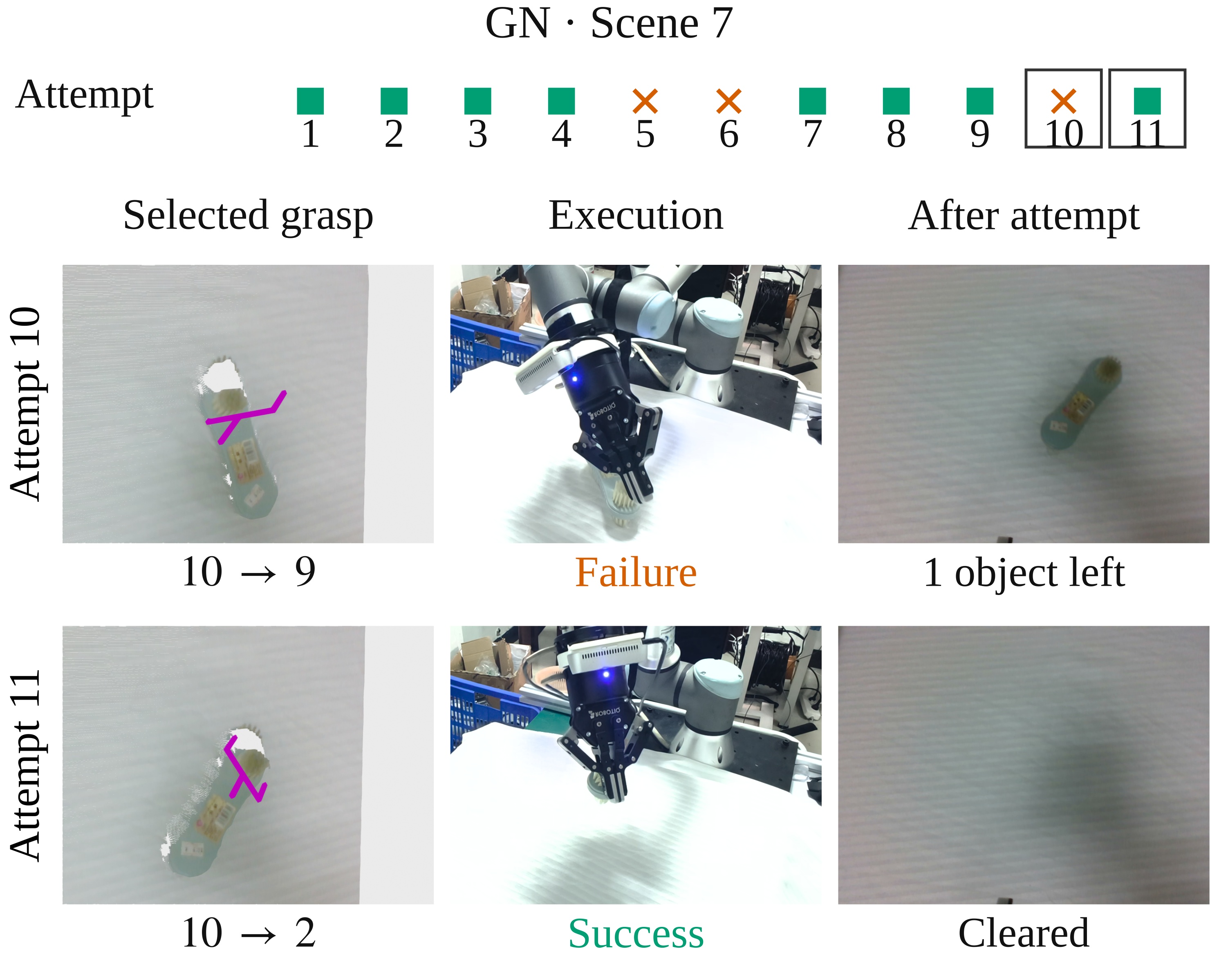}
\caption{GN Scene 7: the brush remains after attempt 10 and is removed on attempt 11. The outcome strip marks the attempts shown in detail.}
\label{fig:additional_real_robot_gn_recovery}
\end{minipage}
\end{figure*}

Figures~\ref{fig:additional_real_robot_cases_1} and~\ref{fig:additional_real_robot_cases_2} connect selected candidates to their physical execution and the scene afterward. For each attempt, the robot selects an executable candidate in the chosen order; its arrow compares that candidate's ranks under $\pi^{\mathcal D}$ and the final GraRe order $\pi^{\mathrm R}$ on the same observation. The robot observes the scene again before selecting a candidate for the next attempt.

Candidates ranked low under $\pi^{\mathcal D}$ can still support successful physical grasps. GN removes a carton using a candidate promoted from rank 60 to rank 9, SBG lifts a banana from a carton using a candidate promoted from rank 12 to rank 1, and EG removes a red carton using a candidate promoted from rank 20 to rank 2. These grasps span carton side faces or the curved body of the banana, and the execution frames show the objects held during lifting. The GN example in Scene 8 shows a remaining error: the candidate promoted from rank 56 to rank 1 targets a screwdriver inside a mug, but the gripper withdraws without lifting it. A large ranking change can therefore bring forward either an effective grasp or a physical failure.

\subsection{Failures and Scene Completion}

The execution sequences show how object removal resumes after unsuccessful attempts. In GN Scene 7, the brush is briefly lifted and falls back on attempt 10, then is removed on attempt 11 (Figure~\ref{fig:additional_real_robot_gn_recovery}). Comparing the selected candidates in the robot base frame, the closing axes differ by $80.7^\circ$ and the candidate width decreases from 69.4 to 53.3\,mm. The successful attempt thus uses a different grasp orientation on the remaining brush.

\begin{figure*}[!t]
\captionsetup{type=figure}
\begin{minipage}[t]{0.482\textwidth}
\vspace{0pt}
{
\centering
\begin{minipage}{.85\columnwidth}
\includegraphics[width=\columnwidth]{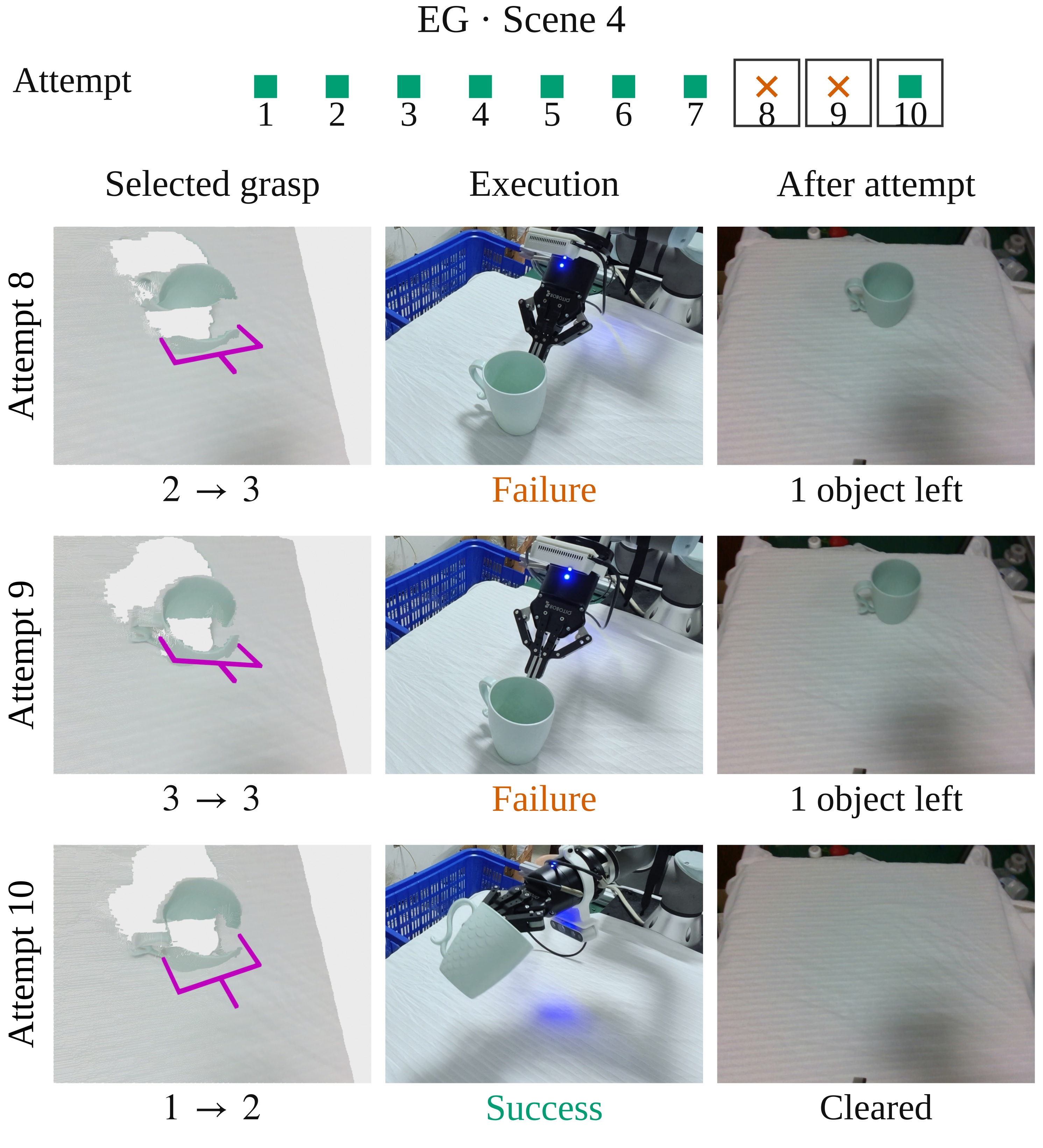}
\end{minipage}
\caption{EG Scene 4: two unsuccessful attempts precede removal of the mug on attempt 10.}
\label{fig:additional_real_robot_eg_recovery}
}
\end{minipage}
\hfill
\begin{minipage}[t]{0.482\textwidth}
\vspace{0pt}
{
\centering
\begin{minipage}{.85\columnwidth}
\includegraphics[width=\columnwidth]{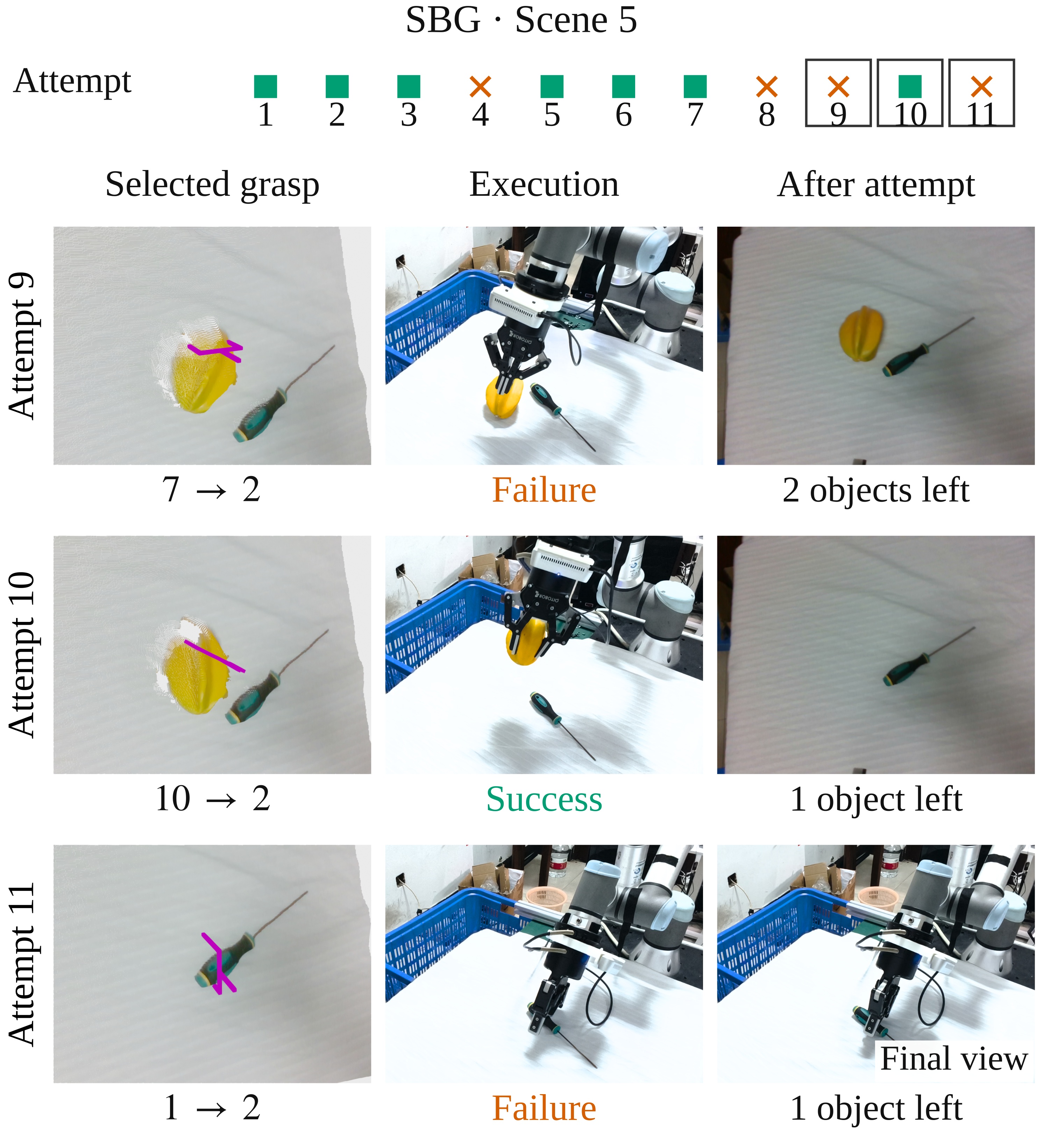}
\end{minipage}
\caption{SBG Scene 5: attempts 9--11 include removal of the starfruit, leaving the screwdriver. The last image is the final recorded view.}
\label{fig:additional_real_robot_sbg_remaining}
}
\end{minipage}
\end{figure*}

In EG Scene 4, the gripper withdraws without lifting the mug on attempts 8 and 9, before removing it on attempt 10 (Figure~\ref{fig:additional_real_robot_eg_recovery}). The two unsuccessful grasps have nearly the same approach direction. Relative to attempt 9, the successful candidate changes its closing and approach directions by $22.6^\circ$ and $22.2^\circ$, respectively. These sequences show different grasp selections across successive observations, with successful removal completing each scene. Even with little clutter remaining, grasp pose and possible finger contacts continue to matter.

SBG Scene 5 shows partial recovery without complete clearance (Figure~\ref{fig:additional_real_robot_sbg_remaining}). The starfruit is briefly lifted and lost on attempts 8 and 9, then removed on attempt 10. Between attempts 9 and 10, the selected approach direction stays nearly unchanged, while the closing axis changes by $29.9^\circ$ and the candidate width increases from 44.9 to 60.7\,mm. Its removal leaves the screwdriver as the sole remaining object. The next grasp targets the screwdriver handle, but the screwdriver remains on the surface at the end of the recording. Successful removal can therefore resume while the scene remains incomplete.

\begin{figure*}[!t]
\captionsetup{type=figure}
\begin{minipage}[t]{0.482\textwidth}
\vspace{0pt}
{
\centering
\begin{minipage}{.95\columnwidth}
\includegraphics[width=\columnwidth]{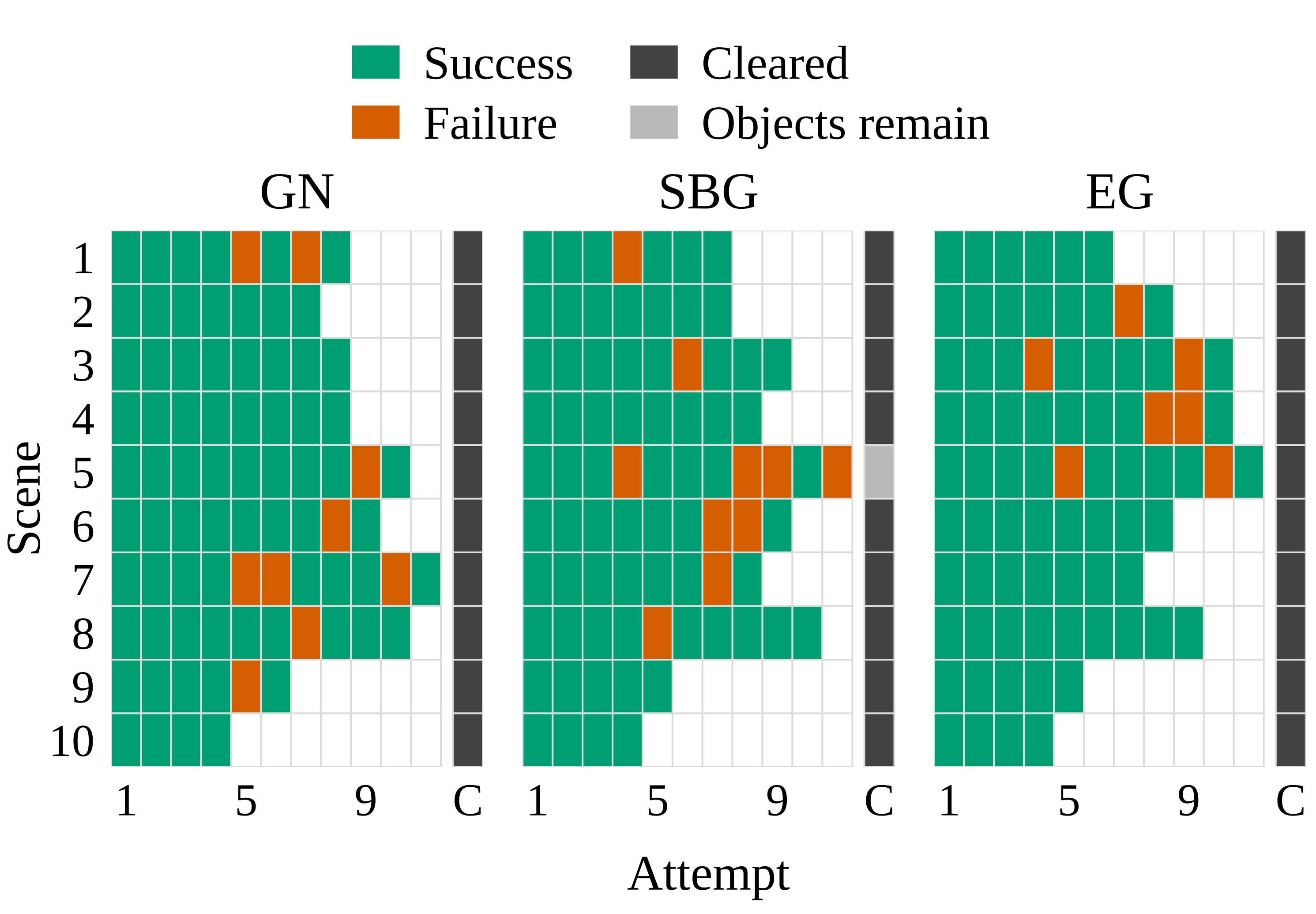}
\end{minipage}
\caption{Outcomes of successive grasp attempts in all 30 GraRe trials. Column C gives the final scene status; white cells mark positions with no attempt.}
\label{fig:additional_real_robot_outcomes}
}
\end{minipage}
\hfill
\begin{minipage}[t]{0.482\textwidth}
\vspace{0pt}
{
\centering
\begin{minipage}{.95\columnwidth}
\includegraphics[width=\columnwidth]{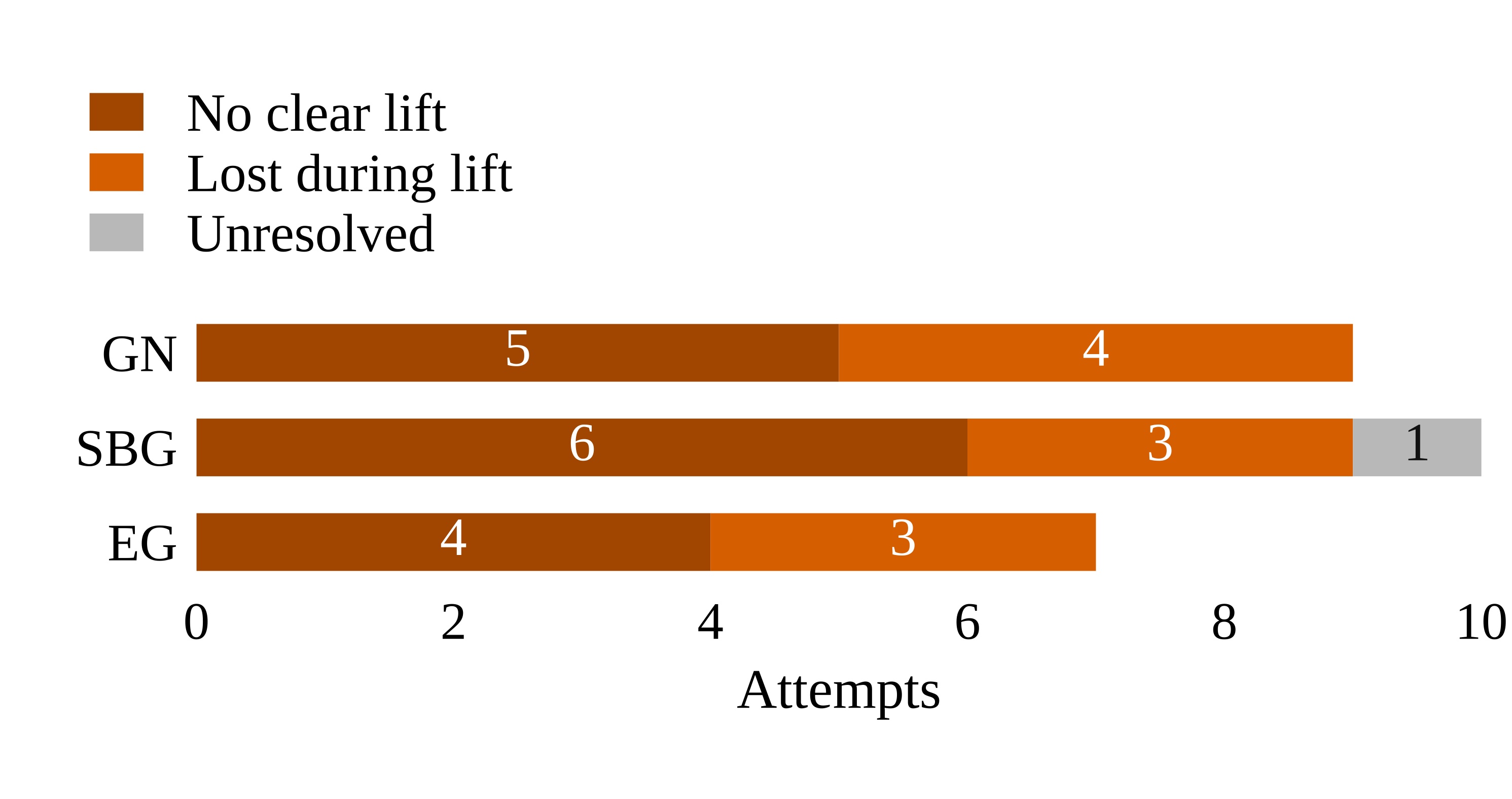}
\end{minipage}
\caption{Visible stages of unsuccessful executions. The remaining recording ends before withdrawal is complete, leaving its failure stage unresolved.}
\label{fig:additional_real_robot_failure_stages}
}
\end{minipage}
\end{figure*}

Figure~\ref{fig:additional_real_robot_outcomes} places these examples within all 30 GraRe trials. Across the complete sequences, 25 unsuccessful attempts are followed by another attempt, and 21 of these next attempts succeed. Failures are thus often followed by renewed object removal, consistent with the high completion rates reported in the main paper.

Among the 26 unsuccessful executions (Figure~\ref{fig:additional_real_robot_failure_stages}), 15 show no clear object lift and 10 show an object lifted but lost during lifting. The remaining recording ends before withdrawal is complete, leaving its failure stage unresolved. The observed failures include both difficulty establishing a grasp and difficulty retaining an object during motion, as illustrated by the mug and brush sequences.

\begin{figure*}[!t]
\captionsetup{type=figure}
\centering
\begin{minipage}[t]{0.482\textwidth}
\vspace{0pt}
\centering
\includegraphics[width=\columnwidth]{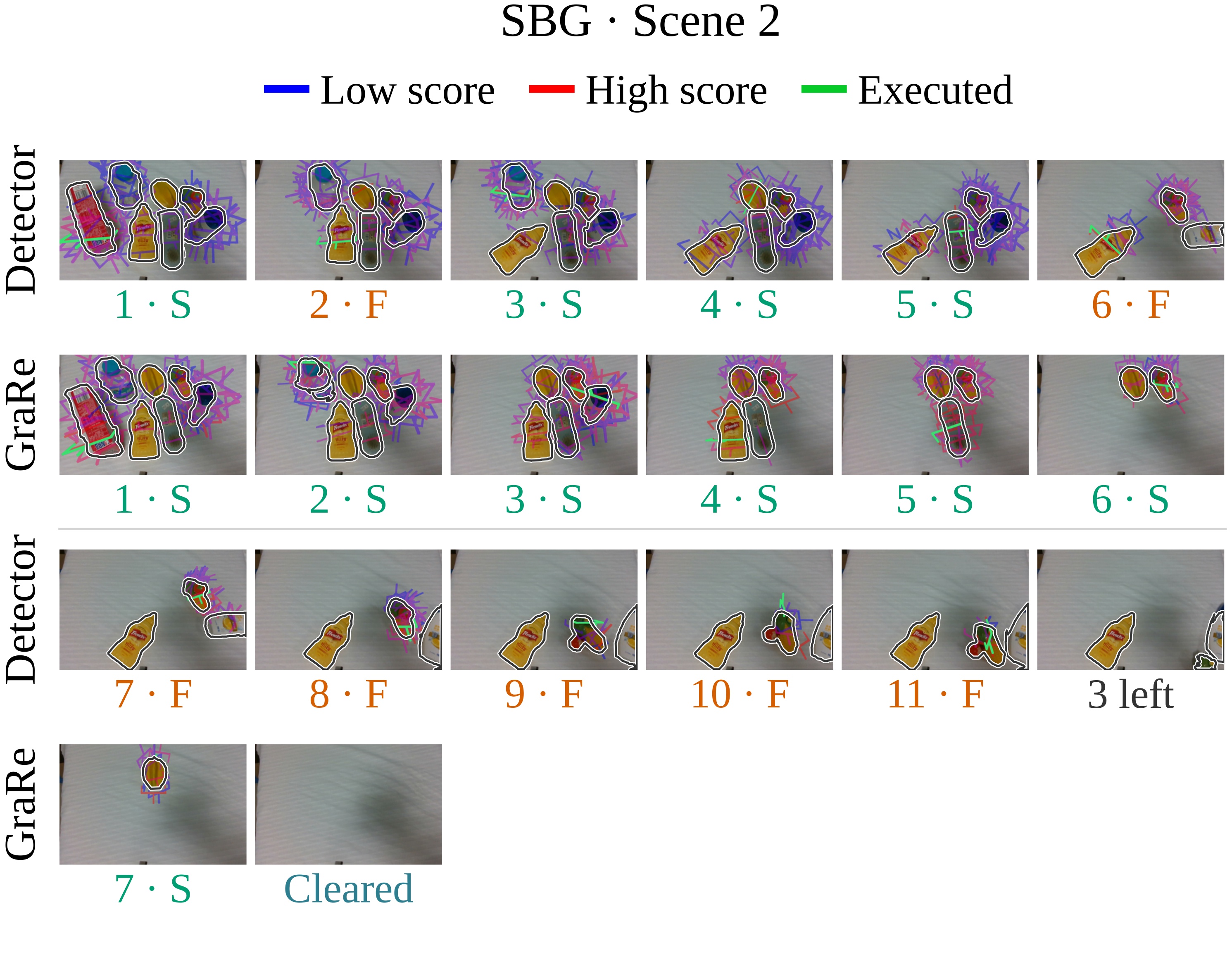}
\caption{Complete grasping sequences for SBG Scene 2. Each block contains up to six consecutive observations per method. Labels give attempt numbers; S and F denote success and failure. Final views show the remaining scene.}
\label{fig:real_robot_paired_sbg}
\end{minipage}
\hfill
\begin{minipage}[t]{0.482\textwidth}
\vspace{0pt}
\centering
\includegraphics[width=\columnwidth]{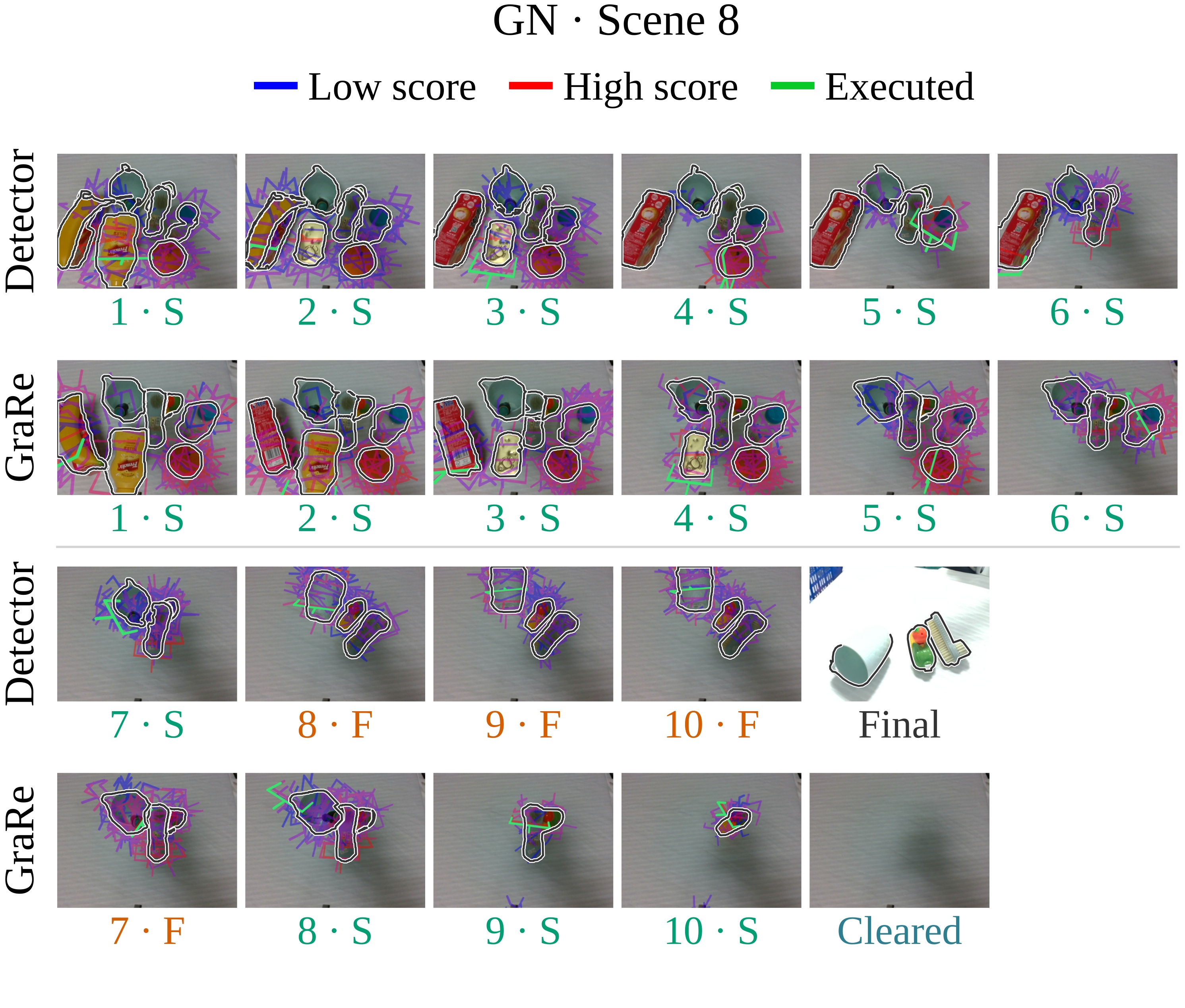}
\caption{Complete grasping sequences for GN Scene 8, using the layout of Figure~\ref{fig:real_robot_paired_sbg}. The final Detector view shows the remaining objects from an external camera.}
\label{fig:real_robot_paired_gn}
\end{minipage}
\end{figure*}

\begin{table*}[!t]
\centering
\begingroup
\normalsize
\setlength{\tabcolsep}{3pt}
\begin{tabular*}{\textwidth}{@{\extracolsep{\fill}}lrrrrrr@{}}
\toprule
& \multicolumn{2}{c}{GN} & \multicolumn{2}{c}{SBG} & \multicolumn{2}{c}{EG} \\
\cmidrule(lr){2-3}\cmidrule(lr){4-5}\cmidrule(lr){6-7}
Stage & Detector & GraRe & Detector & GraRe & Detector & GraRe \\
\midrule
Timed observations & 78 & 81 & 81 & 78 & 88 & 78 \\
Mean candidates & 61.40 & 50.36 & 58.33 & 55.33 & 40.58 & 46.94 \\
\midrule
Instance segmentation & 1657.23 & 1516.64 & 1615.70 & 1408.96 & 1510.30 & 1414.71 \\
Input loading & 27.29 & 25.78 & 24.52 & 23.25 & 22.85 & 23.60 \\
Detector input sampling & 9.90 & 9.14 & 5.30 & 5.29 & 5.40 & 5.33 \\
Detector forward and decoding & 194.33 & 193.82 & 386.48 & 373.73 & 131.47 & 94.56 \\
Detector post-processing & 364.79 & 386.64 & 290.88 & 279.79 & 351.14 & 339.22 \\
\midrule
Object FPS and association$^{\dagger}$ & -- & 400.08 & -- & 237.45 & -- & 243.81 \\
Local-shell construction$^{\dagger}$ & -- & 413.11 & -- & 263.79 & -- & 235.26 \\
Prediction and ranking$^{\dagger}$ & -- & 25.75 & -- & 21.00 & -- & 24.24 \\
\midrule
Scene-safety checks & 0.00 & 0.00 & 5.58 & 5.38 & 3.95 & 4.59 \\
\midrule
Additional GraRe latency & -- & 838.94 & -- & 522.25 & -- & 503.31 \\
Total inference latency & 2253.54 & 2971.10 & 2328.47 & 2618.76 & 2025.10 & 2385.43 \\
\bottomrule
\end{tabular*}
\endgroup
\caption{Per-observation inference latency on an RTX 2060. Values are means in milliseconds except the observation and candidate counts. $^{\dagger}$ marks GraRe-specific stages. Their subtotal gives the additional GraRe latency. Detector totals sum the relevant stage timings. GraRe totals combine instance segmentation with the recorded candidate-processing time. Each policy is averaged over its own trial observations.}
\label{tab:runtime_breakdown}
\end{table*}

\subsection{Paired Scene-Clearing Sequences}

Figures~\ref{fig:real_robot_paired_sbg} and~\ref{fig:real_robot_paired_gn} follow complete Detector and GraRe trials of two scene configurations. For each attempt, candidate grasps and the executed pose are shown on the current observation. The next observation records the scene after execution, with contours marking visible object regions. Candidate colors range from blue to red with increasing score within each view; green marks the executed grasp.

Figure~\ref{fig:real_robot_paired_sbg} shows how the two SBG trials diverge as the table becomes less crowded. Detector succeeds at attempts 1, 3, 4, and 5, but attempts 6--11 all fail. The mustard bottle, shampoo bottle, and vegetable toys remain visible through these final six attempts and in the last view. Candidate grasps are still proposed around the residual objects, but the executed choices do not sustain object removal. GraRe completes seven successful attempts. By its sixth observation, only the starfruit and vegetable toys remain; the sixth grasp removes the vegetable toys, and the seventh removes the starfruit. The difference between the trials therefore persists after the initial clutter has been reduced.

In GN Scene 8 (Figure~\ref{fig:real_robot_paired_gn}), both trials begin with successive successful grasps. Detector succeeds on its first seven attempts but fails on attempts 8--10, leaving the mug, brush, and vegetable toys. GraRe succeeds on its first six attempts and fails on attempt 7, when it selects the screwdriver inside the mug. Its next three attempts all succeed: attempt 8 removes the mug together with the screwdriver, attempt 9 removes the brush, and attempt 10 removes the remaining vegetable toys. The change from the screwdriver to the mug is visible in the executed poses, while the following observations show the remaining objects disappearing in sequence. These final three successes turn an interrupted sequence into a completed trial.

These sequences link attempt-level success to the final state of the scene. Early removals reduce the number of objects, but the last decisions still concern distinct geometries: bottles resting on the table, a screwdriver within a mug, and small residual groups. GSR is the percentage of successful attempts, whereas CR is the percentage of trials that end with a fully cleared scene. The late failures and renewed removals visible here explain why both measures are informative when evaluating repeated grasping.

\raggedbottom
\subsection{Online Inference Latency}

\label{sec:online_inference_latency}

We report mean inference latency per RGB-D observation on an NVIDIA RTX 2060 during the real-robot scene-clearing experiments. The measured stages cover instance segmentation and candidate processing. Image acquisition, scene reconstruction, motion planning, and robot motion are excluded. The real-robot system uses UCN for instance segmentation. Network forward timings use CUDA synchronization. Statistics include all timed execution cycles from the evaluated trials, including startup cycles. Table~\ref{tab:runtime_breakdown} gives the observation counts and mean candidate counts for each detector and ranking policy.

Detector latency sums the times for instance segmentation, input loading and sampling, detector forward inference and decoding, candidate post-processing, and scene-safety checks. Candidate post-processing includes collision filtering, non-maximum suppression, and the remaining detector-side processing. Both ranking policies include instance segmentation. Summing these stages gives Detector latencies of 2.25, 2.33, and 2.03 s for GN, SBG, and EG, respectively.

GraRe adds object preprocessing, local-shell construction, and prediction and ranking. Object preprocessing samples fixed-size object point clouds using FPS and associates grasp candidates with object instances. Local-shell construction expresses neighboring scene points in the gripper coordinate frame and applies FPS within each radial shell. Prediction and ranking combine Candidate Features, Local Geometric Features, and Point-MAE-based Object-Context Features to estimate grasp quality, fuse scores, and reorder candidates. This final stage is timed as a whole.

The three GraRe-specific stages together take 838.94, 522.25, and 503.31 ms for GN, SBG, and EG, respectively. Object preprocessing and local-shell construction account for 95.2\%--96.9\% of this additional latency. Prediction and ranking take 21.0--25.8 ms per observation, making geometric input construction the main source of added computation. Including segmentation and the detector stages, the recorded GraRe inference paths take 2.97, 2.62, and 2.39 s. Each policy's total is averaged over the observations encountered in its own trials. GraRe's additional latency is calculated from its three marked stages in Table~\ref{tab:runtime_breakdown}.

\end{document}